\documentclass[10pt,journal,compsoc]{IEEEtran}
\usepackage[nocompress]{cite}

\ifCLASSOPTIONcompsoc
\usepackage[nocompress]{cite}
\else
\usepackage{cite}
\fi
\ifCLASSINFOpdf
\else
\fi
\usepackage{amsmath} 
\usepackage{graphicx}
\usepackage{epsfig}
\usepackage{epstopdf}
\usepackage{caption}
\usepackage{tabularx}
\usepackage[table]{xcolor}
\usepackage{breakcites}
\newcommand{\etal}{\textit{et al.}}

\usepackage{makecell}
\usepackage{xurl}
\usepackage[colorlinks,linkcolor=red,urlcolor=blue,anchorcolor=blue,citecolor=green]{hyperref}
\usepackage[hyphenbreaks]{breakurl}
\usepackage{amsfonts}
\newcolumntype{H}{>{\setbox0=\hbox\bgroup}c<{\egroup}@{}}
\definecolor{editcolor}{RGB}{0,0,0}
\definecolor{babypink}{rgb}{0.96, 0.76, 0.76}
\definecolor{mygray}{RGB}{221,221,221} 
\definecolor{ForestGreen}{RGB}{34,139,34}

\usepackage{arydshln}
\usepackage{bm}
\usepackage{graphicx}
\usepackage{array}
\usepackage{booktabs} 
\usepackage{bm}
\usepackage{arydshln}
\usepackage{marvosym}
\renewcommand{\arraystretch}{0.999}
\let\oldsubsection\subsection
\renewcommand{\subsection}[1]{%
	\vspace{-3mm}%
	\oldsubsection{#1}%
}
\usepackage{enumitem}
\usepackage{multirow} 
\usepackage[caption=false]{subfig}
\usepackage{tcolorbox} 
\newcommand{\revisefigure}[1]{%
	{#1}
}
\newcommand{\revise}[1]{{#1}}

\begin{document}

\title{SMC++: Masked Learning of Unsupervised Video Semantic Compression}


\author{Yuan~Tian,
	Xiaoyue~Ling,
	Cong~Geng,
	Qiang~Hu,
	Guo~Lu\textsuperscript{\Letter},
	and~Guangtao~Zhai\textsuperscript{\Letter},~\textit{Fellow, IEEE}
\IEEEcompsocitemizethanks{
	\IEEEcompsocthanksitem Yuan Tian is with the Shanghai Artificial Intelligence Laboratory, Shanghai, China. E-mail: tianyuan168326@outlook.com.
	Cong~Geng is with the JIUTIAN Research, E-mail: gengcong@cmjt.chinamobile.com. 
	Qiang Hu is with the Cooperative Medianet Innovation Center, Shanghai Jiao Tong University, Shanghai, China, E-mail: qiang.hu@sjtu.edu.cn.
	Xiaoyue Ling, Guo Lu and Guangtao Zhai are with the Institute of Image Communication and Network Engineering, Shanghai Jiao Tong University, Shanghai, China, E-mail: \{xiaoyue\_ling,luguo2014,zhaiguangtao\}@sjtu.edu.cn.
	Corresponding Authors\textsuperscript{\Letter}: Guo Lu and Guangtao Zhai.
}
}
\markboth{Journal of \LaTeX\ Class Files,~Vol.~14, No.~8, August~2015}%
{Shell \MakeLowercase{\textit{et al.}}: Bare Demo of IEEEtran.cls for Computer Society Journals}

\IEEEtitleabstractindextext{%
\begin{abstract}
Most video compression methods focus on human visual perception, neglecting semantic preservation. This leads to severe semantic loss during the compression, hampering downstream video analysis tasks.
In this paper, we propose a Masked Video Modeling (MVM)-powered compression framework that particularly preserves video semantics, by jointly mining and compressing the semantics in a self-supervised manner.
While MVM is proficient at learning generalizable semantics through the masked patch prediction task, it may also encode non-semantic information like trivial textural details, wasting bitcost and bringing semantic noises. To suppress this, we explicitly regularize the non-semantic entropy of the compressed video in the MVM token space.
The proposed framework is instantiated as a simple Semantic-Mining-then-Compression (SMC) model.
Furthermore, we extend SMC as an advanced SMC++ model from several aspects. First, we equip it with a masked motion prediction objective, leading to better temporal semantic learning ability.
Second, we introduce a Transformer-based compression module, to improve the semantic compression efficacy.
Considering that directly mining the complex redundancy among heterogeneous features in different coding stages is non-trivial, we introduce a compact blueprint semantic representation to align these features into a similar form, fully unleashing the power of the Transformer-based compression module.
Extensive results demonstrate the proposed SMC and SMC++ models show remarkable superiority over previous traditional, learnable, and perceptual quality-oriented video codecs, on three video analysis tasks and seven datasets. \textit{Codes and model are available at: \url{https://github.com/tianyuan168326/VideoSemanticCompression-Pytorch}.}
\end{abstract}
}
\maketitle
\IEEEdisplaynontitleabstractindextext

%
\IEEEpeerreviewmaketitle
 
 \vspace{-8mm}
\IEEEraisesectionheading{ \section{Introduction}}
 \IEEEPARstart{V}{ideo} compression has been researched over the past few decades.
 Most methods, including traditional~\cite{sullivan2012overview}\cite{bross2021overview} and learnable ones~\cite{hu2021fvc}\cite{li2021deep}, aim at accurately reconstructing the video pixels, rather than preserving semantic information such as object shapes. These methods usually perform unfavorably on downstream AI tasks~\cite{yi2021benchmarking}\cite{tanaka2022does}.

 To tackle this problem, lots of research efforts have been devoted to emphasizing the coding of semantics-relevant information.
 For example, early works~\cite{zhang2016joint}\cite{chao2015keypoint}\cite{baroffio2015hybrid} and standards~\cite{duan2018compact}\cite{duan2015overview}\cite{duan2013compact} additionally transport the manually-designed image descriptors.
 Later, some methods\cite{galteri2018video}\cite{choi2018high} improve the traditional codec with hand-crafted designs to better cope with the specific tasks.
 Meanwhile, some methods~\cite{choi2018near}\cite{chen2019lossy}\cite{chen2019toward}\cite{singh2020end}\cite{feng2022image}\cite{sheng2024vnvc} compress the feature maps of AI models instead of the images, where the downstream task modules shall be fine-tuned to accommodate the features.
 Besides, scalable coding methods~\cite{liu2021semantics}\cite{yang2021towards}\cite{yan2021sssic}\cite{choi2021latent}\cite{choi2022scalable11}\cite{sun2020semantic}\cite{lin2023deepsvc} divide the bitstream into two parts, \textit{i.e.}, one for analysis tasks and another one for video reconstruction.
The former part is usually optimized by task-specific losses.

Despite these progresses, rare methods are unsupervised and also support out-of-the-box deployment. \textit{Unsupervised} represents that the semantic learning procedure is supervised by the video itself, without leveraging any human-annotated task labels. This is friendly to data scarcity scenarios and eliminates the labor of video annotation. \textit{Out-of-the-box} denotes the compressed videos can be directly consumed by the pre-trained downstream task models, without adapting or fine-tuning these numerous task models. This reduces the deployment cost and improves the practicality.
 
We refer to this novel and challenging problem as Unsupervised Video Semantic Compression (UVSC).
 Similar to traditional video compression~\cite{bhaskaran1997image}, UVSC also pursues better video quality at the given bitrate.
 But, the main discrepancy lies in the quality metric. Traditional video compression focuses on low-level metrics like PSNR or SSIM~\cite{wang2004image}. Conversely, UVSC prefers a good semantic quality, which is an open challenge, since the high-level semantics is not well-defined like low-level information.

 In this paper, motivated by the superior semantic learning capability of the masked video modeling (MVM)~\cite{he2022masked}\cite{tong2022videomae} scheme, we propose the first MVM-powered video semantic compression framework. During the optimization, we mask out a large proportion of the compressed video patches, and use the unmasked parts to predict the masked regions, facilitating the compressed video patches by our framework to maintain their semantic attributes.
 Further, we notice that, while the MVM scheme is powerful in learning semantics, its inner generative learning paradigm also facilitates the coding framework memorizing non-semantic information~\cite{dong2022bootstrapped}, which makes the learned semantic feature consuming extra bitcost, as well as noisy.

To suppress this deficiency, we explicitly decrease the non-semantic information entropy of the MVM feature space, by formulating it as a parameterized Gaussian Mixture Model conditioned on the mined video semantics. The alternative semantic learning and non-semantic suppressing procedures make the system bootstrapping itself toward more efficient semantic coding.
As a result, it shows remarkable results on a wide range of video analysis tasks.

As for the specific instantiation, we first build a simple Semantic-Mining-then-Compensation (SMC) model, which compensates and amends the semantic loss problem of current lossy video codecs. Specifically, the semantic features of the original video and the lossy video are extracted on the encoder side, and only the residual semantics is compressed by a convolutional neural network. During decoding, the residual-compensated semantic feature is synthesized as the videos, deployed for various AI tasks.

Furthermore, we extend the SMC model to a more advanced SMC++ model from the following aspects:
 
 \textit{First, we augment the masked learning task by additionally predicting motion.} Motion is another critical cue in videos, which benefits a series of video tasks. For example, motion information is necessary for discriminating the visually-ambiguous actions~\cite{simonyan2014two}.
 Therefore, we introduce another decoder to the MVM task header, for predicting the motion targets of the masked regions, expecting to enhance the temporal semantic modeling capability of our framework.
 
\textit{Second, we improve the coding efficacy by introducing a Blueprint-guided compression Transformer (Blue-Tr).} The basic SMC model adopts simple convolution operations for compression, which can not adequately model complex semantic redundancies. The self-attention~\cite{vaswani2017attention} operation is more powerful,
but still faces challenges in capturing redundancies among heterogeneous features, which are distributed in various coding stages of the compression system.
To address this issue, we propose a multi-step ``Blueprint mining-feature aligning-redundancy modeling'' approach. First, we mine the most critical semantic part of the current frame, which we call blueprint semantics.
Subsequently, we employ the blueprint to align diverse heterogeneous features that are distributed in different timestamps and coding layers.
Finally, given the aligned features, a decomposed Transformer model is adopted to capture the semantic redundancies.
Our contributions are:

 \begin{itemize}
 	\setlength{\itemsep}{0pt}
 	\setlength{\parsep}{0pt}
 	\setlength{\parskip}{0pt}
 	\item  We propose the first {masked learning} framework for the unsupervised video semantic compression problem, aiming to better support various semantic analysis tasks under low-birate conditions.
 	This framework is instantiated as a simple SMC model.
 	
 	\item We devise the Non-Semantics Suppressed (NSS) learning strategy to better adapt the general masked learning scheme to the compression problem, aiming to suppress the encoding of non-semantic information.
 	
 	\item We further extend SMC as SMC++ by introducing \textit{1)} a masked motion modeling objective and \textit{2)} a new Blueprint-guided compression Transformer (Blue-Tr),
 	 for \textit{1)} learning better temporal semantics and \textit{2)} compressing semantics more effectively.
 	 The newly proposed Blue-Tr introduces a blueprint semantics concept, and uses it to align diverse features. The aligned features are easier to compress, fully unleashing the power of the Transformer-based compression module.

 	\item The proposed models SMC and SMC++ demonstrate notable superiority over previous traditional, learnable, and perceptual codecs, on three video analysis tasks and seven datasets.
 	Moreover, we append SMC++ with a small detail rendering network. The resulting SMC++* model achieves superior visual detail coding performance, on four video compression datasets.
 \end{itemize}

This work extends our preliminary conference version~\cite{tian2023non} with the following substantial improvements. 
1) We theoretically demonstrate that optimizing the NSS loss is equivalent to maximizing the mutual information between the compressed video patch tokens and their corresponding semantic labels, thereby improving semantic separability. In addition, we introduce an extra masked motion prediction term to the original appearance-only MVM task loss. We also offer further analysis and visualization of the masked learning procedure in the supplementary material.
2) We propose a Blueprint-guided compression Transformer (Blue-Tr) to achieve more effective semantic compression, upgrading the SMC model to the SMC++ model. By appending it with a lightweight detail decoder, we obtain the PSNR-oriented SMC++* model.
3) We conduct more comprehensive evaluations, including results on video question answering and fine-grained action recognition tasks, results on LLM-based downstream models, the flexibility of using multiple codecs (VVC, H.264, H.265, and DCVC-FM) as the base codec, and comparisons with more recent methods. Ablation studies are performed on MOT and VOS tasks, on hyperparameter selection, teacher-model-based MAE learning, various loss terms, attention mechanisms, the number of historical frames used as context, etc.

\vspace{-3mm}
 \section{Related Works}
 \label{sec:related_work}
 \textbf{Video Compression.}
 Previous video codecs, including traditional ones~\cite{wiegand2003overview}\cite{sullivan2012overview}\cite{bross2021overview} and learnable ones~\cite{hu2021fvc}\cite{li2021deep}\cite{lu2019dvc}\cite{li2022hybrid}\cite{li2023neural1}\cite{li2024neural}\cite{hu2025vrvvc}\cite{hu2025varfvv}\cite{hu20254dgc}, are designed to achieve better pixel-wise signal quality metrics, \textit{e.g.}, PSNR and MS-SSIM~\cite{wang2003multiscale}, which mainly serve the human visual experience. Recently, there are also some generative video coding methods~\cite{yang2021perceptual}\cite{tian2024free} that mainly consider visual comfort and perceptual quality.
Despite recent methods achieving significant strides in low-level visual metrics like PSNR and SSIM, their effectiveness on semantic AI tasks is still undesirable~\cite{tanaka2022does}\cite{chen2025just}\cite{aibench}\cite{zhang2025lmmsurvey}\cite{li2025image}\cite{chen2024gaia}\cite{li2025r}.
This prompts the research on the video semantic compression problem.

 \textbf{Video Coding for Machine (VCM).}
 Early standards such as CDVA~\cite{duan2013compact} and CDVS~\cite{duan2015overview} propose to pre-extract and transport the image keypoints, supporting image indexing or retrieval tasks.
 Some works~\cite{choi2018near}\cite{chen2019lossy}\cite{chen2019toward}\cite{singh2020end}\cite{feng2022image} compress the intermediate feature maps instead of images.
 Besides, some works~\cite{choi2018high}\cite{huang2021visual} \cite{cai2021novel}\cite{zhang2021just} improve traditional codecs by introducing downstream task-guided rate-distortion optimization strategy or another task-specific feature encoding stream~\cite{veselov2021hybrid}\cite{chao2015keypoint}.
 Also, some methods~\cite{duan2020video}\cite{choi2022scalable11} optimize the learnable codecs by directly incorporating the downstream task loss.
 Recently, some methods exploit hand-crafted structure maps~\cite{hu2020towards}\cite{duan2022jpd} for semantic coding.
Most above methods rely on task-specific labels or hand-crafted priors, which may not be generalizable enough.
 
Recently, some works have leveraged self-supervised representation learning methods for learning a compact semantic representation.
As a pioneering work, Dubois~\etal\cite{dubois2021lossy} theoretically revealed that the distortion term of the lossy rate-distortion trade-off for image classification can be approximated by a contrastive learning objective~\cite{he2020momentum}. However, the compressed semantics are empirically effective to a group of tasks that share a similar prior (\textit{i.e.
}, labels are invariant to data augmentations), but may severely discard the semantics useful to other tasks such as detection, segmentation, and even medical tasks~\cite{tian2025medical}\cite{tian2025semantics}\cite{tian2025towards}\cite{yuan2025rofi}. Feng~\etal~\cite{feng2022image}\cite{chen2024cross}\cite{jiang2025advancing} proposed to learn a unified feature representation for AI tasks from unlabeled data in a similar manner.
In these methods, the downstream models are required to be fine-tuned to adapt to the features.
 
\revise{As for the incorporation of motion information, most previous approaches~\cite{yi2022task}\cite{ge2024task} directly adopt the architecture of PSNR-oriented codecs~\cite{lu2019dvc}\cite{li2021deep}, utilizing motion information such as optical flow to compress inter-frame redundancy. CMVC~\cite{zhang2024video} integrates motion to guide a diffusion model for generating smooth and natural videos, in ultra-low-bitrate video generative compression. HMFVC~\cite{huang2022hmfvc} extracts semantic information from motion flow for machine analysis. However, none of the previous approaches utilize motion as a learning target, to facilitate preserving temporal semantics during compression.}
 
Our previous work~\cite{tian2022coding} presented a one-bit semantic map that efficiently compresses video semantics into a compact space, advancing semantic compression. Nonetheless, it faces certain restrictions. Firstly, the one-bit semantic map was heuristically hand-designed, based on the assumption that image edges retain most semantics, such as object structures. This work eliminates such human-derived priors. Secondly, the previous approach relies on contrastive learning, which is inferior to the masked learning technique adopted in this work, as evidenced in numerous papers.

 \textbf{Scalable Coding and Visual-Semantic Fusion Coding.}
 {Scalable coding} methods~\cite{hu2020towards}\cite{yang2021towards}\cite{liu2021semantics}\cite{yan2021sssic}\cite{choi2021latent}\cite{choi2020task} can achieve excellent compression efficiency when measured with the trained tasks, but usually show undesirable results on the tasks/data out of the training scope, due to the supervised learning paradigm. 
 {Visual-semantic fusion coding} methods~\cite{duan2022jpd}\cite{akbari2019dsslic}\cite{huang2021deep}\cite{hu2020towards}, \textit{a.k.a}, conceptual coding~\cite{chang2019layered}\cite{chang2022conceptual}\cite{chang2021thousand}\cite{chang2022consistency}, first extract the structure information and the texture information on the encoding side, and then fuse the two parts into a image on the decoding side. 
 The fused images are readily fed into various tasks and achieve superior performance even at low bitrate levels.
 However, almost all these methods employ semantic segmentation map~\cite{duan2022jpd} or edge map~\cite{hu2020towards} as the semantic stream, not fully discarding the task-specific priors.

 \textbf{Compressed Video Analysis.}
 There are also amounts of works such as~\cite{wang2019fast}\cite{li2022end} perform video analysis tasks, such as
 action recognition~\cite{wu2018compressed}\cite{shou2019dmc}\cite{carreira2017quo}\cite{tian2022ean}\cite{tian2020self}
  and multiple object tracking (MOT)~\cite{khatoonabadi2012video}, in the compressed video domain.
 However, these methods focus on developing video analysis models that better leverage the partially decoded video stream, such as the motion vector.
 In contrast, our work focuses on the coding procedure.

 \textbf{Self-Supervised Semantic Learning.}
 Recent methods can be mainly divided into two catogories, \textit{i.e.}, Contrative Learning (CL) ones ~\cite{he2020momentum}\cite{chen2020improved} and Masked Auto-Encoder (MAE) learning ones~\cite{he2022masked}\cite{tong2022videomae}.
MAE simply predicts masked patches from unmasked ones, while showing remarkably strong performance in downstream tasks. 
Our work adapts the MAE to better address the compression problem, by explicitly suppressing non-semantic information within its feature space.
 
\revise{ 
\textbf{Motion-Enhanced MAE.}
There are also some works that utilize motion information as the MAE target, aiming to capture more temporal semantics in videos.
MGMAE~\cite{huang2023mgmae} leverages optical flow to build a temporally-consistent masking strategy, reducing information leakage across frames.
Masked Motion Encoding~\cite{sun2023masked} and MotionMAE~\cite{yang2024motionmae} and Song~\etal~\cite{song2022takes} extend the reconstruction task by not only predicting static appearance, but also motion trajectories from masked regions. MASA~\cite{zhao2024masa} introduces a motion-aware masked autoencoder for sign language recognition.
Our work is the first to introduce the masked motion reconstruction target to the field of video compression.
}

  \begin{figure*}[!t]
 	\centering
 	\includegraphics[width=17.5cm]{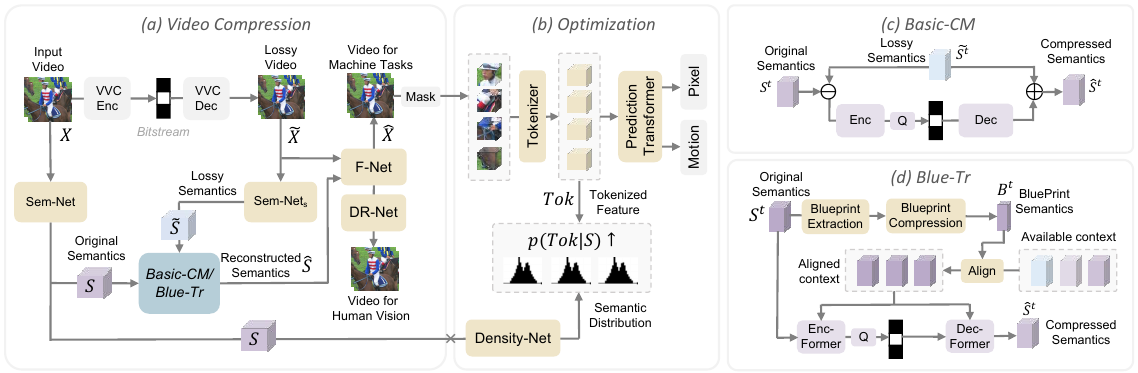}
 	\vspace{-2mm}
 	\caption {
 		Framework overview.
 		First, the semantic features of the original and the lossy videos are separately extracted by Sem-Net. Then, the original semantic feature is compressed with the aid of the lossy semantic by Basic-CM/Blue-Tr.
 		Finally, the reconstructed semantic feature and the lossy video are fused by F-Net, generating the semantically-sound video that supports various analysis tasks.
 		The framework is optimized via a non-semantics suppressed Masked Video Modeling (MVM) objective. 
 		In the basic model SMC,
 		 the MVM task only predicts the pixels, 
 		 meanwhile 
 		 the semantics is compressed by a simple residue-based compression module (Basic-CM).
 		In the improved model SMC++, a motion-prediction MVM objective is further incorporated for better temporal semantic modeling, and a powerful Blueprint-guided compression Transformer (Blue-Tr) is introduced.
 		$\times$ denotes the gradient-stopping operation.
 		Q denotes the quantization operation.
 		Our framework can further support high-fidelity video decoding, by appending a detail rendering network (DR-Net).
 	 }
  	\vspace{-4mm}
 	\label{fig:framework}
 \end{figure*}
 
 \vspace{-3mm}
 \section{Approach}
 \vspace{3mm}
 \label{sec:method}
 \subsection{\textsf{Framework Overview}}
The proposed masked learning-powered video semantic coding framework is shown in Figure~\ref{fig:framework}.
 Let the original video $X$ and its lossy version $\tilde{X}$, which is compressed by a lossy video codec such as VVC.
 On the encoder side, we extract the semantic information from the original video and the lossy video, respectively.
 Then, the original video semantics is compressed with the aid of the lossy semantics.
 On the decoder side, we fuse the reconstructed semantic feature $\hat{S}$ and the lossy video $\tilde{X}$ as the video $\hat{X}$, which is deployed to support various machine analysis tasks.
 
 The framework is optimized via a non-semantics suppressed Masked Video Modeling (MVM) objective.
 The framework sub-components are detailed as follows.
 
 \textbf{Semantic Extraction Network (Sem-Net)}.
 To transform the videos from RGB space to semantic space, the original video $X$ and the lossy video $\tilde{X}$ are encoded as the semantic
 features $S$ and $\tilde{S}$, respectively.
 The tensor shape of $S$ and $\hat{S}$ are both $\mathbb{R}^{T \times 512 \times \frac{H}{32} \times \frac{W}{32}}$, where $T$ and $H \times W$ represent the temporal length and spatial dimensions of the input video.
 For producing $S$, we adopt the ResNet18~\cite{he2016deep} network as the Sem-Net, but replace its first Max-pooling layer with a stride two convolution layer for retaining more information.
 For producing $\tilde{S}$, we adopt a more lightweight network denoted Sem-Net$_{\mathrm{s}}$, since it has to be performed twice on both the encoder and decoder sides.
 Sem-Net$_{\mathrm{s}}$ mainly consists of five convolution layers of stride size two and kernel size three.
 We also insert depth-wise convolutions of large kernel size 7$\times$7 to effectively enlarge the receptive field size, without substantially increasing the computational cost.
 The weights of the above two networks are randomly initialized.
 
 \textbf{Semantic Coding}.
 In the basic model SMC, the residual semantic feature $Res$ between the original video feature $S^t$ and lossy one $\tilde{S}^t$ will be computed by a simple subtraction operation, and then compressed with an auto encoder-decoder CNN network, as shown in Figure~\ref{fig:framework} (c).
  Adding back the reconstructed residual semantic feature to $\tilde{S}^t$, we produce the compensated semantic feature $\hat{S}^t$.
 Both the encoder and the decoder networks, \textit{i.e.}, Sem-Enc and Sem-Dec, are composed of three residual blocks~\cite{he2016deep}.

 The above residue-based compression strategy only exploits the simple local redundancies, neglecting both long-range spatial redundancy and temporal redundancy.
 In the improved SMC++ model, we employ the self-attention mechanism to model the spatial-temporal redundancy among current semantics $S^t$, previous semantics $S^{t-1}$, current lossy semantics $\tilde{S}^t$ and previous lossy semantics $\tilde{S}^{t-1}$, where lossy semantics is extracted from the lossy video compressed by the base VVC layer.
 However, it is not trivial to directly model the redundancy among these highly heterogeneous features.
 For example, the features $S^t$ and $\tilde{S}^{t-1}$ of different timestamps are not spatially aligned.
The features $S^t$ and $\tilde{S}^t$ extracted from original and lossy videos exhibit domain gap.
To address this issue, we introduce the blueprint semantic feature to align the above features in terms of both spatial arrangement and domain.
The aligned features are then compressed by a pair of encoding and decoding Transformers, as shown in Figure~\ref{fig:framework} (d).

 \textbf{Semantic-Visual Information Fusion}.
 After obtaining the reconstructed semantic feature $\hat{S}$ and the lossy video $\tilde{X}$ on the decoder side,
 we use a UNet-style generator network termed F-Net to synthesize the final video $\hat{X}$,
which will be consumed by downstream AI task models.
 Considering that global structures instead of local details are more critical to video semantics, the above feature fusion procedures within the UNet are conducted in four-, eight-, and sixteen-times downsampled feature spaces.
 
\textbf{High-Fidelity Decoding Support}.
In practical applications, videos serve a dual purpose: machine task support and human viewing. To ensure our framework can decode high-fidelity details suitable for human inspection, we further append it with a lightweight Detail Rendering Network (DR-Net).
The decoded high-fidelity videos also allow for the manual review of AI analysis results.

\revise{ 
\textbf{Discussion with the Scalable Coding Approaches~\cite{liu2021semantics}\cite{yan2021sssic}}.
These approaches~\cite{liu2021semantics,yan2021sssic} use the base layer to deliver semantic information, whereas our approach uses the enhancement layer for this purpose.
This provides the following benefits:
First, the base-layer semantic information in~\cite{liu2021semantics,yan2021sssic} is task-specific. This contradicts the core principle of our framework: ``A single model that generates generalizable bitstream for various tasks''.
In contrast, we utilize a traditional codec as the base layer, which is task-agnostic and not bound to specific tasks.
Second, the bitstream produced by our method is partially compatible with traditional codec decoders. This compatibility allows the video stream to meet diverse user requirements. For example, users with low-end devices incapable of running deep learning models can still watch video, using the traditional codec for decoding.
}
 
 \subsection{\textsf{Semantic Learning Objective}}
 In this section, we describe how to learn compressible semantic representation from unlabeled videos, which is one of the core challenges for the unsupervised semantic coding problem.
 The learning objective is based on the MAE framework~\cite{he2022masked}, inspired by the fact that MAE learns strong and generalizable semantic representation from unlabeled image/video data.
 Furthermore,  to improve the semantic coding efficiency,
 \textit{i.e.}, excluding semantic-less information for saving bitcost, we introduce a Non-Semantics Suppressed (NSS) learning strategy to better adapt the vanilla MAE objective to the coding problem.

 \textbf{MAE Learning.}
 Given the decoded video $\hat{X}$, we first divide it into regular non-overlapping patches of size $16\times16$, and each patch is transformed to tokens by linear embedding.
 Then, a proportion of tokens are randomly masked, and the remaining unmasked token set ${Tok}$ is fed into the prediction network $\phi$, which includes an encoder and a decoder, for reconstructing the video.
 The pixel-wise reconstruction loss of the MAE task is given by,
 \begin{equation}	
 	\label{eq_mae_pixel}
 \mathcal{L}_{MAE-pixel} = \frac{1}{\mathcal{M}} \sum_{i \in \mathcal{M}}^{} | \phi (Tok)[i] - {X}[i] |,
 \end{equation}
 where $i$ is the masked token index, $\mathcal{M}$ is the index set of masked tokens, and $X$ is the ground-truth video, $|\cdot|$ denotes the $\ell_1$ distance function. We adopt the $\ell_1$ instead of the patch-normalized $\ell_2$~\cite{he2022masked}\cite{tong2022videomae}, as we prefer the stable gradient values \textbf{1} of $\ell_1$.
 This stabilizes the joint training with other loss functions in our system, such as GAN loss.

 Furthermore, considering that temporal dynamics, \textit{e.g.}, motion information, is another crucial component of videos, we also extend the MAE learning objective from the above pixel-only one to a motion-enhanced one,
 \begin{equation}
  	\label{eq_mae_motion}
 \mathcal{L}_{MAE-motion} = \frac{1}{\mathcal{M}} \sum_{i \in \mathcal{M}}^{} | \phi_{m}(Tok)[i] - \mathcal{T} \cdot {X}[i] |, 
 \end{equation}
where $\phi_{m}$ shares the parameters of the encoder with $\phi$ but includes a motion decoder that is not shared.
\revise{Compared to the appearance loss in the vanilla MAE framework, the motion loss ensures that motion pattern information is also preserved in the token space. This enhancement benefits a wide range of video-related tasks by capturing both spatial and temporal dynamics effectively.}
$\mathcal{T}$ denotes the motion calculation function, such as the RGB difference map and the optical flow map.
In the experimental section, we extensively compare these design choices.

Guided by the above objectives, the shared encoder implicitly clusters the video patches into some semantic centers, and the decoder builds a spatial-temporal reason graph among these semantic primitives to predict the pixels or motions of the masked region.
This facilitates preserving semantics-relevant information within each patch of the compressed video, as well as interactions among different patches.
However, these detail-reconstruction objectives also encourage $\hat{X}$ over-memorizing some non-semantic information, such as the object texture details, which degrades the quality of the learned semantic representation, as well as decreases the compression efficiency.
 
 \textbf{Non-Semantics Suppressed (NSS) Semantic Learning.}
 To suppress the non-semantic information leaked from $X$ to the compressed video $\hat{X}$, we explicitly regularize the information entropy of the tokens $Tok$ in MAE feature space, conditioned on the mined semantic feature $S$.
 However, due to the difficulty of estimating the entropy of a continuous variable, we insert a quantization operation in the tail of the tokenization procedure, so that $Tok$ is a discrete variable.
 Then, we use a Gaussian Mixture Model (GMM)~\cite{reynolds2009gaussian} with component number $K$ to approximate its distribution.
 The distribution of each token ${Tok}(i)$ is defined by the dynamic mixture weights $w_{i}$, means $\mu_{i}$ and log variances $\sigma_{i}$, which are produced by a density parameter estimation network (Density-Net).
 With these parameters, the distributions can be determined as,
 \begin{equation}
 	{p}({Tok}[i] | {S})  \sim \sum_{k=1}^{K} w_{i}^k \cdot \mathcal{N}  (\mu_{i}^k, e^{\sigma_{i}^k}).
 	\label{eq:gmm_dist}
 \end{equation}
 
 Then, the discretized likelihoods of each video patch token can be given by,
 \begin{equation}
 	p(Tok[i]|{S}) = c({Tok}[i] + 0.5) - c(Tok[i] - 0.5),
 	\label{eq:entropy_estimate}
 \end{equation}
 where $c(\cdot)$ is the cumulative function~\cite{papoulis2002probability} of the GMM in Equation~\eqref{eq:gmm_dist}.
 Finally, the non-semantics suppressed mask learning objective can be given by,
 \begin{equation}
 	\mathcal{L}_{Sem} = -\beta \frac{1}{\mathcal{M}} \sum_{i \in \mathcal{M}}^{}\log(p({Tok[i]} \mid {S})) + \mathcal{L}_{MAE},
 	\label{eq:nss}
 \end{equation}
\revise{where $\mathcal{M}$ denotes the index set of masked tokens}. $\beta$ is the balancing weight.

 In our basic SMC model, $\mathcal{L}_{MAE}$ is pixel-based reconstruction loss $\mathcal{L}_{MAE-pixel}$ for simplicity.
 In our improved SMC++ model, $\mathcal{L}_{MAE}$ incorporates both the pixel- and the motion-based reconstruction objectives, \textit{i.e.},
 $\mathcal{L}_{MAE} = \frac{1}{2} \mathcal{L}_{MAE-pixel} + \frac{1}{2} \mathcal{L}_{MAE-motion}$, aiming to learn enhanced spatial-temporal representation capability.

 \textbf{Discussion.}
 Although Equation~\eqref{eq:entropy_estimate} shares a similar format with the bit estimation procedure of the hyper-prior-based image compression methods~\cite{balle2018variational}\cite{cheng2020learned},
 our goal is fundamentally different from theirs.
 Our approach aims to suppress the extra non-semantic information that is introduced by the MAE task, transporting zero bits,
 while the method~\cite{balle2018variational} explores the hierarchical redundancies within images, and the estimated bits are additionally transported.

 \textbf{Density-Net.}
 It consists of two stacked convolutions of kernel size three, aligning the semantic feature ${S}$ to the MAE feature space.
 Then, we append a multiple layer perceptron (MLP) to predict $w$, $\mu$, and $\sigma$ within Equation~\eqref{eq:gmm_dist}.
 The gradient of ${S}$ is detached during the back-propagation procedure, forming a self-bootstrapping paradigm and avoiding the collapsing solution~\cite{chen2021exploring}.
 NSS scheme enforces the masked learning procedure to focus on mining semantic information, while the optimized semantic-rich ${S}$ leads to a more principled NSS objective.
 
\subsection{{Overall Optimization Target}}
\label{method_learning}
The overall optimization target is to enforce the decoded video $\hat{X}$ of rich semantics and good visual quality, while minimizing the bitrate of the transported semantic feature. The whole loss function can be given by,
\begin{equation}
	\mathcal{L} =   \alpha \mathcal{L}_{Sem}+ 
	\mathcal{L}_{lpips}  +  \mathcal{L}_{GAN}
	+ H,
\end{equation}
where
$\alpha$ is the balancing weight.
Following~\cite{esser2021taming}, we introduce the combined $\mathcal{L}_{lpips}  +  \mathcal{L}_{GAN}$ item to regularize the visual quality of the compressed video, where $\mathcal{L}_{lpips}$ denotes the perceptual loss~\cite{zhang2018unreasonable}, and $\mathcal{L}_{GAN}$ denotes the GAN loss.
$H$ represents the semantic feature bitrate, which is estimated with the factorized bitrate estimation model~\cite{balle2016end}.
 
 \begin{figure*}[!thbp]
	\centering
	\includegraphics[width=16.5cm]{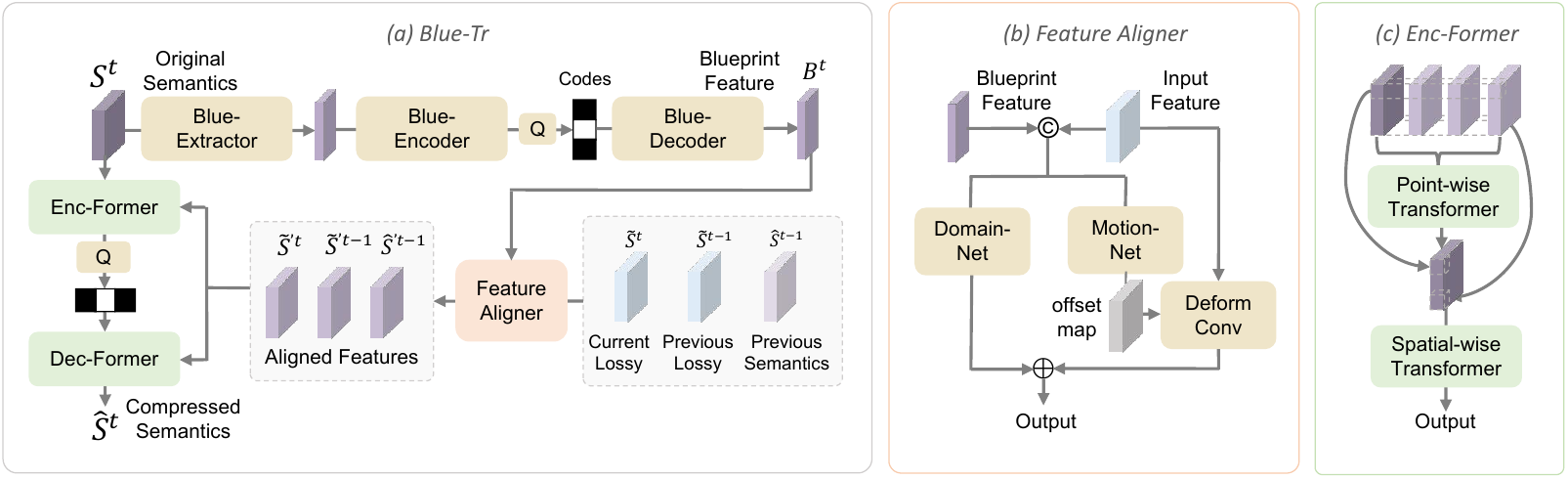}
	\caption {
		Blueprint-guided compression Transformer (Blue-Tr) first extracts a blueprint semantic feature, which is then employed to align diverse available features. Finally, a decomposed Transformer compresses redundancies among the current frame's semantic feature and the aligned features.
		The feature aligner module (b) align different features by using the blueprint feature as the guidance.
		Q denotes the quantization operation.
		The current lossy semantics $\tilde{S}^t$ is also fed into the Blue-Encoder/Decoder for aiding the compression of the blueprint feature.
		We omit this in the figure for simplicity.
	}
	\vspace{-4mm}
	\label{fig:group_entropy_transformer}
\end{figure*}
 
 \revise{ 
\subsection{Theoretical Justification of the NSS Objective}
\label{sec:proof}
\textbf{{Notation.}} To begin, we introduce the notations used throughout this section.
We refer to video patch token features and the semantic features, of all videos, as the random variables $\mathbf{{TOK}}$ and $\mathbf{{S}}$. The true conditional distribution of $\mathbf{TOK}$ given $\mathbf{S}$ is denoted as $P_r(\mathbf{TOK} \mid \mathbf{S})$, which will be simplified to $P_r$ in the following for clarity. The distribution estimated by Density-Net is represented as $P(\mathbf{TOK} \mid \mathbf{S})$.
Let $\mathcal{C}$ be the discrete representation of $\mathbf{S}$ obtained via a clustering algorithm such as KMeans~\cite{ahmed2020k}.

\textbf{{Lemma 1: Upper Bound on Conditional Entropy.}}
To establish the theoretical basis, we begin by analyzing the KL divergence between $P_r$ and $P(\mathbf{TOK} \mid \mathbf{S})$,
\begin{equation} \label{eq:KL}
	KL\Bigl(P_r \,\|\, P\Bigr) = \mathbb{E}_{P_r}\!\left[\log\frac{P_r(\mathbf{TOK} \mid \mathbf{S})}{P(\mathbf{TOK} \mid \mathbf{S})}\right] \geq 0.
\end{equation}
Expanding the logarithm yields,
\begin{equation}
	\mathbb{E}_{P_r}\!\left[\log P_r(\mathbf{TOK} \mid \mathbf{S}) - \log P(\mathbf{TOK} \mid \mathbf{S})\right] \ge 0,
\end{equation}
then,
\begin{equation}
	\label{eq:cond_entropy_upper}
	-\mathbb{E}_{P_r}\,\log P(\mathbf{TOK} \mid \mathbf{S}) \ge -\mathbb{E}_{P_r}\,\log P_r\big(\mathbf{TOK} | \mathbf{S}\big),
\end{equation}
where 
\begin{equation}
		\label{eq:cond_entropy}
	-\mathbb{E}_{P_r}\!\left[\log P_r \big(\mathbf{TOK} | \mathbf{S}\big) \right] = H\big(\mathbf{TOK} | \mathbf{S}\big),
\end{equation}
is the conditional Shannon entropy.

\textbf{{Lemma 2: Entropy Decomposition.} }
Next, we decompose the total entropy of $\mathbf{TOK}$ to better understand the relationships between conditional entropy and mutual information. According to the entropy decomposition rule~\cite{cover1999elements}:
\begin{equation} \label{eq:decomposition}
	H\big(\mathbf{TOK}\big)=H\big(\mathbf{TOK}\mid \mathbf{S}\big)+I\big(\mathbf{TOK};\mathbf{S}\big),
\end{equation}
where $I\big(\mathbf{TOK};\mathbf{S}\big)$ denotes the mutual information between $\mathbf{TOK}$ and $\mathbf{S}$.  
Since entropy terms evolve during training, we can rewrite Equation~\eqref{eq:decomposition} as:
\begin{equation} \label{eq:entropy_delta}
	\Delta H\big(\mathbf{TOK}\big)= \Delta H\big(\mathbf{TOK} | \mathbf{S}\big) + \Delta I\big(\mathbf{TOK};\mathbf{S}\big).
\end{equation}

\textbf{{Lemma 3: Information Equivalence During Feature Clustering.}}  
Assuming the clustering algorithm uses a sufficiently large centroid number and that $\mathbf{S}$ is highly compressible (since our learning objective minimizes its bitrate), the clustered feature $\mathcal{C}$ retains nearly all the information from $\mathbf{S}$,
\begin{equation}
H\big(\mathbf{S} \mid \mathcal{C}\big) \approx 0.
\end{equation}  
The chain rule~\cite{cover1991entropy} for mutual information yields
\begin{equation}\label{eq:chain_eq1}
I\big(\mathbf{TOK}; \mathbf{S},\mathcal{C}\big) = I\big(\mathbf{TOK};\mathcal{C}\big) + I\big(\mathbf{TOK}; \mathbf{S}\mid\mathcal{C}\big),
\end{equation}  
\begin{equation}\label{eq:chain_eq2}
I\big(\mathbf{TOK}; \mathbf{S},\mathcal{C}\big) = I\big(\mathbf{TOK}; \mathbf{S}\big) + I\big(\mathbf{TOK}; \mathcal{C}\mid\mathbf{S}\big).
\end{equation}  
Equating Equations~\eqref{eq:chain_eq1} and ~\eqref{eq:chain_eq2} gives,
\begin{equation} \label{eq:chain_rule}
	\begin{aligned}
	&I\big(\mathbf{TOK};\mathbf{S}\big) - I\big(\mathbf{TOK};\mathcal{C}\big) \\
	& = \Bigl[ I\big(\mathbf{TOK}; \mathbf{S}\mid\mathcal{C}\big) - I\big(\mathbf{TOK}; \mathcal{C}\mid\mathbf{S}\big) \Bigr].
 	\end{aligned}
\end{equation}
According to the entropy decomposition rule~\cite{cover1999elements} and the non-negativity of the entropy, we have,
\begin{equation} \label{eq:tok_s_c}
0 \le I\big(\mathbf{TOK}; \mathbf{S}\mid\mathcal{C}\big) \leq H\big(\mathbf{S}\mid\mathcal{C}\big) \approx 0.
\end{equation}
Since $\mathcal{C}$ is derived from $\mathbf{S}$, we also have,
\begin{equation}\label{eq:tok_c_s}
I\big(\mathbf{TOK}; \mathcal{C}\mid\mathbf{S}\big) = 0.
\end{equation}
Combining Equation~\eqref{eq:chain_rule}, \eqref{eq:tok_s_c}, and \eqref{eq:tok_c_s}, we obtain
\begin{equation} 
\begin{aligned}
&\left| I\big(\mathbf{TOK};\mathbf{S}\big) - I\big(\mathbf{TOK};\mathcal{C}\big) \right| \\
& =\left|   I\big(\mathbf{TOK}; \mathbf{S}\mid\mathcal{C}\big) - I\big(\mathbf{TOK}; \mathcal{C}\mid\mathbf{S}\big) \right| \approx 0.
\end{aligned}
\end{equation}

This indicates that the mutual information between \(\mathbf{TOK}\) and \(\mathcal{C}\) is very close to that between \(\mathbf{TOK}\) and \(\mathbf{S}\),  
\begin{equation}
	I\big(\mathbf{TOK}; \mathbf{S}\big) \approx I\big(\mathbf{TOK}; \mathcal{C}\big).
	\label{eq:approx_MI}
\end{equation}

\textbf{{Theorem 1: NSS Loss Increases the Semantic Separability.}}
Now, building upon the previous lemmas, we demonstrate how the NSS loss enhances the video token semantic separability. Recall that the NSS loss in Equation~\eqref{eq:nss} can be reformulated as,
\begin{equation}
	\mathcal{L}_{Sem} = \beta \underbrace{\big[	-\mathbb{E}_{P_r}\,\log P(\mathbf{TOK} \mid \mathbf{S}) \big]}_{term~1} + \mathcal{L}_{MAE},
\end{equation}
where $\beta$ is a balancing parameter.
\begin{itemize}
	\item The term 1 is an upper bound on $H\big(\mathbf{TOK} \mid \mathbf{S}\big)$, from Equation~\eqref{eq:cond_entropy_upper} and~\eqref{eq:cond_entropy}. Minimizing it reduces $H\big(\mathbf{TOK} \mid \mathbf{S}\big)$, i.e., $\Delta H\big(\mathbf{TOK} \mid \mathbf{S}\big) < 0$.
	\item The term $\mathcal{L}_{MAE}$ encourages highly semantic token representations such as local concepts, intuitively reducing the overall entropy, i.e., $\Delta H\big(\mathbf{TOK}\big) < 0$.
\end{itemize}

By choosing appropriate $\beta$, $\Delta H\big(\mathbf{TOK} \mid \mathbf{S}\big)$ can be decreased more quickly than $\Delta H\big(\mathbf{TOK}\big)$, while keeping that $\mathcal{L}_{MAE}$ still converges.
From Equation \eqref{eq:entropy_delta}, this achieves,
\begin{equation}
	\label{eq:increase_I}
	\Delta I\big(\mathbf{TOK};\mathbf{S}\big)=\Delta H\big(\mathbf{TOK}\big)-\Delta H\big(\mathbf{TOK}\mid \mathbf{S}\big) > 0,
\end{equation}
With Equation~\eqref{eq:approx_MI},
we have,
\begin{equation}
	\label{eq:tok_c}
	\Delta I\bigl(\mathbf{TOK};\mathcal{C}\bigr) > 0,
\end{equation}
where $\mathcal{C}$ denotes discrete clustered label set.

Then, considering that the variational bounds~\cite{poole2019variational} of mutual information between the video patch token feature and the discrete label can be given as,
\begin{equation}
	\label{eq:mutual-info-bound}
	\begin{aligned}
		&I\big(\mathbf{TOK};\mathcal{C} \big) \\ 
		&\le \mathbb{E}\!\left[\frac{1}{N}\sum_{i=1}^{N} \log \frac{P_r\big(\mathcal{C}[i] \mid {Tok}[i]\big)}{\frac{1}{N-1}\sum_{k\ne i} P_r\big(\mathcal{C}[i] \mid {Tok}[k]\big)}\right], \\
	\end{aligned}
\end{equation}
where $\mathcal{C}[i]$ and $Tok[i]$  are at the same spatial position in the same video.
$N$ is the number of all video patch tokens within the training batch.

Combining Equations~\eqref{eq:tok_c} and~\eqref{eq:mutual-info-bound}, we conclude that the NSS objective increases the mutual information between the token and clustered semantic features, thereby enhancing the semantic separability of the token features.
}

 \begin{figure}[!tbp]
	\centering
	\tabcolsep=0mm
	\scalebox{0.99}{
		\begin{tabular}{cc}
			\includegraphics[width=0.24 \textwidth]{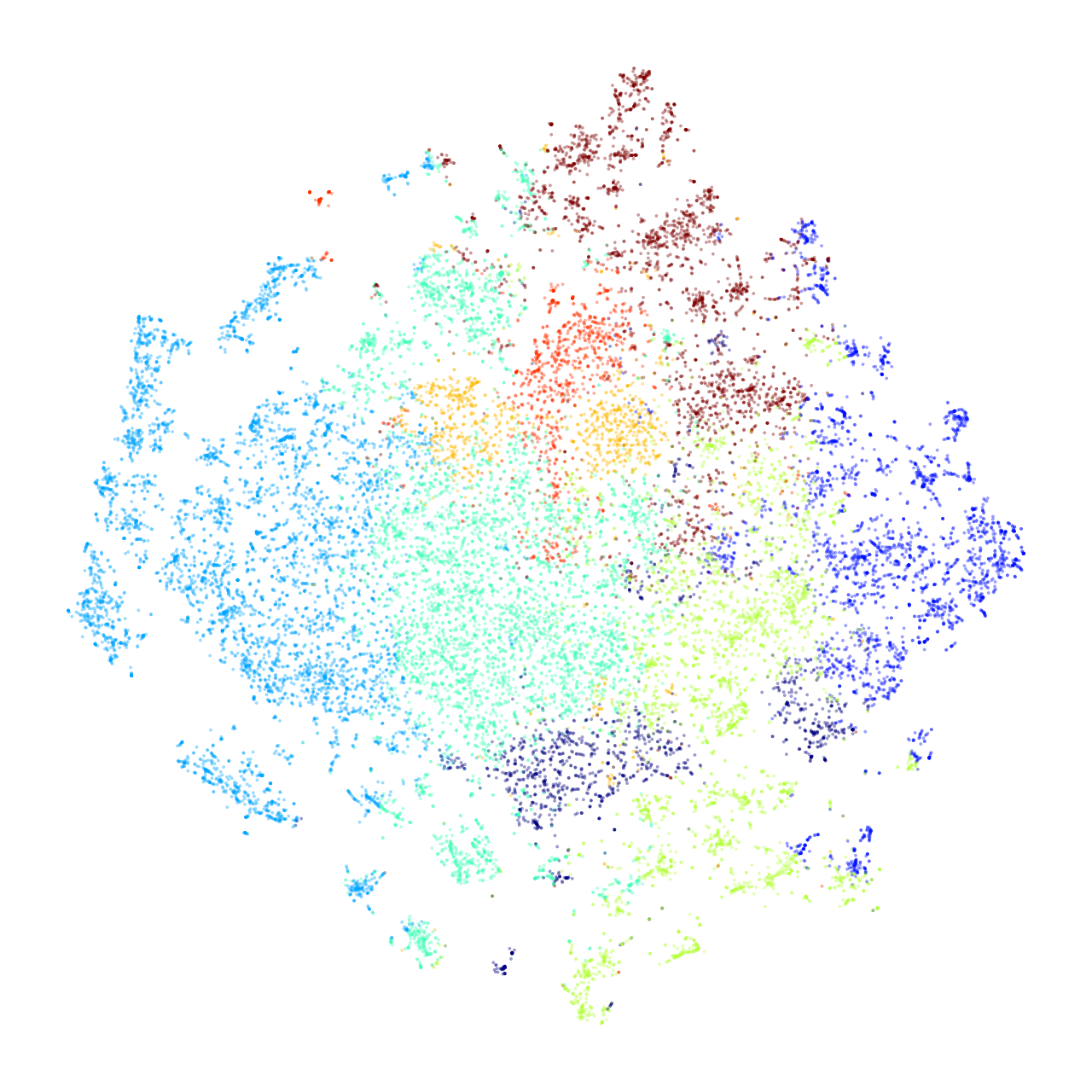}&
			\includegraphics[width=0.24 \textwidth]{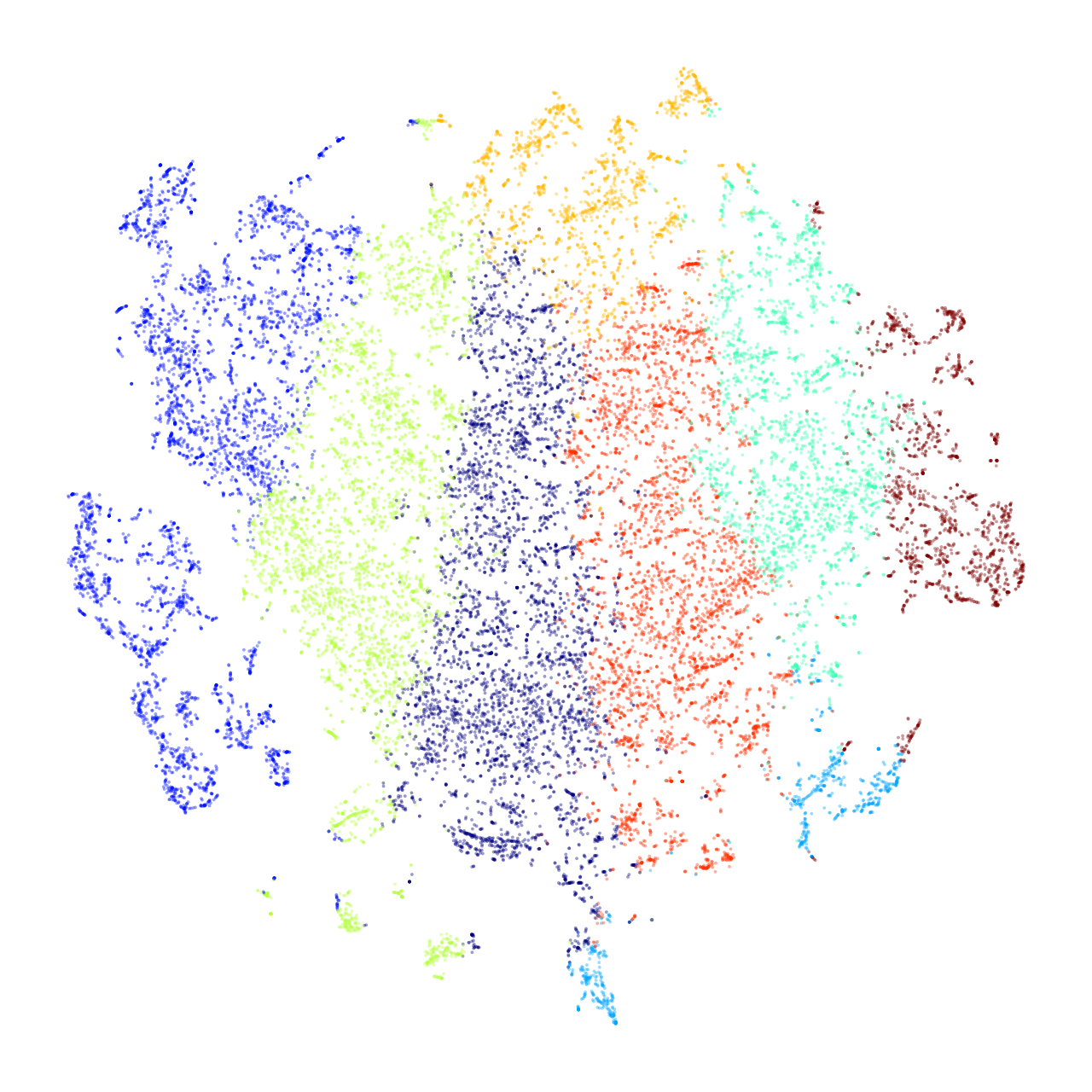} \\
			(a) & (b)	 
		\end{tabular}
	}
	\vspace{-2mm}
	\caption{
		Visualization of the patch features learned by different semantic objectives, \textit{i.e.},
		(a) vanilla MAE loss $\mathcal{L}_{MAE}$ and
		(b) our non-semantics suppressed MAE loss $\mathcal{L}_{Sem}$.
	}
	\vspace{-4mm}
	\label{fig:feature_tsne}
\end{figure}
\textbf{{Visualization.}}
To provide a more intuitive understanding how NSS loss improves the semantic boundaries, we further visualize the patch features.
First, we randomly select one hundred videos from the UCF101 dataset.
Then, we cluster the patch features of these videos into individual groups using the K-Means algorithm~\cite{hartigan1979algorithm}.
The cluster ID is then utilized as the pseudo-category for each patch. Subsequently, the features are visualized using the t-SNE technique~\cite{van2008visualizing}.
As compared in Figure~\ref{fig:feature_tsne}, although the vanilla MAE loss can learn roughly separable semantic representation, our non-semantics suppressed loss $\mathcal{L}_{Sem}$ learns much clearer semantic boundaries, indicating less semantic noises.
This also saves the semantic bitcost (0.0074bpp \textit{v.s.} 0.0028bpp for $\mathcal{L}_{MAE}$ and $\mathcal{L}_{Sem}$, respectively).

\subsection{{Blueprint-Guided Compression Transformer}}
In this section, we introduce another new component of the SMC++ model, 
\textit{i.e.}, Blueprint-guided compression Transformer (Blue-Tr), which effectively captures redundancy among heterogeneous features across different frames and coding layers. 
As illustrated in Figure~\ref{fig:group_entropy_transformer}, Blue-Tr operates under an ``align-then-compress" paradigm, guided by a novel blueprint semantic representation. The compression procedure consists of the following three steps:

\textbf{Blueprint Semantic Extraction and Compression.}
Drawing inspiration from coarse-to-fine compression approaches~\cite{hu2020coarse}\cite{hu2022coarse}, we extract the most salient video feature component, termed the blueprint semantic feature, as the coding guidance. Instead of downsampling the spatial dimension as in previous works, we condense the channel dimension for extracting the salient semantics, since semantic information is mostly encoded in the channels~\cite{chen2017sca}. Specifically, given the current frame's semantic feature $S^t$, we employ a Blue-extractor network to produce the blueprint feature with a channel dimension one-fourth that of $S^t$. This blueprint feature is then compressed using a pair of lightweight CNN-based compression networks, resulting in the compressed blueprint feature $B^t$, which is available at both the encoder and decoder to aid in compressing $S^t$.

The Blue-extractor network consists of three convolution layers with gradually decreasing channel number.
Its first two convolutions are followed by a residual block to enhance the non-linear transformation capability.
The encoder of the compression network is the same as the Blue-extractor network, while the decoder part is symmetric to the encoder.
Besides, $\tilde{S}^t$ is together fed into the encoder and decoder network to reduce the blueprint feature bitcost.

\textbf{Feature Alignment.}
Existing learnable video compression networks typically leverage motion information to compensate for the previous frame.
The produced frame is mostly aligned with the current frame, serving as an effective coding reference. However, our framework goes beyond just the previous frame, additionally involving the (current and previous) lossy frames from the base VVC coding layer.
Therefore, we use the blueprint feature $B^t$ to align all these features simultaneously in a unified manner.

As shown in Figure~\ref{fig:group_entropy_transformer} (b), the alignment operation includes two branches: one for motion compensation and another for domain correction. The motion branch estimates the motion between the input feature and $B^t$ using a Motion-Net and aligns their spatial arrangements via deformable convolution~\cite{dai2017deformable}. The domain branch uses a Domain-Net to correct the input feature domain. The features from both branches are subsequently fused.
Both Motion-Net and Domain-Net comprise three convolution layers equipped with the LeakyReLU~\cite{xu2015empirical} activation.

\textbf{Contextual Compression.}
Leveraging the aligned current lossy feature, aligned previous lossy feature, and aligned previous semantic feature as coding context, we compress the current semantic feature $S^t$ in an contextual coding manner~\cite{li2021deep}. During encoding, we feed these context features and $S^t$ together into an Enc-Former, generating a compressed feature. This feature is then quantized and encoded into bitcodes. On the decoder side, the compressed feature is fed into a Dec-Former together with context features, to reconstruct the semantic feature $\hat{S}^{t}$.

 Due to the previous alignment step, features at corresponding spatial locations exhibit high correlation. Consequently, the Enc-Former adopts a decomposed architecture: a point-wise Transformer eliminates redundancy across different features, followed by a spatial-wise Transformer to reduce spatial redundancy. This strategy significantly reduces the computational cost of the self-attention operation, approximately $\frac{1}{N^2}$ compared to the vanilla one, where $N=4$ is the type number of features fed into the Transformer. As the feature types expand, for example, by increasing the previous frame number, the cost reduction can be further improved. Dec-Former mirrors the Enc-Former's architecture.
Each Transformer includes three self-attention blocks~\cite{vaswani2017attention}.

\textbf{Discussion.}
Our approach inherits the strengths of two popular neural coding paradigms: motion-based explicit modeling approaches such as FVC~\cite{hu2021fvc}, and Transformer-based implicit modeling approaches like VCT~\cite{mentzer2022vct}. Similar to FVC, we warp feature maps and align features across timestamps. However, our Transformer-based approach excavates more intricate redundancies between features at different timestamps, positions, and coding layers.
Compared to VCT, we incorporate motion modeling to guide the Transformer, simplifying optimization and enhancing the model's robustness against large motion.
 \begin{figure*}[!thhp] 
	\centering
	\renewcommand\arraystretch{0.0}
	\newcommand{\widthscalefive}{0.24}
	\tabcolsep = 1.1mm
	\scalebox{0.95}{
		\begin{tabular}{cccc}
			\includegraphics[width=\widthscalefive \textwidth]{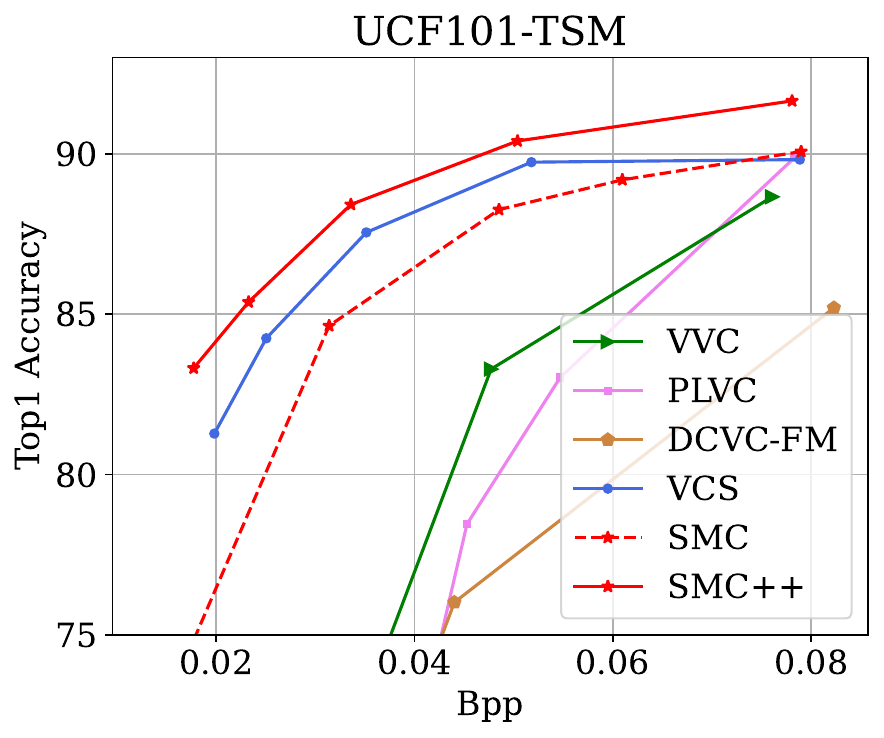} &
			\includegraphics[width=\widthscalefive \textwidth]{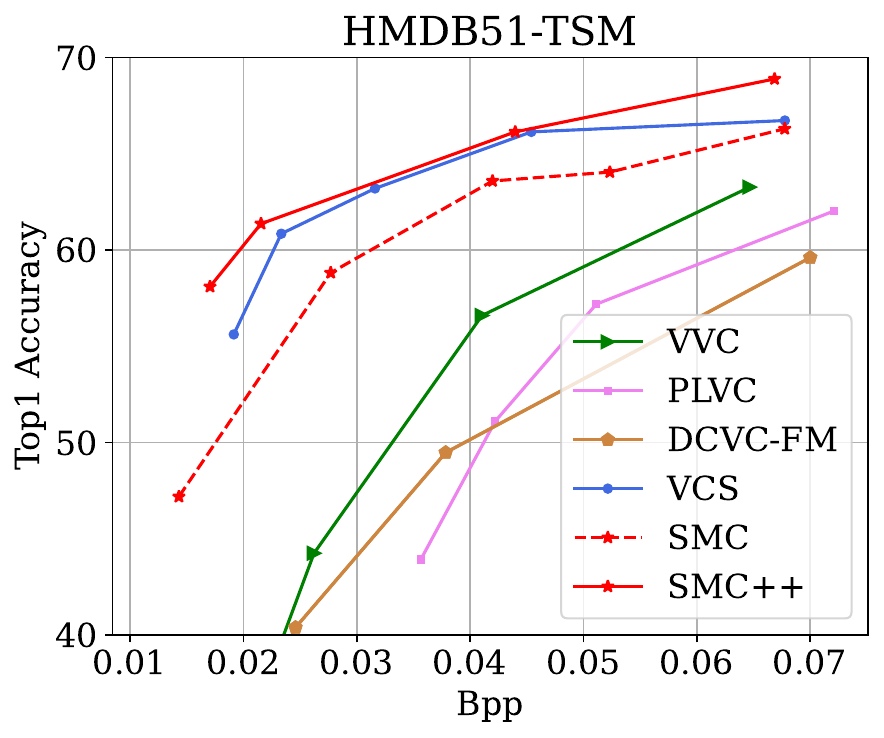}&
			\includegraphics[width=\widthscalefive \textwidth]{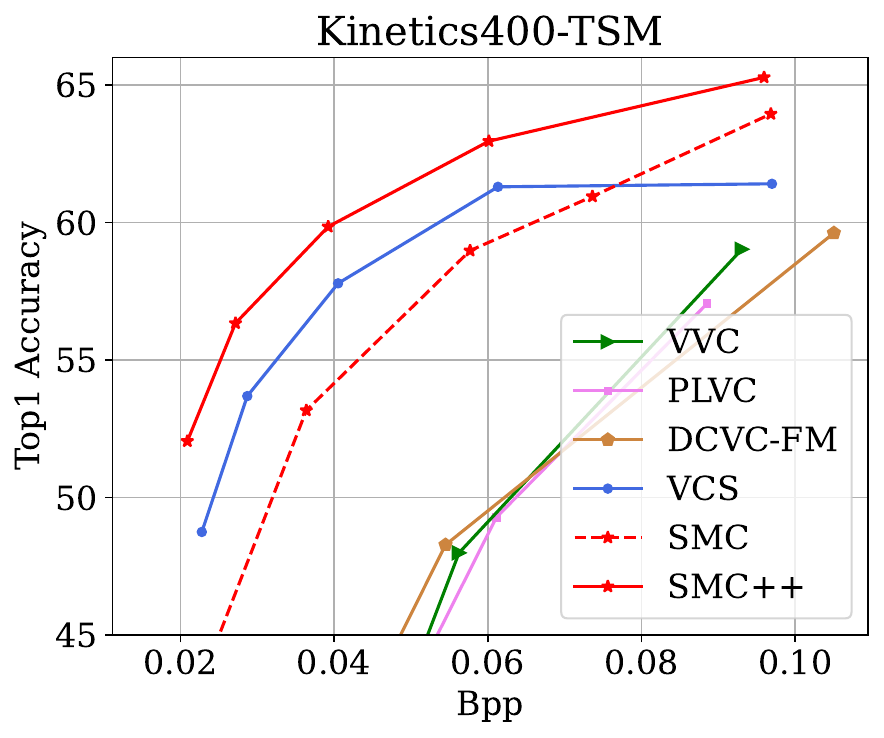} &
			\includegraphics[width=\widthscalefive \textwidth]{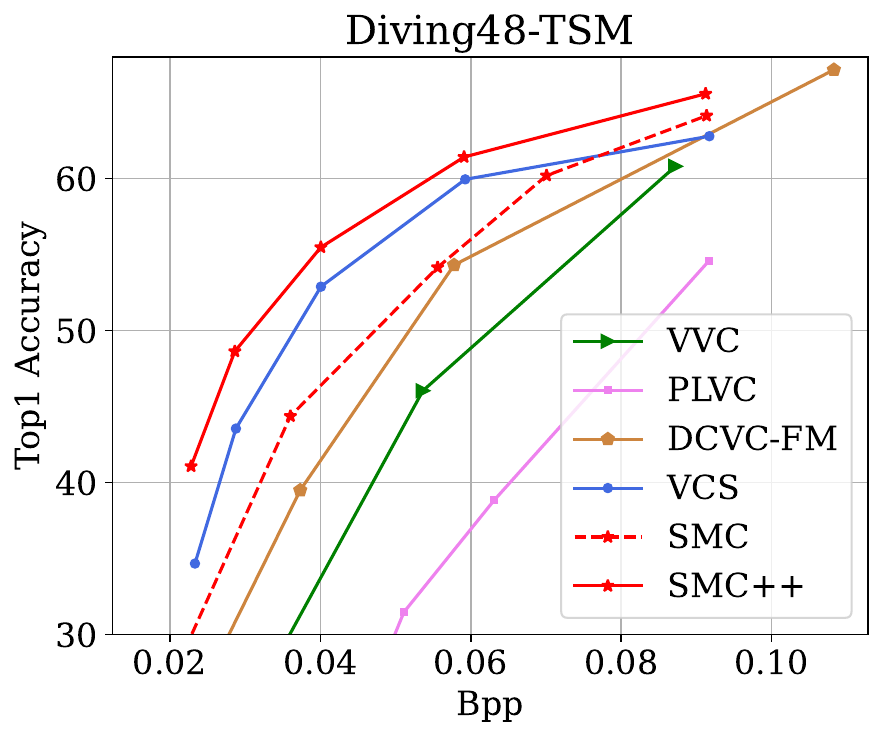} 
			\\
			\includegraphics[width=\widthscalefive \textwidth]{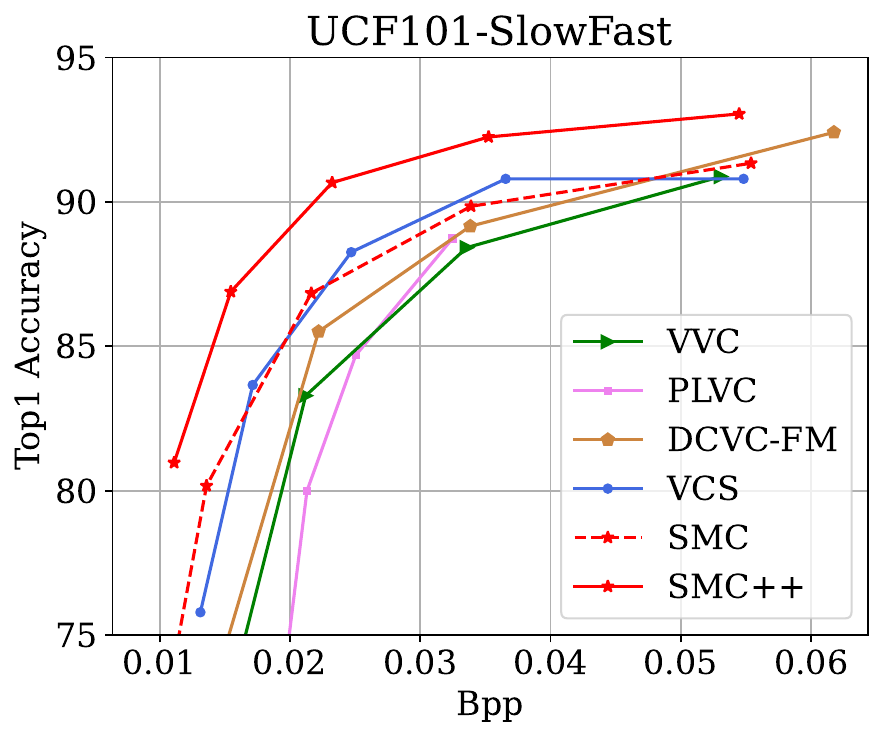}&
			\includegraphics[width=\widthscalefive \textwidth]{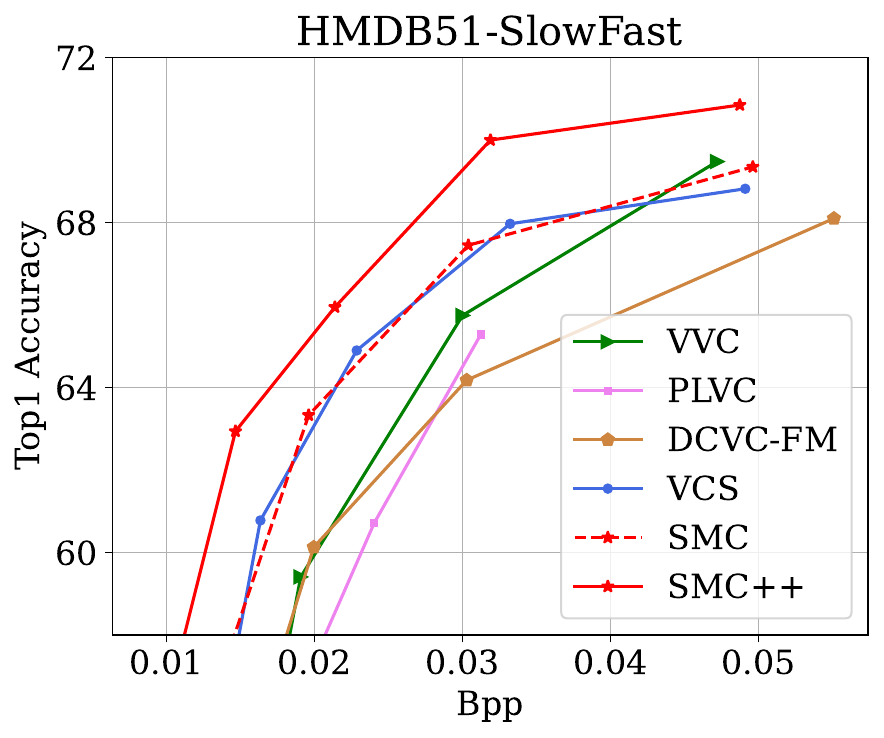} &
			\includegraphics[width=\widthscalefive \textwidth]{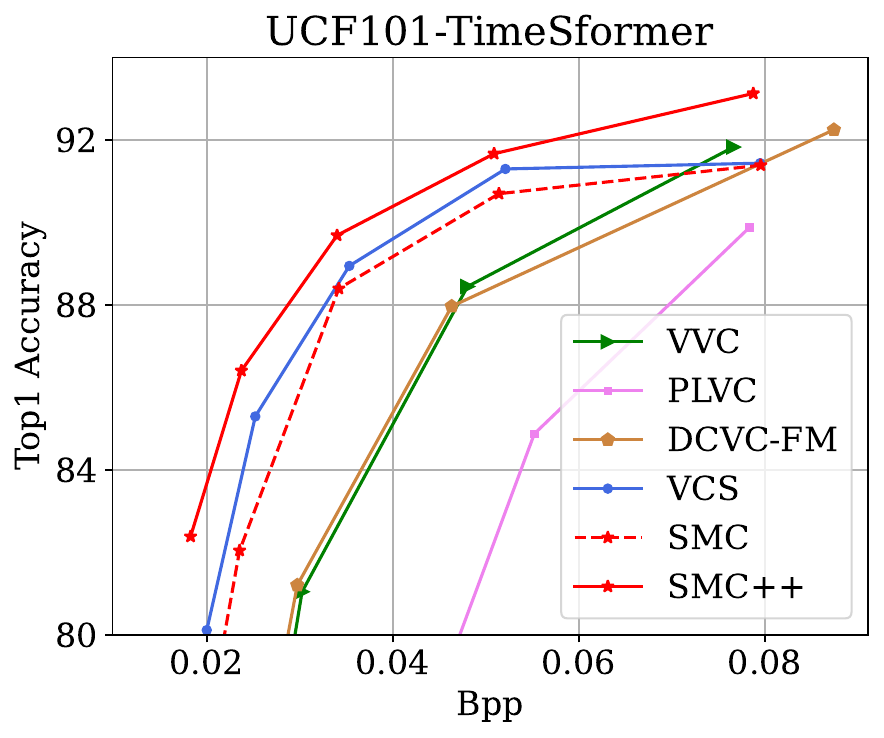}& 
			\includegraphics[width=\widthscalefive \textwidth]{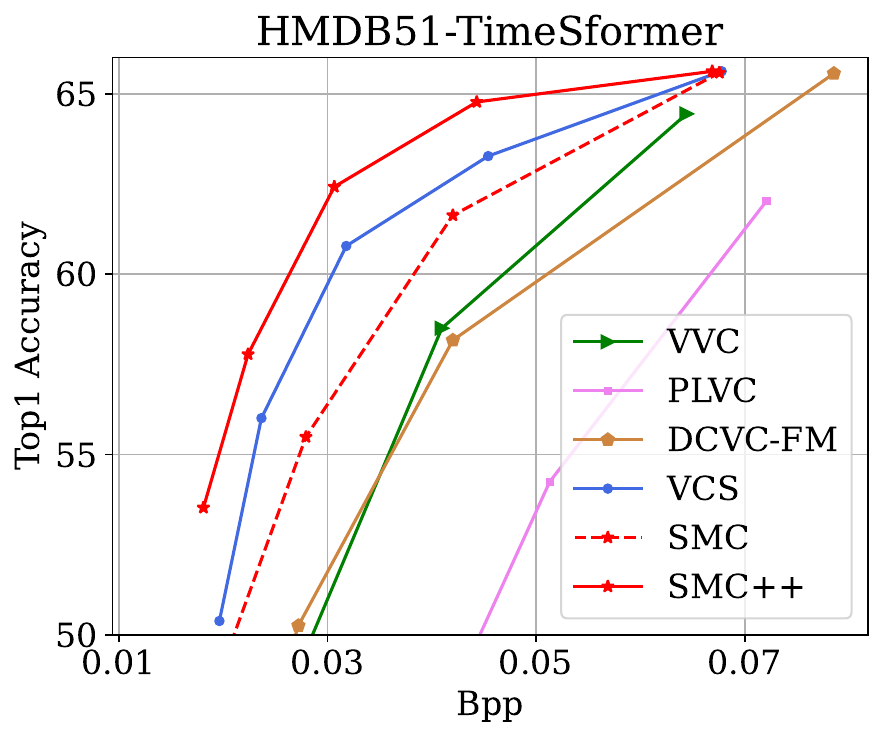}
			\\
			\includegraphics[width=\widthscalefive \textwidth]{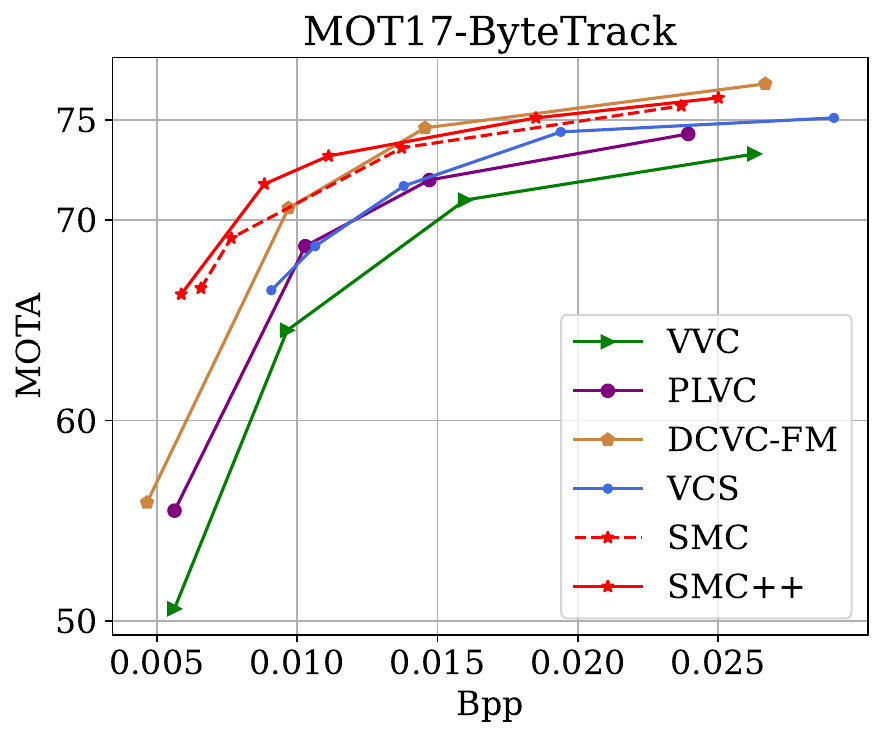} &
			\includegraphics[width=\widthscalefive \textwidth]{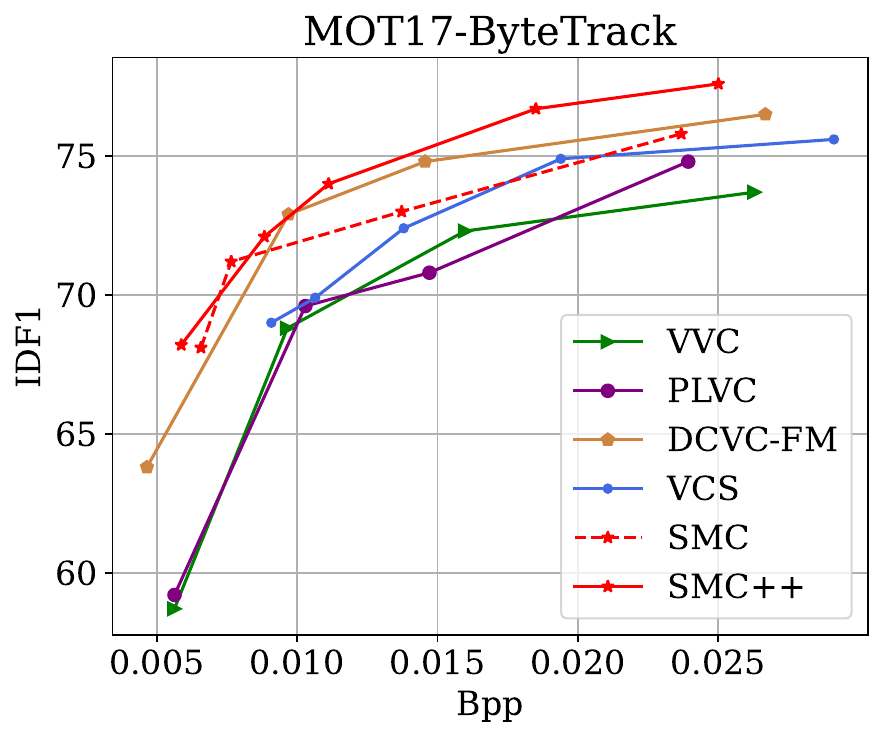} &
			\includegraphics[width=\widthscalefive \textwidth]{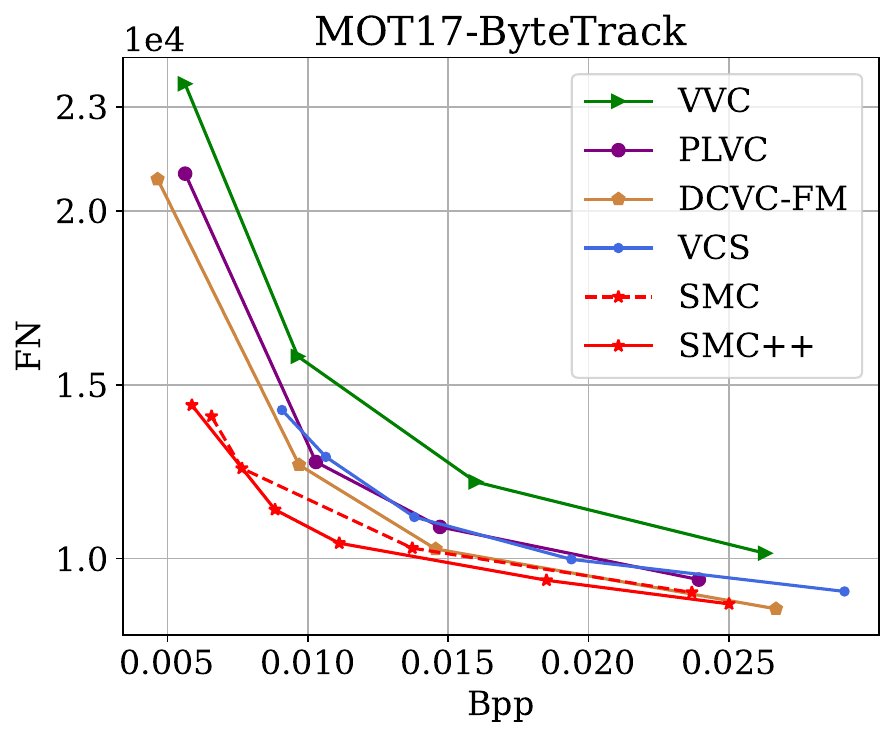} &
			\includegraphics[width=\widthscalefive \textwidth]{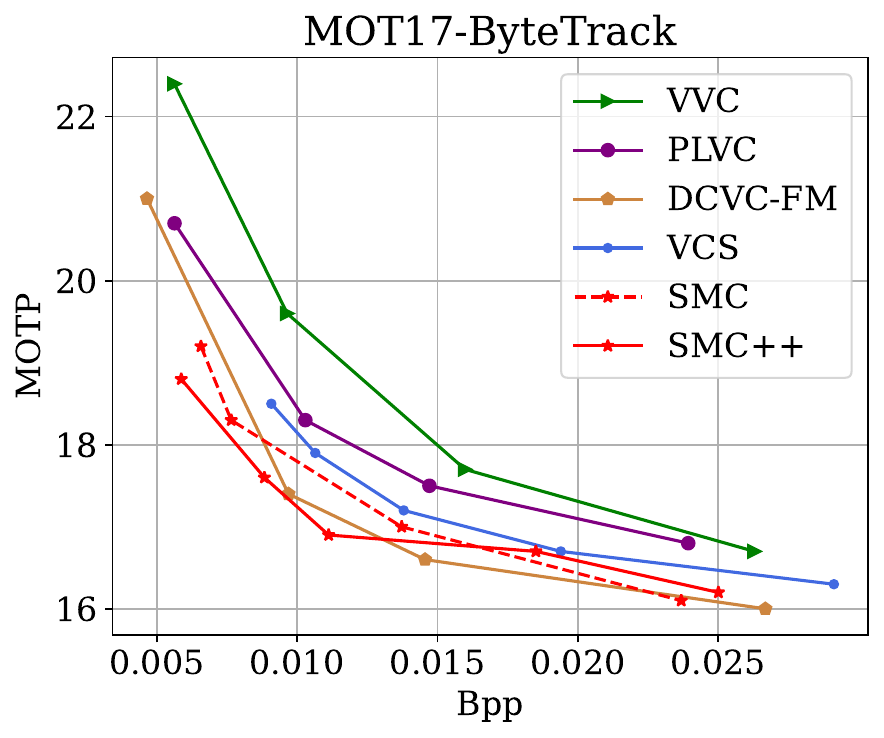} 
			\\
			\includegraphics[width=\widthscalefive \textwidth]{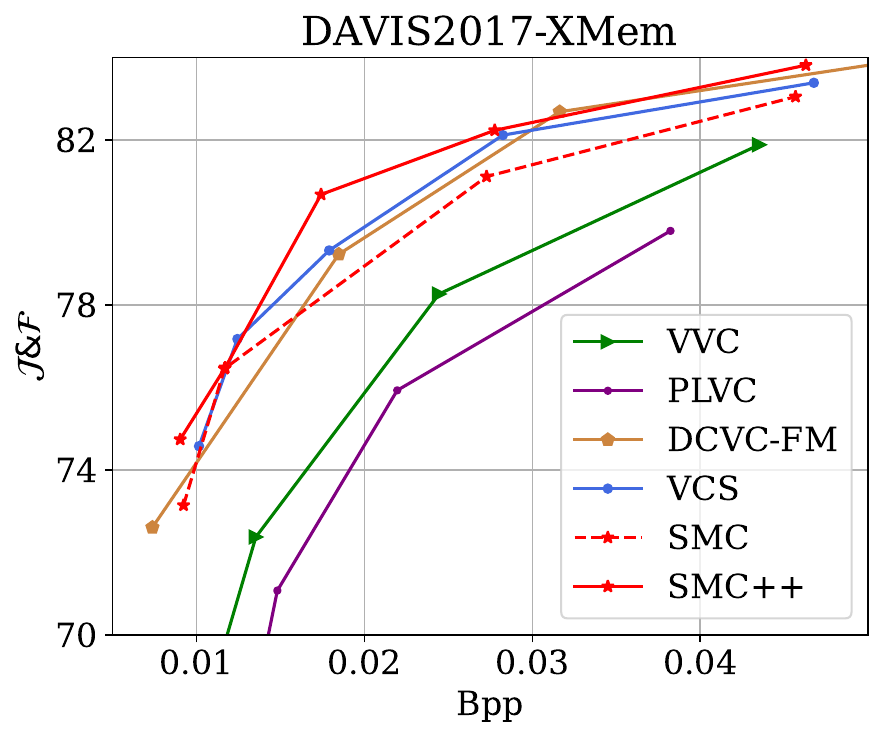} &
			\includegraphics[width=\widthscalefive \textwidth]{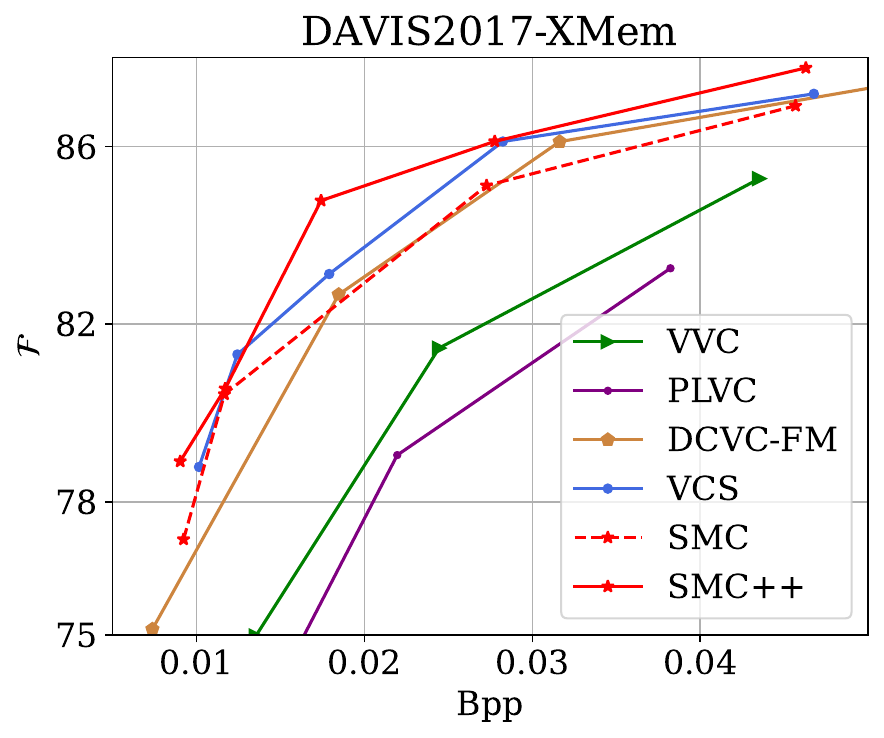} &
			\includegraphics[width=\widthscalefive \textwidth]{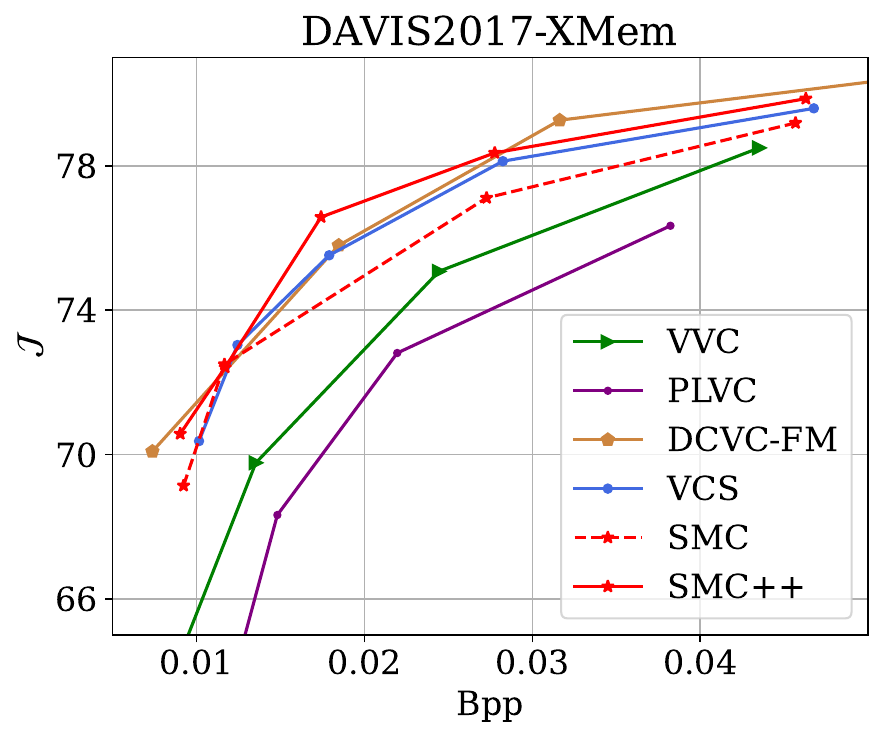} &
			\includegraphics[width=\widthscalefive \textwidth]{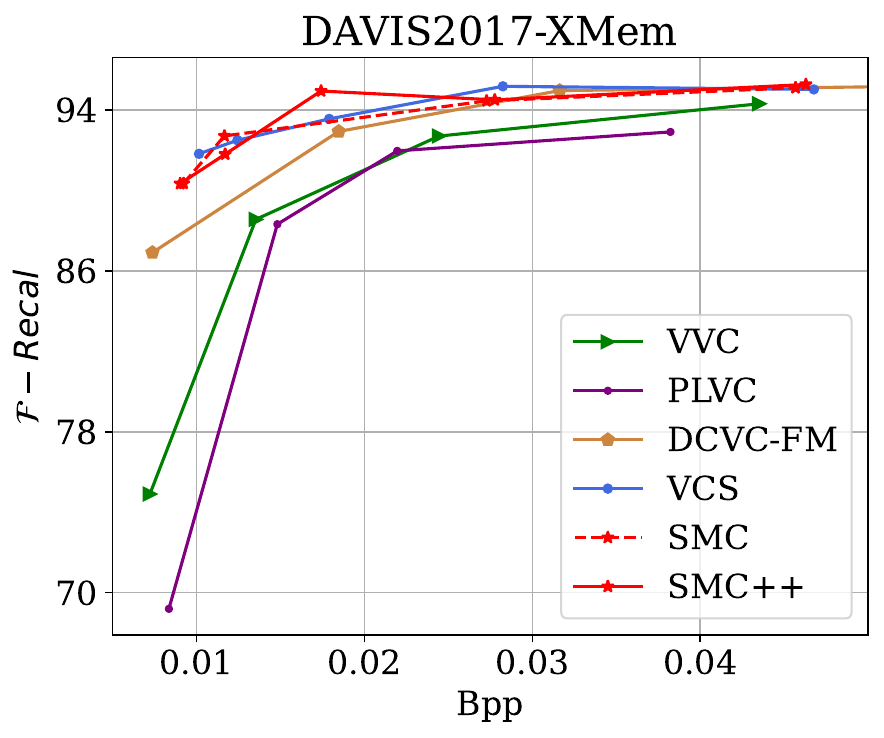}
		\end{tabular}
	}
	\vspace{-4mm}
	\caption{
		Semantic coding performance on Action Recognition (1st and 2nd rows), VOS (3rd row), and MOT (4th row) tasks.
		The plot titles are in \{Dataset\}-\{Task Model\} format. The codec setting is LDP mode with GOP size 10.
		The results on more advanced codec settings are provided in the supplementary material.
	}
	\vspace{-5mm}
	\label{fig:lbuv_mot_blind}
\end{figure*}
   
\subsection{{High-Fidelity Decoding Support}}
\label{sec:detail_render}
To serve human viewing, we introduce the Detail Rendering Network (DR-Net) to fuse the videos from our framework and the base VVC layer, yielding high-fidelity video output.
Note that this procedure does not involve transporting any new information,
since our framework is built upon the VVC codec, and decoding the VVC-compressed video is an intermediate step of our approach.

The fusion procedure comprises three sub-steps: 
\textit{1) Intra-semantic fusion:} Given the semantic-rich video from SMC++, an encoder network extracts features from each frame. The previous frame features are aligned to the current one with the Feature Aligner in Figure~\ref{fig:group_entropy_transformer} (b), producing the fused semantic feature.
\textit{2) Intra-lossy fusion:} This step is similar to the above procedure, except the input frames are from the VVC layer, producing the fused lossy feature.
\textit{3) Pixel decoding:} The above fused features are concatenated, and then fed into a decoder network to generate the final frames.

The encoder network comprises two stride-two convolution layers, succeeded by three residual blocks. The decoder network is symmetric with the encoder network.
All convolution layers have 64 channels. We optimize DR-Net using the $\ell1$ loss between the generated and original frames.
The parameters within SMC++ are frozen during the optimization, so that most inference costs can be shared for decoding semantic-rich videos and detail-rich videos.

 \vspace{-4mm}
 \section{Experiments}
 \label{sec:exp}
 \vspace{3mm}
\subsection{\textsf{Experimental Details}}
 \textbf{{Evaluation Datasets.}}
 For {action recognition} task, we evaluate on four large-scale video datasets, {UCF101}~\cite{soomro2012ucf101}, {HMDB51}~\cite{kuehne2011hmdb}, {Kinetics}~\cite{carreira2017quo}, and {Diving48}~\cite{li2018resound}.
 For {multiple object tracking (MOT)} task, we evaluate on {MOT17}~\cite{milan2016mot16}.
 For {video object segmentation (VOS)} task, we evaluate on {DAVIS2017}~\cite{pont20172017}.
 We also compare the compression performance on Standard HEVC Test Sequences~\cite{sullivan2012overview}, including HEVC Class B/C/D/E datasets.
 Dataset processing pipeline is provided in the supplementary material.

 \textbf{Experimental Setting.}
For the downstream tasks, we use TSM~\cite{lin2019tsm}, SlowFast~\cite{feichtenhofer2019slowfast}, and TimeSformer~\cite{bertasius2021space} for action recognition, ByteTrack~\cite{zhang2022bytetrack} for MOT, and XMem~\cite{cheng2022xmem} for VOS; for baseline codecs, HEVC, VVC~\cite{VVenC}, FVC~\cite{hu2021fvc}, PLVC~\cite{yang2021perceptual}, DCVC-FM~\cite{li2024neural}, and VCS~\cite{tian2022coding} are evaluated.
All codecs and our framework are evaluated on Low-Delay with P frame (LDP) coding mode.
The evaluation metrics include bpp, Top1 accuracy (action recognition), MOTA~\cite{kasturi2008framework}, MOTP, FN, and IDF1 (MOT), as well as the Jaccard index ($\mathcal{J}$), contour accuracy ($\mathcal{F}$), $\mathcal{J\&F}$, and $\mathcal{F}$-Recall (VOS).
For more details, please refer to the supplementary material.

\begin{table*}[!thbp]
	\centering
		\caption{
		Results on action recognition.
		``Original'' denotes the performance upper-bound, which is evaluated on the original dataset.
		The codec setting is LDP mode with GOP size 10.
	}
	\vspace{-2mm}
	\renewcommand\arraystretch{1.1}
	\tabcolsep=1.5mm
	
	\scalebox{0.99}{
		\begin{tabular}{|c|c|c|c|c|c|c|c|c|c|c|c|}
			\hline
			{}&\multicolumn{5}{c|}{\textbf{TSM Top1} (\%)}  &\multicolumn{3}{c|}{\textbf{Slowfast Top1} (\%)}&\multicolumn{3}{c|}{\textbf{TimeSformer Top1} (\%)} \\
			\hline
			{}  & 
			UCF101 & HMDB51 & Kinetics400 &                      Diving48 & Diving48~              & UCF101        & HMDB51          & HMDB51             &UCF101& UCF101 & HMDB51    \\
			\hline
			\textit{Bpp}&
			\textit{@0.04} & \textit{@0.04} &  \textit{@0.06} & \textit{@ 0.03} & \textit{@ 0.05~} &\textit{@0.03} & \textit{@0.02} & \textit{@0.03}&  \textit{@0.02} &  \textit{@0.04} & \textit{@0.04}   \\
			\arrayrulecolor{gray}\cdashline{1-12}[5pt/3pt]
			HEVC &  52.70 & 36.64 &  35.28 &  - & 22.48 &85.13 &47.78 &62.68 &47.16& 69.92 &37.45   \\
			\hline
			FVC & 64.61 &  47.54 &  37.23 & - &22.94 &79.43 & 52.15 & 59.81 &58.90& 72.47 & 45.11   \\
			\hline
			PLVC & 71.40 &  48.67 &  48.65 & - & 30.13  & 87.38 & 56.33 & 64.50 &38.92& 70.06 & 44.59\\
			\hline
			VVC & 76.97 &  55.72 & 49.11 & 24.38 & 42.75  &86.93 & 59.95 & 65.74 &66.98& 85.10  & 57.83\\
			\hline
			DCVC-FM & {72.71} &  {50.15} &  {49.51} &{32.16}  &{48.67} &{87.95} &{60.15}  &{64.06} &{69.50}&{85.39}   &{57.08}\\
			\hline
			VCS & {88.19} &  	{64.98} &  {61.07} &{44.57}  &{56.54} &{89.39} &{63.09}  &{67.01} &{80.11}&{89.60}   &{62.27}\\
			\hline
			SMC & {86.46} &  	{62.92} &  {59.26} &{37.79}  &{51.36} &{88.89} &{63.48}  &{67.30} &{77.61}&{89.17}   &{60.74}\\
			\hline
			SMC++ & \textbf{89.17} &  	\textbf{65.29} &  \textbf{62.93 } &\textbf{49.45}  &\textbf{58.58} &\textbf{91.56} &\textbf{65.33}  &\textbf{69.27} &\textbf{83.66}&\textbf{90.39}   &\textbf{64.02}\\
			\arrayrulecolor{gray}\cdashline{1-12}[5pt/3pt]
			\textcolor{gray}{Original} & \textcolor{gray}{93.97} &  \textcolor{gray}{72.81} &  \textcolor{gray}{70.73} & \textcolor{gray}{75.99}& \textcolor{gray}{75.99}
			& \textcolor{gray}{94.92} &  \textcolor{gray}{72.03} &  \textcolor{gray}{72.03} & \textcolor{gray}{95.43} & \textcolor{gray}{95.43} &  \textcolor{gray}{71.44}
			\\
			\hline
			
		\end{tabular}
	}
	 	\vspace{-4mm}
	\label{tab:benchmark_ar}
\end{table*}

\begin{table}[!tbp]
	\centering
	\caption{
		MOT performance comparison of different coding methods on MOT17.
		The codec setting is LDP mode with GOP size 10.
		``Original'' denotes the results with original videos, \textit{i.e.}, the performance upper bound.
	}
	\vspace{-2mm}
	\tabcolsep=0.3mm
	\scalebox{0.99}{
		\begin{tabular}{|c|c|c|c|c|c|c|}
		\hline
		Bpp&\multicolumn{2}{c|}{@0.01bpp} & \multicolumn{2}{c|}{@0.015bpp} & \multicolumn{2}{c|}{@0.02bpp} \\
			\hline
		Metric	&{\makecell{MOTA(\%)}}  	& {\makecell{IDF1(\%)}} &{\makecell{MOTA(\%)}}  	& {\makecell{IDF1(\%)}}&{\makecell{MOTA(\%)}}  	& {\makecell{IDF1(\%)}}\\
			\hline
			{{HEVC}}&61.30 &64.32 & 68.61 &69.62 & 71.90 &72.18 \\
			\hline
			{{FVC}}  &44.24&52.53 &51.84 &58.61 &59.43  & 64.68\\
			\hline
			{{PLVC}}  &67.87 &68.95  &72.07  & 70.92 &73.31&73.09 \\
			\hline
			{{VVC}}  &64.86  &68.99 &69.99 & 71.75 &71.89& 72.84\\
			\hline
			DCVC-FM  &70.85 &73.02 &\textbf{74.68} &74.86 & \textbf{75.58} & 75.56\\
			\hline
			VCS  &67.79 &69.53 &72.28 &72.93 & 74.44& 74.94 \\
			\hline
			{{SMC}} &{70.84}  &{71.89} &73.86  &73.35 & 74.92 & 74.76\\
			\hline
			{{SMC++}} &\textbf{72.51}   &\textbf{73.07}  &74.19 & \textbf{75.41}&75.33 &\textbf{76.90} \\
			\arrayrulecolor{gray}\cdashline{1-7}[5pt/3pt]
			{\textcolor{gray}{Original}} &\textcolor{gray}{78.60}  &\textcolor{gray}{79.00}&\textcolor{gray}{78.60}  &\textcolor{gray}{79.00}&\textcolor{gray}{78.60}  &\textcolor{gray}{79.00}  \\
			\hline
		\end{tabular}
	}
	\label{tab:MOT_SOTA}
\end{table}

\begin{table}[!tbp]
	\centering
	\caption{
		VOS performance comparison of different coding methods on DAVIS2017.
		The codec setting is LDP mode with GOP size 10.
		``Original'' denotes the results with original videos, which is the performance upper bound.
	}
	\vspace{-2mm}
	\tabcolsep=1.0mm
	\scalebox{0.99}{
		\begin{tabular}{|c|c|c|c|c|c|c|}
			\hline
			Bpp &\multicolumn{2}{c|}{@0.01bpp} & \multicolumn{2}{c|}{@0.02bpp} & \multicolumn{2}{c|}{@0.03bpp} \\
			\hline
			Metric &{$\mathcal{J}$\&$\mathcal{F}$} (\%) & \textbf{$\mathcal{F}$} (\%)&
			{$\mathcal{J}$\&$\mathcal{F}$} (\%) & \textbf{$\mathcal{F}$} (\%)&
			{$\mathcal{J}$\&$\mathcal{F}$} (\%) & \textbf{$\mathcal{F}$} (\%)
			   \\
			\hline
			{{HEVC}}&57.68   &58.51&73.28  &75.96&77.45 &80.56 \\
			\hline
			{{FVC}}  &62.39  &63.55&75.87 &68.85&71.68 &74.15 \\
			\hline
			{{PLVC}}  &61.45&62.87&74.60&77.62 &77.84 &81.13\\
			\hline
			VVC  &67.47   &69.36&75.87&78.82 &79.32&82.57\\
			\hline
			DCVC-FM  &74.18 &76.91&79.63 &83.06 &82.25&85.68\\
			\hline
			VCS  &72.44 &76.44 &79.89 &83.73&82.24&86.21\\
			\hline
			{{SMC}} &{74.50}  &{78.70} &78.98 & 82.93 &81.40&85.38 \\
			\hline
			{{SMC++}} &\textbf{75.48}   &\textbf{80.12} &\textbf{80.55}  &\textbf{84.91} &\textbf{82.43} &\textbf{86.43}\\
			\arrayrulecolor{gray}\cdashline{1-7}[5pt/3pt]
			{\textcolor{gray}{Original}} &\textcolor{gray}{87.70}  &\textcolor{gray}{91.33}&\textcolor{gray}{87.70}  &\textcolor{gray}{91.33} &\textcolor{gray}{87.70}  &\textcolor{gray}{91.33}  \\
			\hline
		\end{tabular}
	}
	\label{tab:VOS_SOTA}
	 	\vspace{-2mm}
\end{table}

 \textbf{Implementation Details.}
We set the masking ratio for the MAE task to 90\%, and employ six encoder and two decoder ViT blocks with divided space-time attention~\cite{dosovitskiy2020image, bertasius2021space}. All network weights are randomly initialized. $K$, $\alpha$, and $\beta$ are set to 5, 1 and, 0.1, respectively. Training is performed on 256$\times$256 video clips (length 8) using the Adam optimizer~\cite{kingma2014adam} with an initial learning rate of $1\times10^{-4}$, decayed to $1\times10^{-5}$ during the final 100k iterations.
The total iteration number is 1M. The mini-batch size is 24. The quantization step is omitted for the first 100k iterations to allow precise gradient propagation. The SMC model is trained in approximately six days, on eight Nvidia 3090Ti GPUs. SMC++ is subsequently initialized from SMC and further trained for 500k iterations, with the learning rate maintained at $1\times10^{-4}$ for the first 400k iterations and decayed to $1\times10^{-5}$ during the final 100k iterations, also taking about six days on eight Nvidia 3090Ti GPUs.

 \subsection{\textsf{Experimental Results}}

 \textbf{{Action Recognition.}}
In Table~\ref{tab:benchmark_ar}, we evaluate various video compression methods on the action recognition task. Our basic SMC model remarkably outperforms traditional codecs HEVC and VVC, the learnable codecs FVC and DCVC-FM, and the perceptual codec PLVC.
For example, on the UCF101-TSM setting, our SMC model shows 22\% and 10\% Top1 accuracy improvement over FVC and VVC, respectively, at 0.04bpp. On the HMDB51-TSM setting, SMC outperforms the DCVC-FM by 18\% at 0.03bpp.
Although the most recent semantic coding approach VCS surpasses our basic SMC model, our enhanced SMC++ model outperforms it again, notably with over 5\% Top1 accuracy gain on the Diving48-TSM setting at 0.03bpp.
Moreover, on the large-scale Kinetics400 dataset, our SMC++ model surpasses VCS and DCVC-FM by 1.86\% and 13.42\% Top1 accuracy, respectively, at 0.06bpp.
When being evaluated with the Slowfast and TimeSformer action recognition networks, our SMC++ model still outperforms all other approaches by a large margin, \textit{i.e.}, about 4\% Top1 gain over VCS at 0.02bpp.
The rate-performance (RP) curves of different methods are provided in the 1st and 2nd rows of Figure~\ref{fig:lbuv_mot_blind}

\begin{table}[!t]
	\centering
	\caption{
		\revise{Comparison of different coding methods on FineGym99~\cite{shao2020finegym}, using TSM action model.}
	}
	\vspace{-2mm}
	\tabcolsep=2mm
	\revisefigure{
		\begin{tabular}{|c|c|c|c|c|}
			\hline
			& \multicolumn{4}{c|}{TSM Top1 (\%)} \\
			\hline
			Bpp&  @0.02bpp & @0.04bpp & @0.06bpp & @0.08bpp \\
			\hline
			VVC    &  42.84 & 71.17 &80.91& 85.53\\
			\hline
			DCVC-FM&  54.67 & 77.34 &83.46& 85.85\\
			\hline
			VCS   &  61.16 & 77.71 &83.02& 85.21\\
			\hline
			SMC   &  62.43 & 80.00 &85.17& 87.30\\
			\hline
			SMC++ &  \textbf{70.67} & \textbf{82.13} &\textbf{86.21}& \textbf{87.70}\\
			\hline
		\end{tabular}
	}
	\label{tab:finegym_tsm}
	\vspace{-4mm}
\end{table}

 \revise{
Finally, we evaluate the performance of our framework on the fine-grained action recognition dataset FineGym99~\cite{shao2020finegym}, which requires distinguishing subtle action differences in various gymnasium videos.  
As shown in Table~\ref{tab:finegym_tsm}, our SMC method already outperforms previous approaches by a large margin. For instance, at 0.02bpp, SMC achieves a Top-1 accuracy of 62.43\%, whereas state-of-the-art learnable video codec DCVC-FM and semantic video codec VCS achieve only 54.67\% and 61.16\%, respectively. This demonstrates the strong fine-grained action discrimination capability of our approach.  
Our enhanced SMC++ further establishes a new state-of-the-art by building upon and improving SMC’s performance.
}

 \textbf{Multiple Object Tracking (MOT).}
Beyond action recognition, we evaluated our coding approach on the more challenging Multiple Object Tracking (MOT) task, demanding precise localization and robust feature extraction under occlusions. Results are shown in Table~\ref{tab:MOT_SOTA}.
At low bitrates like 0.01bpp, our SMC model outperforms FVC, PLVC, and VVC by 26.50\%, 2.97\%, and 5.98\% in MOTA, respectively, matching DCVC-FM. Our SMC++ surpasses DCVC-FM by a large margin of 1.66\%.
Compared to the semantic coding approach VCS, SMC and SMC++ show a 3.05\% and 4.72\% MOTA gain, respectively.

%
%
%

We also provide the RP curves in Figure~\ref{fig:lbuv_mot_blind}. SMC outperforms all previous methods except DCVC-FM, while SMC++ leads across most metrics. At higher bitrates, SMC++'s MOTP performance is unsatisfactory, which denotes the bounding-box accuracy of the detected objects.
We conjecture the reasons is that the Transformer-based Blue-Tr module within SMC++ emphasizes high-level features
and compromises local positioning accuracy. We plan to fix this by adaptively choosing between CNN- and Transformer-based compression modules.

\textbf{Video Object Segmentation (VOS).}
We further evaluate different coding methods on the VOS task. As shown in Table~\ref{tab:VOS_SOTA}, our basic SMC model surpasses all previous approaches, except VCS and DCVC-FM. For example, SMC exceeds VVC and PLVC by 6.73\% and 12.75\% in terms of $\mathcal{J}$\&$\mathcal{F}$ at the 0.01bpp level.
After introducing the masked motion prediction and Blue-Tr module, our resultant SMC++ model outperforms VCS by 3.04\% and 3.68\% in terms of $\mathcal{J}$\&$\mathcal{F}$ and $\mathcal{F}$ respectively, and DCVC-FM by 1.30\% and 3.21\% in the same metrics, at 0.01bpp. The RP curves of recent approaches are provided in the fourth row of Figure~\ref{fig:lbuv_mot_blind}.

\revise{We further evaluate E-Nerv~\cite{li2022nerv} on the VOS task, as the NerV~\cite{chen2021nerv} series has recently shown promising performance in video compression but has yet to be assessed on video understanding tasks.  
	Due to the high computational cost of the current NerV~\cite{chen2021nerv} approaches, we tested only one sequence ``bike-blackswan\_1'', from the DAVIS2017 dataset.  
	E-Nerv~\cite{li2022nerv} achieves a $\mathcal{J}$ value of 82.28\% @0.079 bpp, which is much inferior to our approach, i.e., $\mathcal{J}$ value of 90.28\% at a even lower bitrate 0.048 bpp. This highlights the limitations of NerV approaches, as their overfitting nature restricts them to compressing low-level details.  
	In contrast, our method, which leverages semantics learned from large-scale videos, achieves superior results.}
	
\revise{
\textbf{Comparison to Other Recent Approaches.}
We compare our approach with two more recent methods:
(1) DeepSVC~\cite{lin2023deepsvc}, a scalable video semantic compression approach, and
(2) HVFVC~\cite{li2023high}, a recent perceptual video compression method.
As shown in Figure~\ref{fig_more_sota}, HVFVC is much inferior to both SMC and SMC++ models, in both recognition and segmentation tasks. This is because HVFVC prioritizes perceptual quality but fails to effectively preserve semantic information critical for downstream tasks. At 0.025 bpp, HVFVC achieves only 77\% accuracy, while SMC++ reaches 85\% on UCF101-TSM, representing an 8\% performance gap.
DeepSVC is comparable to the baseline SMC model but falls notably short of SMC++. For instance, on UCF101-TSM at 0.075 bpp, DeepSVC, SMC, and SMC++ achieve 90.01\%, 89.82\%, and 91.41\%, respectively.
Similarly, on the VOS task, DeepSVC achieves about 81\% $\mathcal{J}\&\mathcal{F}$ score, significantly lower than the 82.3\% achieved by SMC++ on the DAVIS2017-XMEM setting at 0.03 bpp.

 \begin{figure}[!t]
 	\centering
 	\tabcolsep=0mm
 	\revisefigure{
 		\begin{tabular}{cc}
 			\includegraphics[width=0.48 \linewidth]{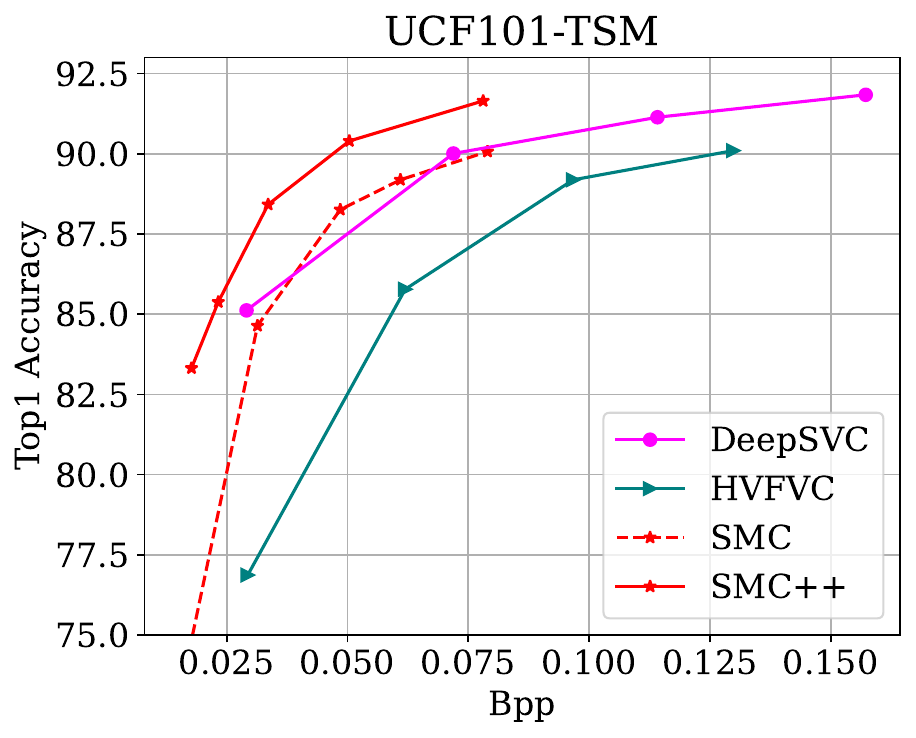}&
 			\includegraphics[width=0.48 \linewidth]{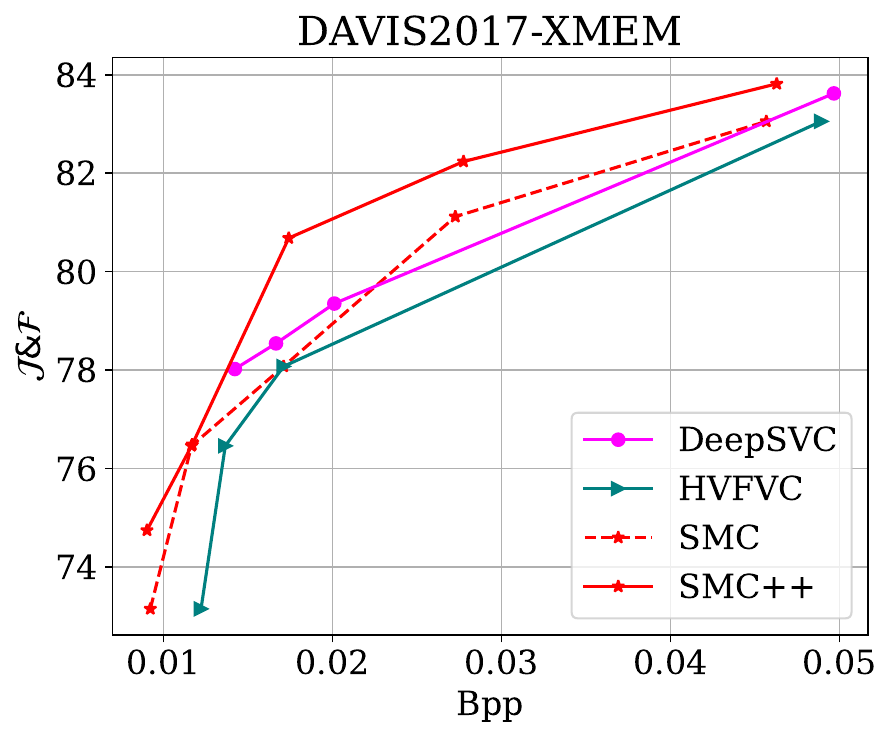} 
 		\end{tabular}
 	}
 	\vspace{-4mm}
 	\caption{
 	\revise{Comparison of our approach with two recent approaches, i.e., DeepSVC~\cite{lin2023deepsvc} and HVFVC~\cite{li2023high}, on action recognition and video object segmentation (VOS) tasks.}
 	}
 	\vspace{-2mm}
 	\label{fig_more_sota}
 \end{figure}
 
\textbf{Compatibility to Other Codecs as the Base Layer.}
Beyond using VVC as the base coding layer, we also evaluate our framework with the fast H.264 and H.265 codecs to assess improvements in coding efficiency, as well as with the state-of-the-art DCVC-FM to verify whether our proposed framework remains effective when applied to an already-strong codec. As shown in Figure~\ref{fig:ablation_base_codec}, our approach consistently enhances performance across these various codecs on semantic tasks, action recognition, and VOS tasks.

  \begin{figure}[!t]
  	\vspace{-2mm}
  	\centering
  	\tabcolsep=1mm
  	\revisefigure{
  		\begin{tabular}{cc}
  			\includegraphics[width=0.48 \linewidth]{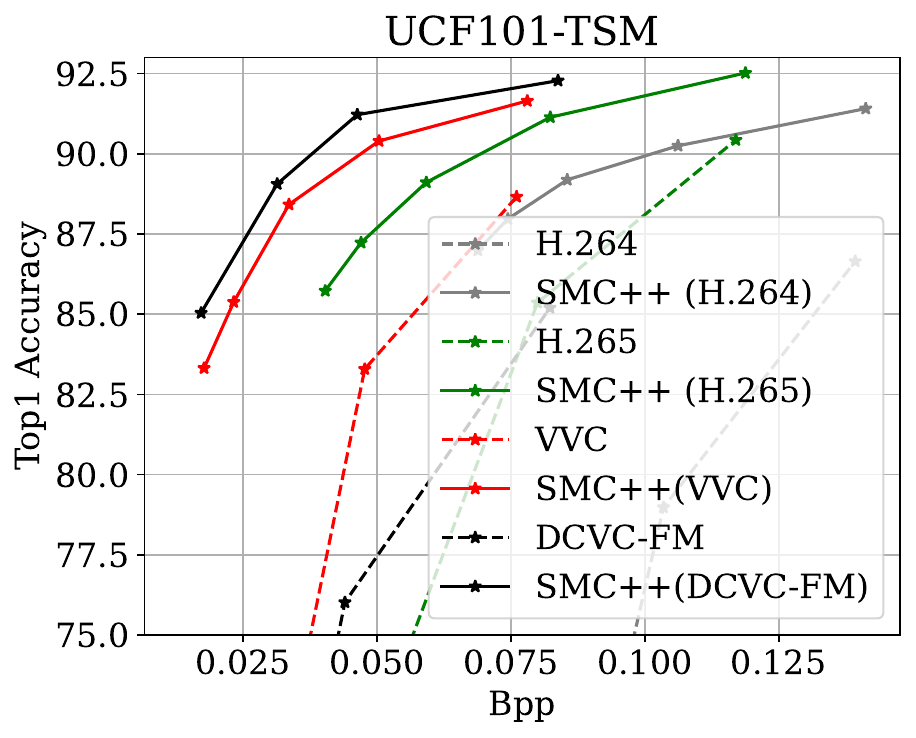}&
  			\includegraphics[width=0.48 \linewidth]{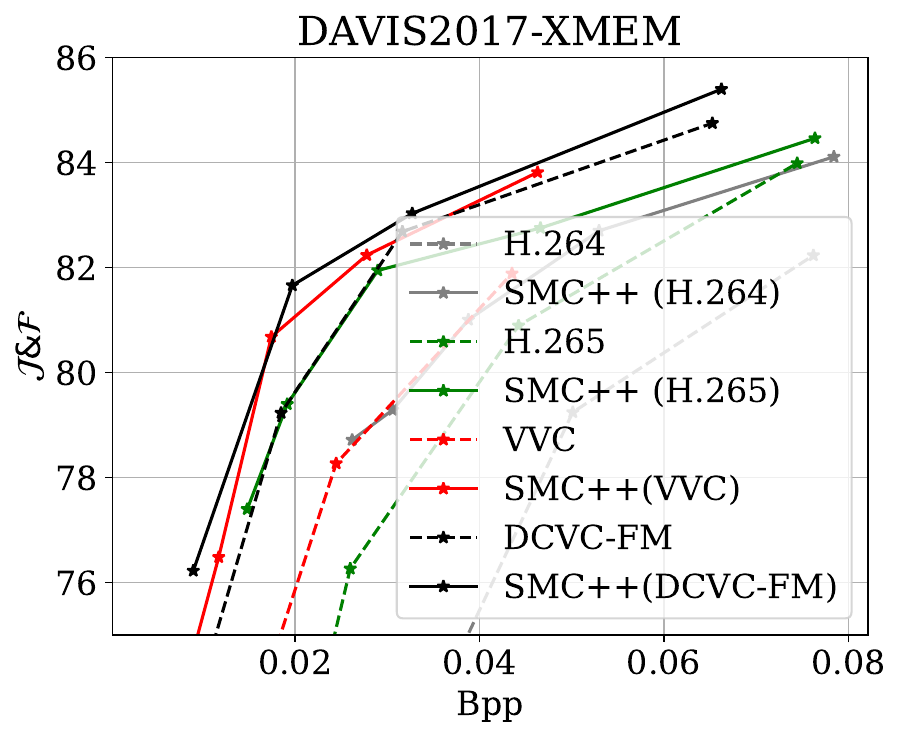} 
  		\end{tabular}
 }
  	\vspace{-4mm}
  	\caption{
  		\revise{Performance enhancement of SMC++, when using H.264, H.265 and DCVC-FM as the base coding layer.}
  	}
  	\vspace{-4mm}
  	\label{fig:ablation_base_codec}
  \end{figure}
  \begin{figure}[!t]
  	\centering
  		\renewcommand\arraystretch{0.0}
  	\tabcolsep=1mm
 		\revisefigure{
 			\begin{tabular}{cc}
 				\includegraphics[width=0.48 \linewidth]{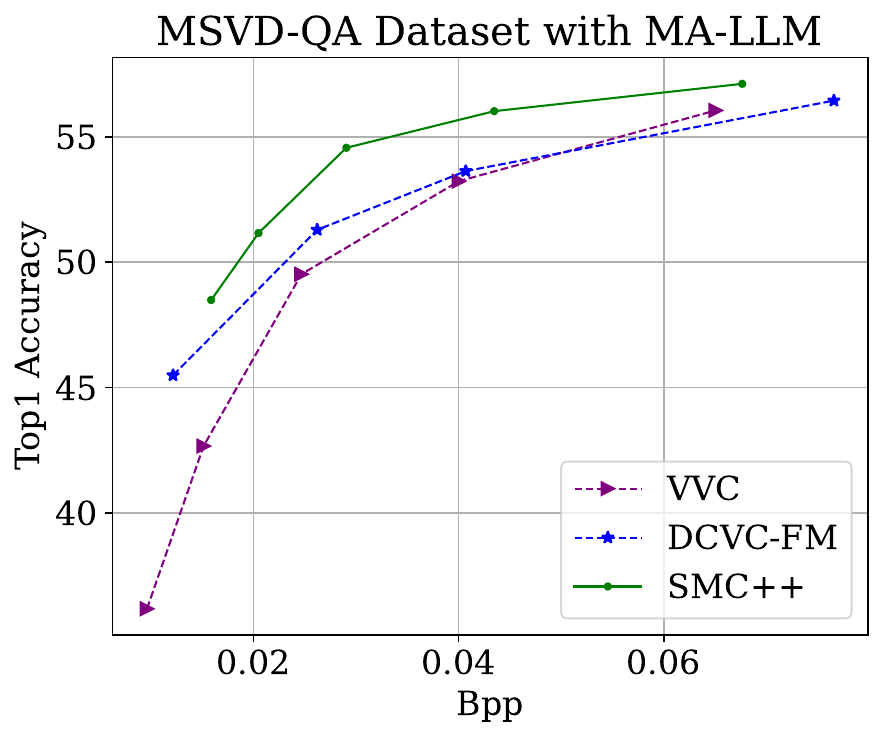}&
 				\includegraphics[width=0.48 \linewidth]{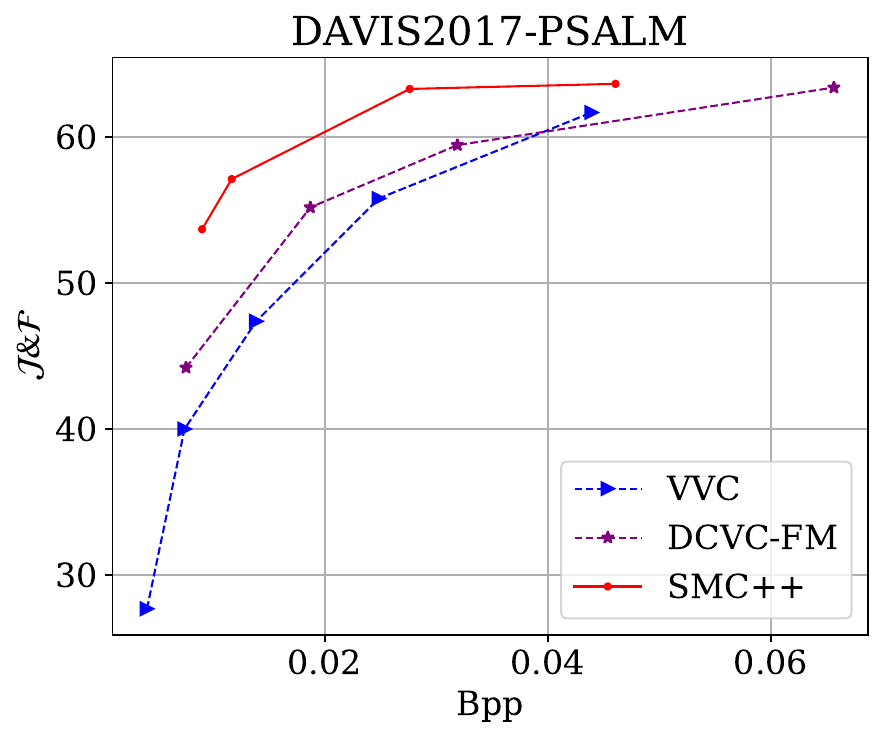}
 			\end{tabular}
 		}
  	\vspace{-2mm}
  	\caption{
  		\revise{Comparison of different approaches on VQA and VOS tasks, with LLM-based downstream models.}
  	}
  	\vspace{-6mm}
  	\label{fig:perform_vqa_seg}
  \end{figure}
  }

\textbf{Effectiveness on LLM-based downstream models.}
Large multi-modal models are becoming increasingly popular for video analysis tasks. To evaluate the compatibility of our method with such models, we conducted additional experiments on the video question-answering (VQA) task using the MSVD-QA dataset~\cite{xu2017video} with MA-LMM~\cite{he2024ma} model, and on the video object segmentation (VOS) task using the DAVIS2017 dataset with PSALM~\cite{zhang2024psalm} model.  

As shown in Figure~\ref{fig:perform_vqa_seg}, for the VQA task, our approach demonstrates consistently strong performance. For example, at 0.02 bpp, our method outperforms VVC and DCVC-FM by 5.85\% and 3.12\%, respectively. This proves that our method effectively preserves semantic information crucial for cross-modality reasoning tasks, even at very low bitrates.  
For the VOS task, we evaluate with the LLM-based approach PSALM~\cite{zhang2024psalm}. Our approach remarkably surpasses previous coding methods. For instance, at 0.01 bpp, our method exceeds VVC and DCVC-FM by 11.23\% and 7.15\%, respectively, in terms of the $\mathcal{J\&F}$ metric. This demonstrates that, even in the era of LLM, our video semantic compression strategy remains highly effective, supporting LLM-based video analysis models at low bitrate levels.

\begin{figure}[!t] 
 	\centering
 	\renewcommand\arraystretch{0.0}
 	\newcommand{\widthscalefive}{0.22}
 	\tabcolsep = 0.6mm
 	\scalebox{1}{
 		\begin{tabular}{cc}
 			\includegraphics[width=\widthscalefive \textwidth]{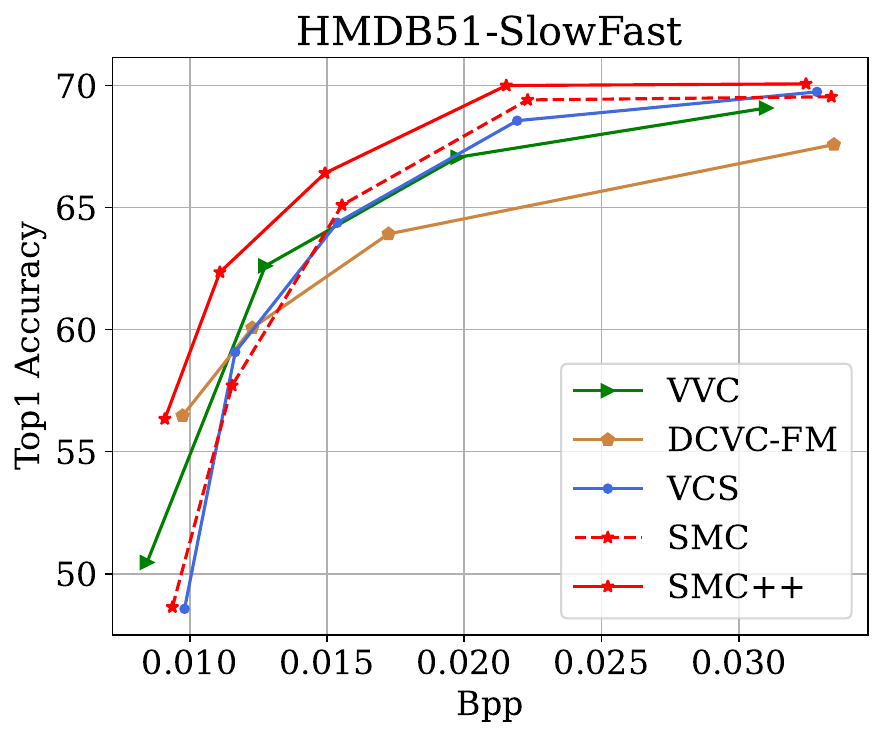} &
 			\includegraphics[width=\widthscalefive \textwidth]{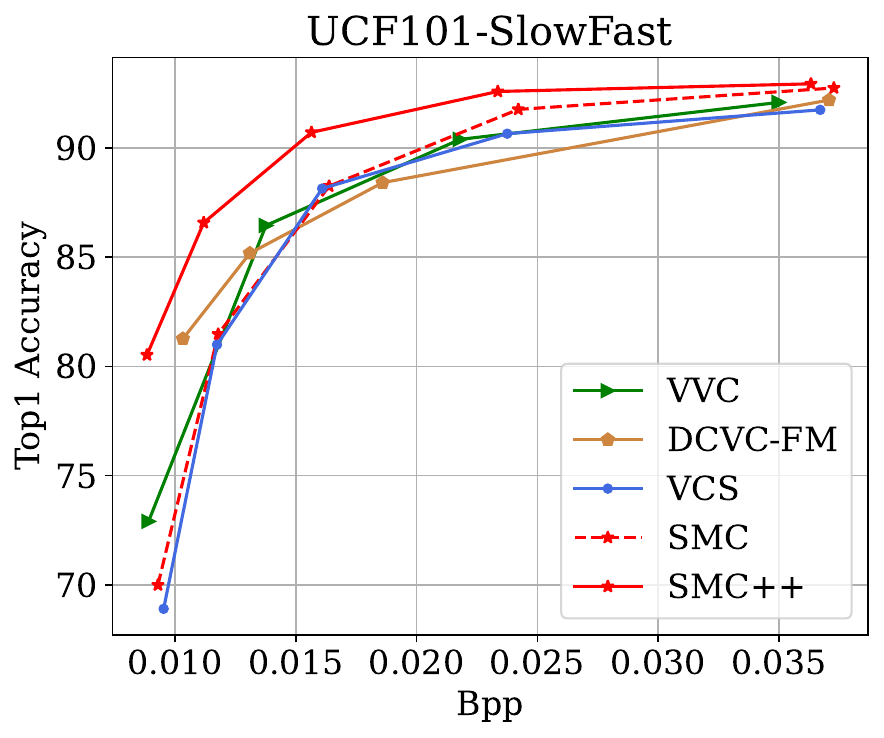} \\
 			\includegraphics[width=\widthscalefive \textwidth]{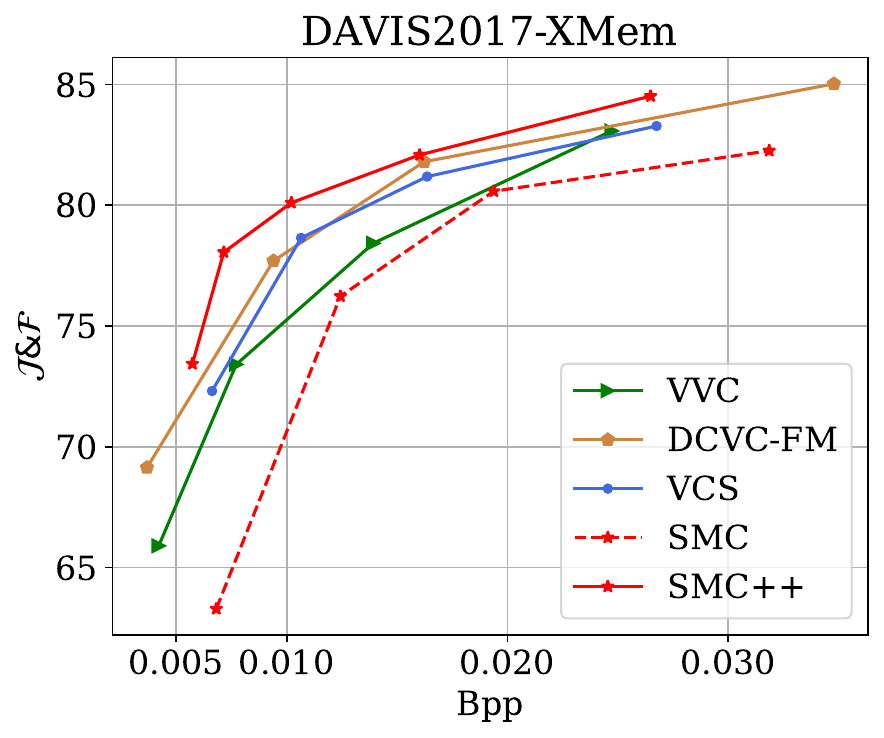} &
 			\includegraphics[width=\widthscalefive \textwidth]{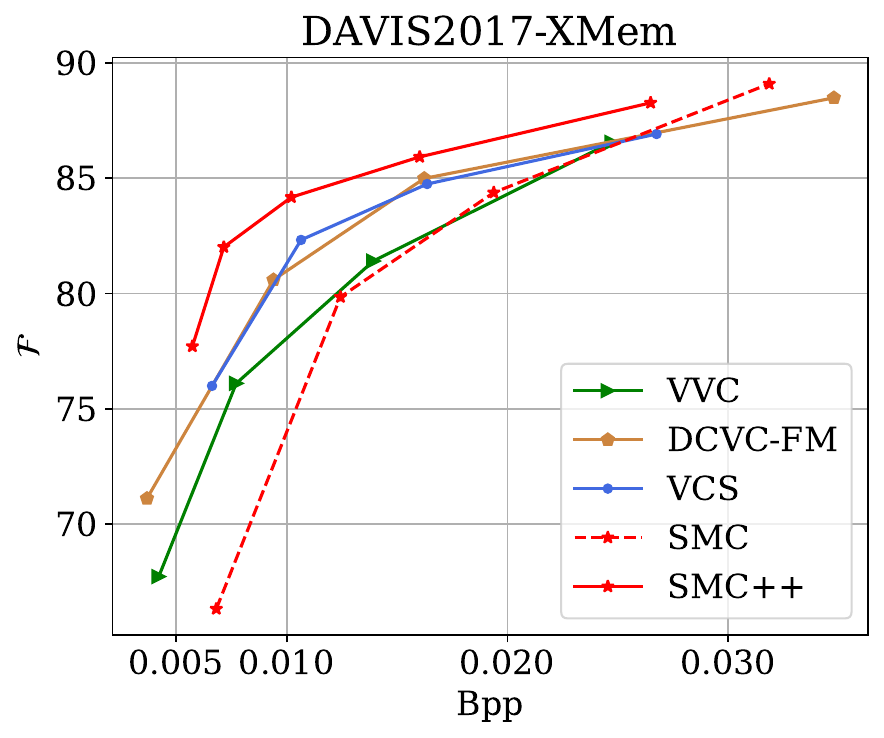} 
 			
 		\end{tabular}
 	}
 	\vspace{-2mm}
 	\caption{
 		Semantic coding performance of different methods on Action Recognition (\textit{1st} row) and VOS (\textit{2nd} row) tasks, evaluated on LDP mode with GOP size 32.
 		The plot titles are in \{Dataset\}-\{Task Model\} format.
 	}
 	\vspace{-2mm}
 	\label{fig:results_gop32}
 \end{figure}

\textbf{More Advanced Test Conditions.}
Recent learnable codecs~\cite{li2021deep}\cite{li2023neural1} have adopted a more challenging test condition, \textit{i.e.}, group-of-picture (GOP) size of 32. This larger GOP size poses a greater challenge than small GOP sizes, because of the error propagation issue.
As shown in Figure \ref{fig:results_gop32}, our advanced SMC++ model maintains its effectiveness under this more advanced test condition.

\begin{figure}[!t] 
	\centering
	\renewcommand\arraystretch{0.0}
	\newcommand{\widthscalefive}{0.22}
	\tabcolsep = 0.6mm
	\hspace{2mm}\scalebox{1}{
		\begin{tabular}{cc}
			\hspace{-4mm}\includegraphics[width=\widthscalefive \textwidth]{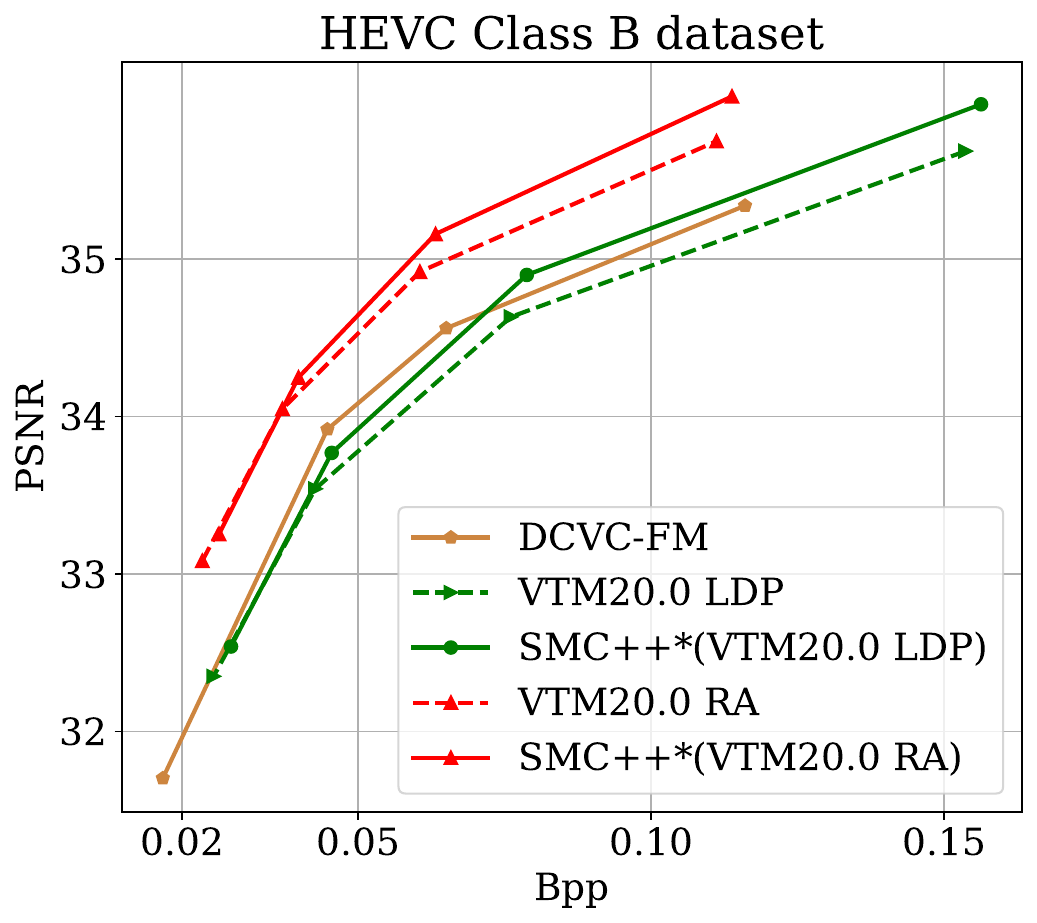} &
			\includegraphics[width=\widthscalefive \textwidth]{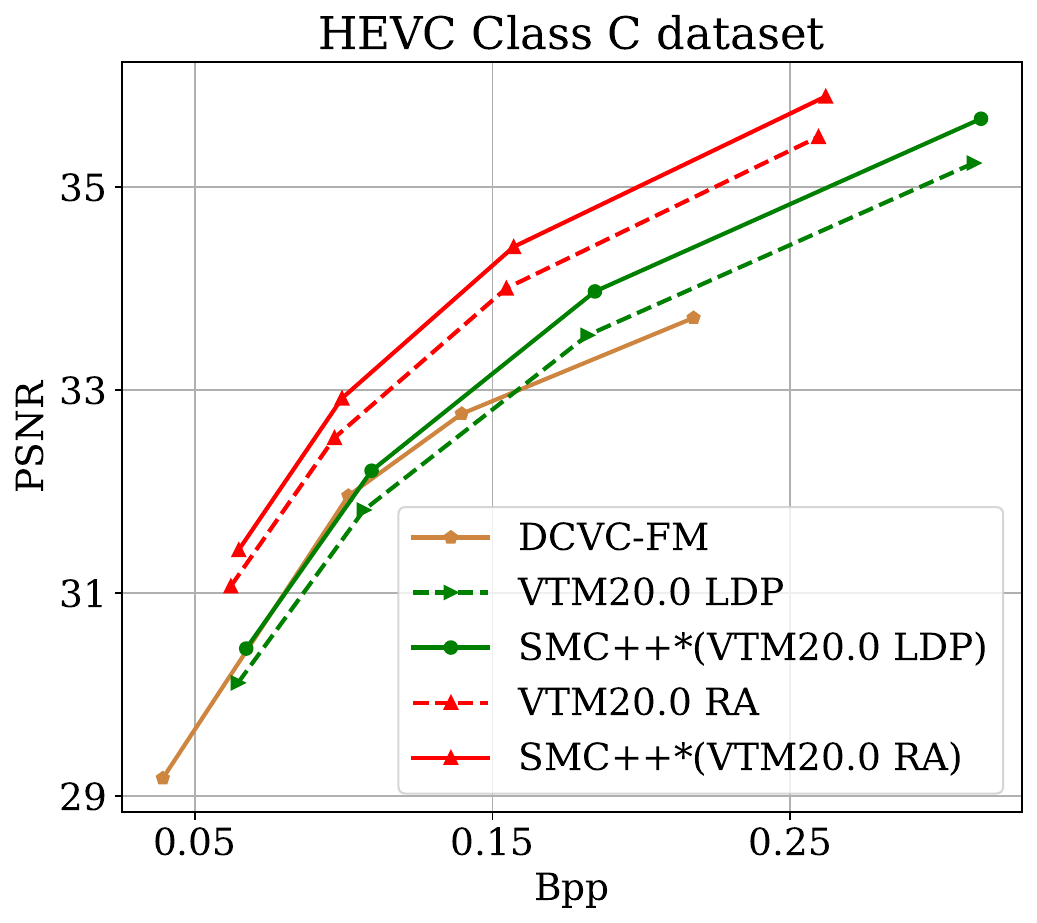} \\
			\hspace{-4mm}\includegraphics[width=\widthscalefive \textwidth]{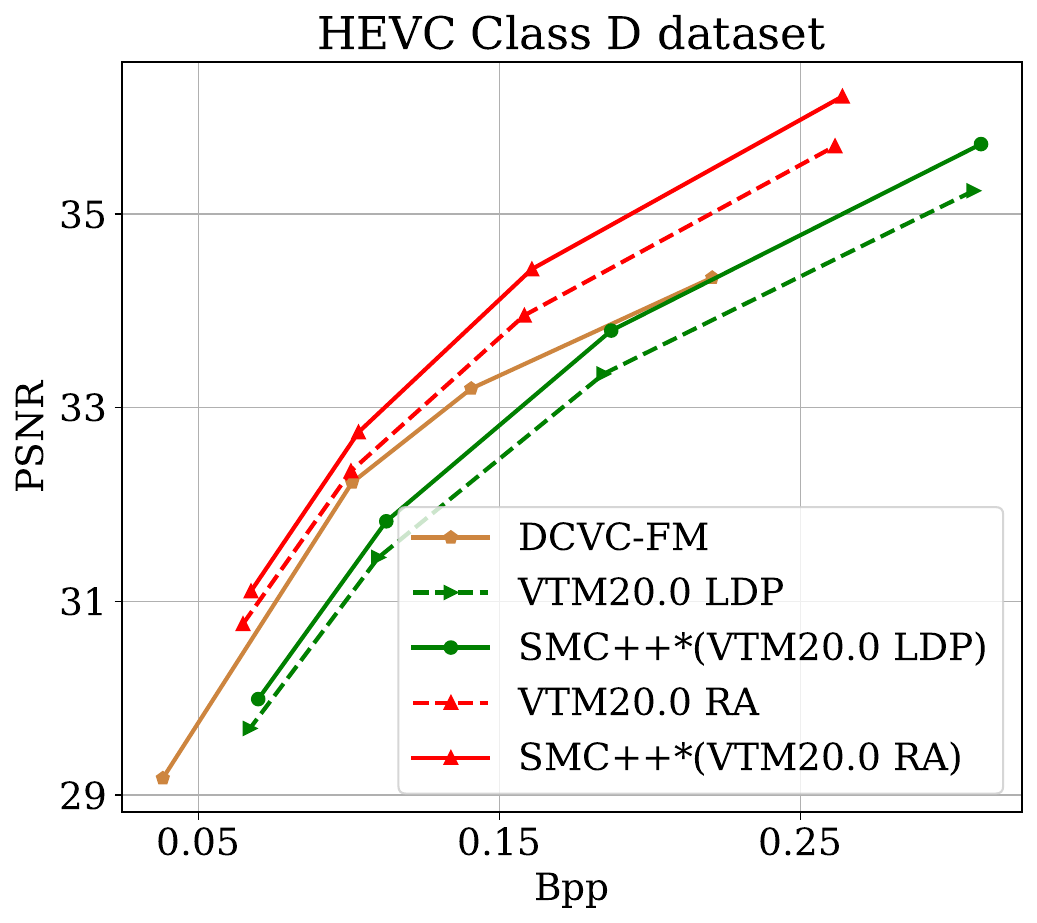} &
			\includegraphics[width=\widthscalefive \textwidth]{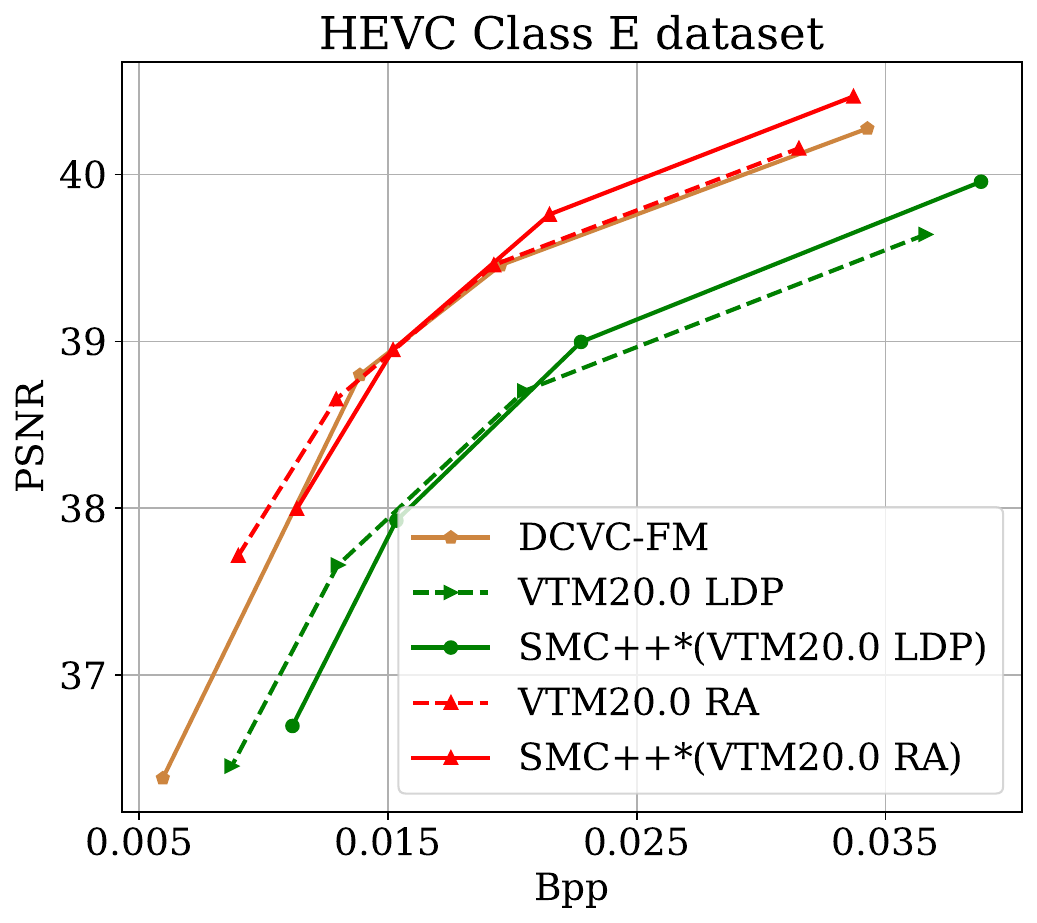} \\
		\end{tabular}
	}
	\vspace{-2mm}
	\caption{
		Comparison of different coding methods in terms of PSNR.
		The test condition is 96 frames with intra-period 32.
		LDP and RA denote the Low-Delay P frame and Random Access mode, respectively.
	}
	\vspace{-4mm}
	\label{fig:visual_coding}
\end{figure}

\textbf{Towards High-fidelity Compression.}
To enable high-fidelity decoding, we enhance SMC++ with DR-Net to form SMC++*, employing VTM20.0 as the base VVC layer.
As shown in Figure~\ref{fig:visual_coding}, SMC++* achieves an approximately 0.3dB gain on the HEVC Class C dataset, compared to the advanced RA mode of VTM at 0.25bpp.

 \subsection{\textsf{Ablation Studies and Analysis for SMC Model}}\label{sec:method_ana}

\begin{figure}[!t]
 	\centering
 	\newcommand{\widthscalefive}{0.222}
 	\tabcolsep=2mm
 	\begin{tabular}{cc}
 		\includegraphics[width=\widthscalefive \textwidth]{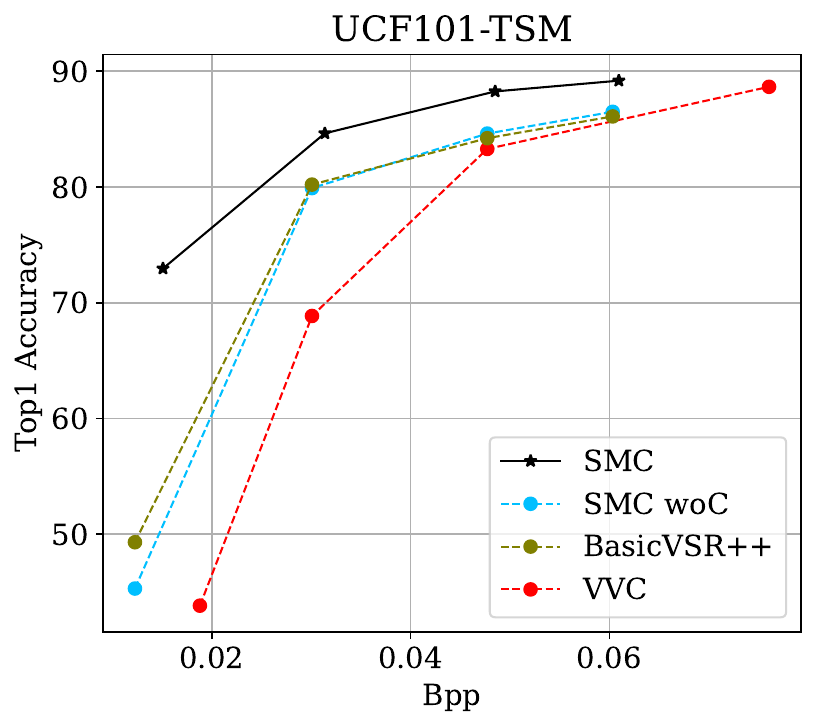} &\includegraphics[width=\widthscalefive \textwidth]{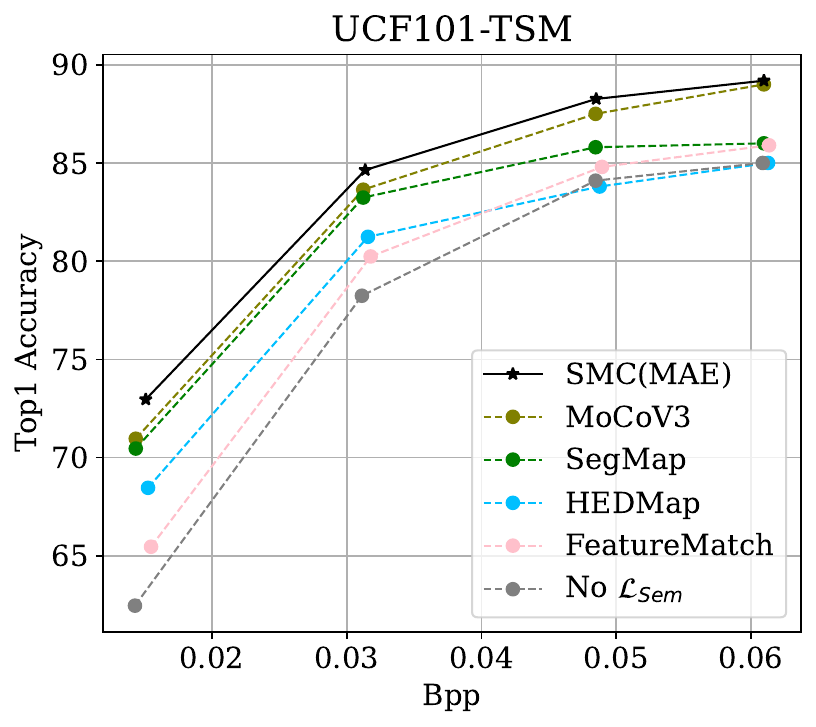}
 	\end{tabular}
 	 	\vspace{-4mm}
 	\caption {
 		\textit{Left:}
 		Ablation study on SMC framework.
 		\textit{SMC woC} denotes the semantic compensation procedure is removed.
 		\textit{BasicVSR++} denotes the compressed videos by VVC codec are enhanced by state-of-the-art BasicVSR++ method~\cite{chan2022basicvsr++}.
 		\textit{Right:}
 		Comparison of different semantic learning objectives. 
 	}
 	\label{fig:ablation_smc}
 	\vspace{-1mm}
 \end{figure}

 {\textbf{Component Study.}}
 We first train a variant model denoted by \textit{SMC wo Comp}, in which the semantic compensation operation is removed.
 As shown in Figure~\ref{fig:ablation_smc} (\textit{Left}), the performance is drastically decreased by 16\% in terms of Top1 accuracy at 0.02bpp.
 Further, we replace the simple UNet-Style visual-semantic fusion network of \textit{SMC wo Comp} with a compressed video enhancement method \textit{BasicVSR++}~\cite{chan2022basicvsr++}\cite{chan2022generalization}.
 The framework is degraded to a ``lossy compression plus enhancement'' paradigm.
 The improvement over the simple UNet is marginal and is still far behind our full semantic compensation framework by -14.23\%@0.02bpp.
 These results prove that the distorted semantics cannot be fixed by video enhancement approaches and should be particularly compensated.
 
  \begin{table}[!t]
 	\centering
 	\caption{
 		Impact of different self-supervised learning objectives to SMC performance, on action recognition and MOT tasks.
 		Our masked learning paradigm performs much better, especially at the fine-grained MOT task.
 	}
 	\vspace{-2mm}
 	\tabcolsep=2mm
 	\begin{tabular}{|c|c|}
 		\hline
 		UCF101-TSM@{0.02bpp} &  MOT17-ByteTrack@{0.01bpp} \\
 		\hline
 		{MoCoV3 $\rightarrow$ \textbf{Masked Leaning}}   &   {MoCoV3 $\rightarrow$ \textbf{Masked Leaning}}  \\
 		\hline
 		Top1: 75.20\% $\rightarrow$ \textbf{76.48\%}& MOTA: 67.62\% $\rightarrow$\textbf{70.84\%}(+3.22\%) \\
 		\hline
 	\end{tabular}
 	\vspace{-5mm}
 	\label{tab:ab_mae_vs_cl}
 \end{table}

 {\textbf{Different Semantic Learning Objectives $\mathcal{L}_{Sem}$.}}
 In this section, we train several SMC models by equipping them with different learning objectives, \textit{i.e.}, {MoCoV3}~\cite{chen2020improved}, {SegMap}, {HEDMap}, and {FeatureMatch}.
 The first one is purely self-supervised.
 The latter three ones regularize the compressed video to be similar to the original video in terms of DeepLabv3~\cite{chen2017rethinking} semantic segmentation map, HED~\cite{xie2015holistically} edge map, and VGG16~\cite{simonyan2014very} feature map, respectively.

 As shown in Table~\ref{tab:ab_mae_vs_cl}, our MAE-powered approach consistently outperforms the contrastive learning-based method {MoCoV3} on two tasks of different granularities, \textit{i.e.}, action recognition and MOT, by 1.28\% Top1 accuracy and 3.22\% MOTA, respectively.
 This clearly proves the semantics learned by MAE is more generalizable than that learned by contrastive learning, as well as more fine-grained.
 
We also demonstrate the superiority of the MAE learning objective over other heuristic objectives through experiments.
A similar comparison has been conducted in our prior work~\cite{tian2022coding}. We replicate this analysis for the sake of self-containment and to underscore that a similar conclusion applies to simpler network architectures of SMC model.

First, we find that the hand-crafted {SegMap} loss achieves similar performance to {MoCoV3} in the lower bitrate ranges.
 However, this paradigm is still far behind our learning objective, \textit{e.g.}, about a 3\% performance gap at 0.05bpp level.
 After replacing the segmentation map with the HED edge map, the performance is further dropped, because the edge map does not contain the category information of each object region and is of less semantics.
 The model trained with {FeatureMatch} shows the worst results, probably because feature values are denser than the discretized segmentation/edge maps, which increases the bitcost.
 
\begin{figure}[!t]
	\renewcommand\arraystretch{0.0}
	\centering
	\tabcolsep=0.5mm
	\revisefigure{
	\hspace{2mm}	\begin{tabular}{cc}
			\includegraphics[width=0.48 \linewidth]{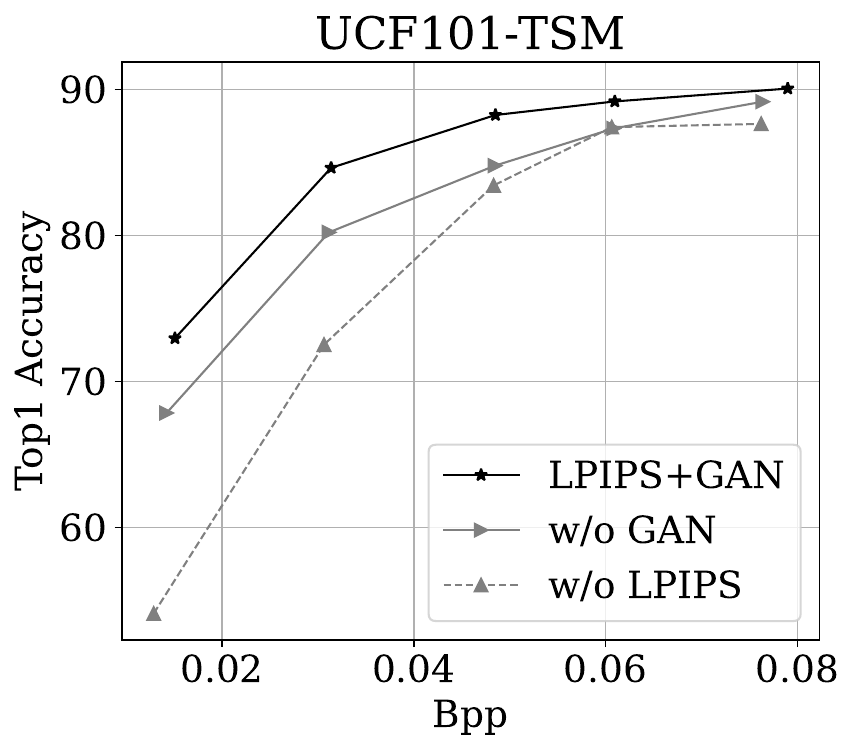}&
			\includegraphics[width=0.49 \linewidth]{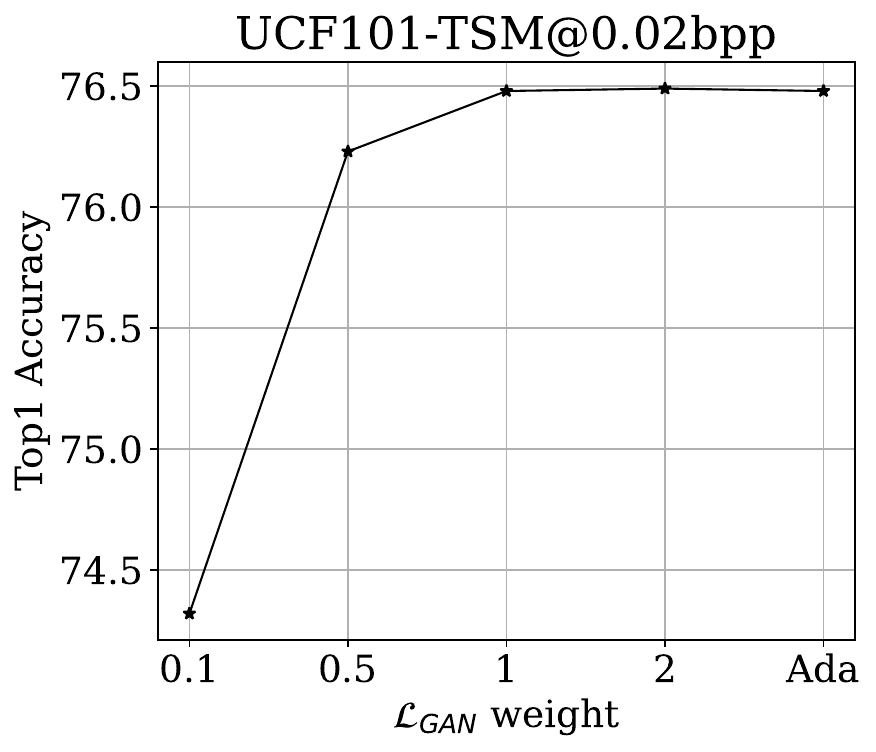}\\
			{\scriptsize ~~(a)} & {\scriptsize ~~~(b)}
		\end{tabular}
	}
	\vspace{-2mm}
	\caption{
		\revise{
			(a) Ablation study on the LPIPS+GAN terms for SMC.
			``w/o'' denotes removing the term.
			(b) \revise{Impact of $L_{GAN}$ weight to SMC. ``Ada'' denotes adaptive weight.}
			}
	}
	\vspace{-2mm}
	\label{fig:impact_gan_lpips_gan}
\end{figure}

 \revise{
\textbf{Ablation on Visual Quality Loss Terms.}
Remembering that we ensure the reconstructed video photo-realistic, by using the combined LPIPS plus GAN terms in VQGAN.
We conduct more studies to study if the two terms are critical to our problem.
As shown in Figure~\ref{fig:impact_gan_lpips_gan}, when the GAN loss is removed, the performance degrades by approximately 5\% at 0.02 bpp. In contrast, removing the LPIPS loss severely degrades the performance, with an accuracy drop of about 15\% at 0.02 bpp. This degradation arises because the local structural features within each patch are no longer preserved, whereas our MAE loss only retains the global interactions among different features.
In summary, both the two loss terms are critical for a high-performance video semantic compression system.\revise{
	Furthermore, we explored other weighting strategies for the GAN loss term, including the adaptive weighting approach employed in VQGAN~\cite{esser2021taming}. As shown in Figure~\ref{fig:impact_gan_lpips_gan} (b), these strategies did not result in obvious performance improvements.
}
}
 
\textbf{Analysis on Masked Learning Objective.}
First, we demonstrate that the lower MAE loss is positively related to the video semantic completeness, as demonstrated in Figure 1 and 2 of supplementary material.
 
Further, we study the influence of the masking ratio on the semantic compression performance.
 As shown in Table~\ref{tab:mask_ratio}, the 90\% masking ratio achieves the optimal performance, and setting the masking ratio to 75\% and 95\% achieves slightly inferior performances.
 The 50\% masking ratio shows a remarkable performance drop, as the textures can be simply completed by extending local textures, instead of reasoning high-level scene and structure semantics.
  \begin{table}[!t]
 	\centering
 	\caption{
 		Impact of different masking ratios to SMC performance, on the UCF101-TSM setting at 0.02bpp.
 	}
 	\vspace{-2mm}
 	\tabcolsep=4.5mm
 	\begin{tabular}{|c|c|c|c|c|}
 		\hline
 Masking ratio&  50\%  &   75\% &90\% & 95\% \\
 		\hline
\makecell[c]{ 
	Top1 (\%)} & 73.85 & 75.62  & \textbf{76.48} & 76.08 \\
 		\hline
 	\end{tabular}
 	\label{tab:mask_ratio}
 \end{table}
  \begin{table}[!t]
  	\vspace{-2mm}
 	\centering
 	\caption{\revise{Comparison of different MAE prediction targets for SMC, on UCF101-TSM setting.}}
 	\vspace{-2mm}
 	\tabcolsep=1.6mm
 	\revisefigure{
 		\hspace{1mm}\begin{tabular}{|c|c|c|c|c|}
 			\hline
 			\multirow{2}{*}{Method} & \multicolumn{4}{c|}{UCF101-TSM Top1 (\%)} \\
 			\cline{2-5}
 			&@ 0.02bpp & @ 0.04bpp & @ 0.06bpp& @ 0.07bpp \\
 			\hline
 			\makecell{Raw pixel}  &
 			76.48 & 86.46 & 89.11 &89.63 \\
 			\hline
 			\makecell{DINOv2 feature} &
 			77.00 & 86.48 & 89.51 & 89.60 \\
 			\hline
 		\end{tabular}
 	}
 	\vspace{-4mm}
 	\label{tab:mae_teacher}
 \end{table}
 \revise{
Finally, we explore using features extracted from pre-trained neural networks as prediction targets for the MAE task, instead of raw pixels. Specifically, we employ the pre-trained DINOv2~\cite{oquab2023dinov2} model to extract features, which serve as the prediction target during masked learning.  

As shown in Table~\ref{tab:mae_teacher}, the DINOv2-based approach achieves better results at very low bitrates due to its compact feature representation. However, at higher bitrates, it underperforms the pixel-prediction approach. For example, at 0.06 bpp, the DINOv2-based model performs 0.60\% worse. Additionally, the feature extraction process increases training time by 32\%. These results indicate that while feature-based targets are beneficial for extremely low bitrates, pixel-based targets remain stable across all bitrate levels.  
}

 \textbf{Effectiveness of NSS Strategy.}
 Our method adapts the vanilla MAE to the semantic compression task by suppressing the non-semantic information within its token space.
 
  \begin{table}[!t]
 	\centering
 	\caption{
 		Comparison of different non-semantic suppression (NSS) strategies on UCF101-TSM setting.
 		$Res$ (bpp) denotes the bitcost of the residual semantic stream of SMC.
 		The QP of the base codec VVC is set to 51.
 	}
 	 \vspace{-2mm}
 	\tabcolsep=2.2mm
 	\begin{tabular}{|c|c|c|c|c|}
 		\hline
 		&   {SMC VQ}   &    {SMC woSem}  & {SMC woNSS} &   {SMC}  \\
 		\hline
 		$Res$ (bpp) & 0.0048 & 0.0042  &0.0074&\textbf{0.0028 }\\
 		\hline
 		Top1 (\%) &   71.22& 71.38  & 	 69.57&\textbf{72.96}\\
 		\hline
 	\end{tabular}
 \vspace{-2mm}
 	\label{tab:ab_NSS}
 \end{table}
   \begin{table}[!t]
 	\centering
 	\caption{
 		\revise{Impact of $\beta$ to SMC, on UCF101-TSM@0.02bpp.
 		}
 	}
 	 \vspace{-2mm}
 	\tabcolsep=3.7mm
 	\revisefigure{
 		\begin{tabular}{|c|c|c|c|c|c|}
 			\hline
 			$\beta$&  0.01 & 0.05& 0.1&0.5&1 \\
 			\hline
 			Top1(\%)& 72.94 & 75.21 &\textbf{76.48} & 76.23 &  75.93 \\
 			\hline
 		\end{tabular}
 	}
 	\vspace{-2mm}
 	\label{tab:ablation_beta}
 \end{table}
 
 \begin{table}[!t]
 	\centering
 	\caption{
 		\revise{Impact of $K$ to SMC, on UCF101-TSM@0.02bpp.}
 	}
 	 \vspace{-2mm}
 	\tabcolsep=1.9mm
 	\revisefigure{
 		\begin{tabular}{|c|c|c|c|c|c|c|c|}
 			\hline
 			$K$&  1  &  3 & 5 & 7 & 9 & 11&15 \\
 			\hline
 			Top1 (\%)& 76.13 & 76.42  & \textbf{76.48} & 76.47 &  76.42 &  76.23 & 75.92\\
 			\hline
 		\end{tabular}
 	}
 	\vspace{-4mm}
 	\label{tab:ablation_k_value}
 \end{table}
 
 To study the necessity of this design, we first train a variant model \textit{SMC woNSS} by removing the $\mathcal{L}_{NSS}$ item from the loss function.
 As shown in Table~\ref{tab:ab_NSS}, the bitcost of \textit{SMC woNSS} is 2.6$\times$ larger than the full SMC model, because the low-level pixel information is back-propagated to the coding system without any selection. Moreover, this non-semantic information may be noises to the downstream tasks, \textit{i.e.}, the action recognition accuracy is dropped from 72.96\% to 69.57\%.
 Then, we train a variant model \textit{SMC woSem} by using a plain learnable variable as the condition of the GMM distribution, instead of the semantic feature $S$.
 Both the compression efficiency and the recognition accuracy of \textit{SMC woSem} are superior to \textit{SMC woNSS}, but still inferior to our SMC model.
 This implies that explicitly regularizing the information entropy of the MAE token space is beneficial to a coding system, and using learned semantics as the guidance further improves this idea.
 Finally, we use the vector quantization (VQ)~\cite{van2017neural} codebook to discretize the token space of MAE, and the resultant \textit{SMC VQ} model has similar performance to \textit{SMC woSem}, indicating that the idea of information suppression is important for a compression system, instead of its concrete implementation.

 {\textbf{Hyper-parameter Sensitivity.}}
 Setting the semantic loss item weight $\alpha$ to the values in the range [1,10] gives similar results.
 \revise{
	Then, we investigate the impact of $\beta$, which balances the Non-Semantics Suppressing term and the MAE term. As shown in Table~\ref{tab:ablation_beta}, increasing $\beta$ from 0.01 to 0.1 leads to consistent performance improvements, due to the increased semantic separability. However, when $\beta$ becomes too large, the representation capability of the video token features in MAE becomes overly constrained, resulting in degraded performance.
 	Regarding the GMM component number $K$, setting it within the range $[3,9]$ yields comparable results, with $K=5$ empirically achieving optimal performance, as demonstrated in Table~\ref{tab:ablation_k_value}. A small value such as $K=1$ results in an overly simplistic mixture distribution that inadequately captures the semantic features, while excessively large values such as $K=15$ create distributions that are too complex for effective optimization.
}

\begin{figure}[!t]
 	\centering
 	\newcommand{\widthscalefive}{0.45}
 	\tabcolsep=-4mm
 	\includegraphics[width=\widthscalefive \textwidth]{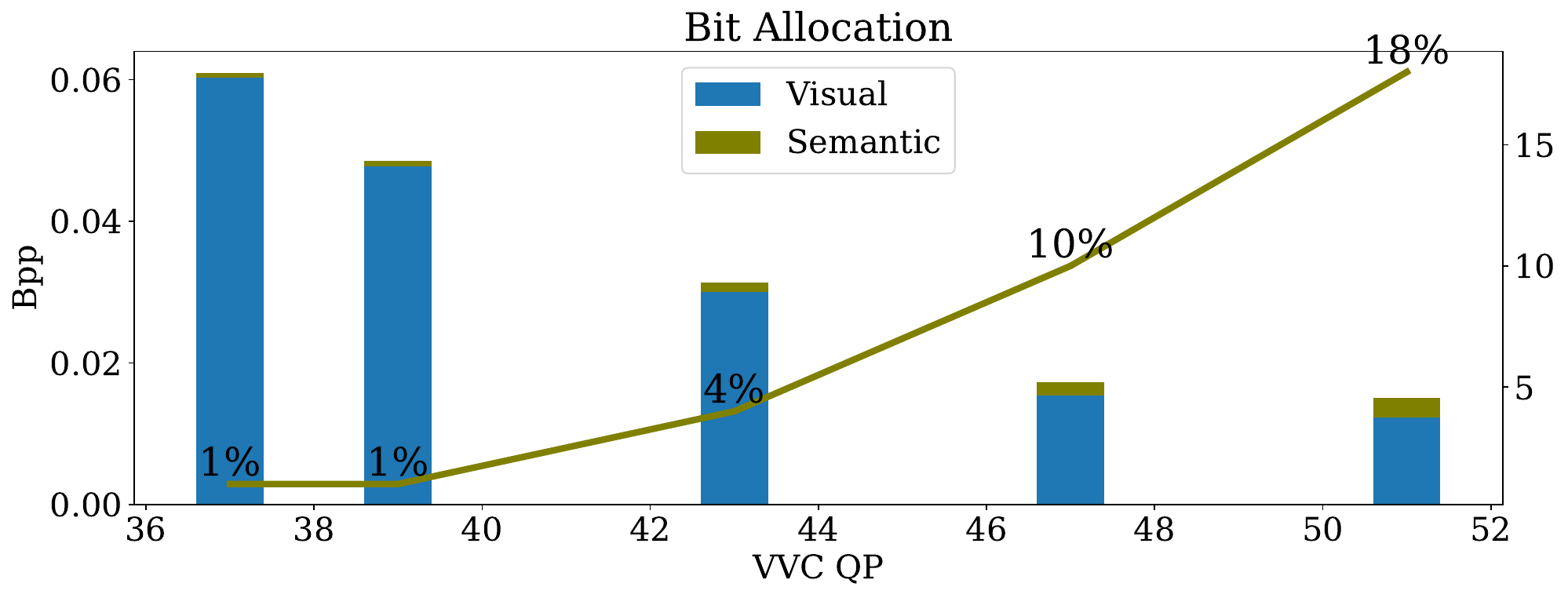}
 	\vspace{-2mm}
 	\caption {
 		Bit consumption of visual and residual semantic information in the SMC model.
 		The x-axis indicates the QP value of the base-layer VVC codec in SMC.
 	}
 	\vspace{-2mm}
 	\label{fig:bit_alloc}
 \end{figure}

 \textbf{Bit-allocation Analysis.}
 As shown in Figure~\ref{fig:bit_alloc}, the semantic stream is always of high compression efficiency.
 Moreover, the bit allocation strategy is adaptive to the QP of the lossy visual stream, although we do not introduce any explicit adaptive design and time-consuming online rate-distortion optimization (RDO) strategies like~\cite{lu2020content}.
 The proportion of semantic information has been decreased to about 1\% when QP is larger than 40 (the second column).
 This 1\% bitcost boots Top1 accuracy by 7\% on UCF101-TSM.
 
  \begin{figure}[!t]
 	\centering
 	\newcommand{\widthscalefive}{0.46}
 	\tabcolsep=-0mm
 	\includegraphics[width=\widthscalefive \textwidth]{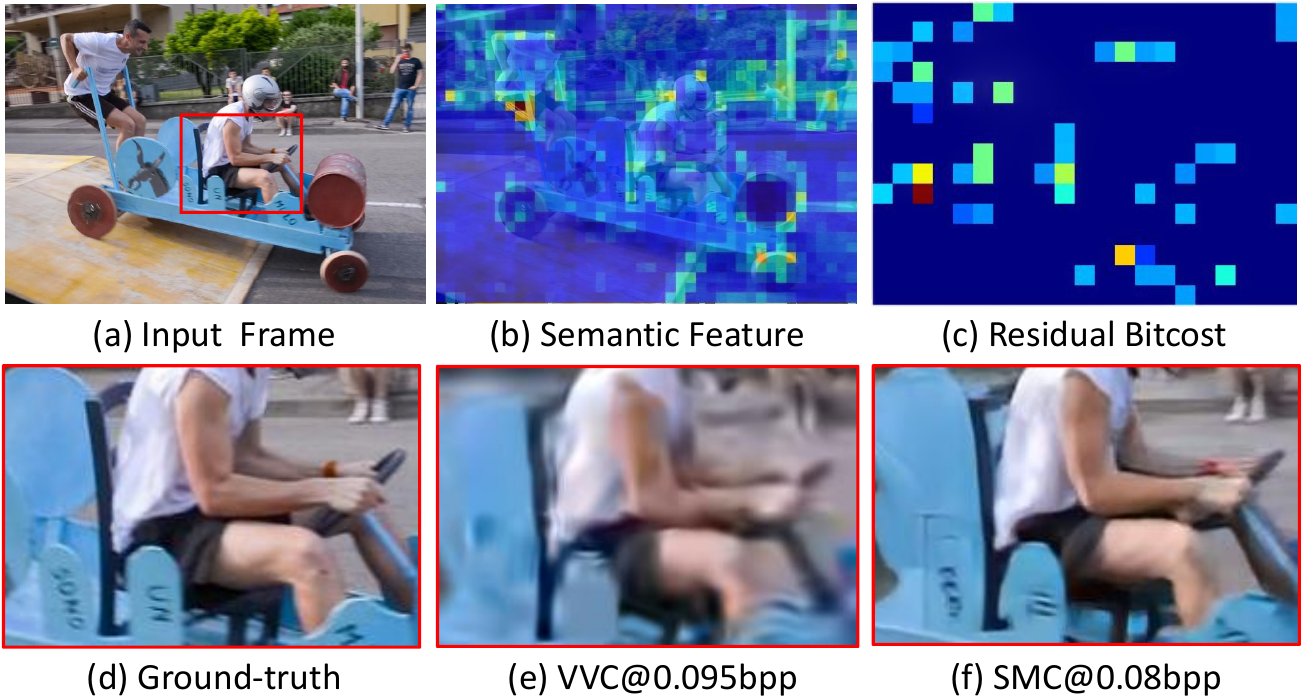}
 	\vspace{-2mm}
 	\caption {
 		Visualization of the semantic feature and the bitcost map of the residual semantics within SMC.
 		The red box region is zoomed in for more clearer visual comparison.
 	}
 	\label{fig:vis_semantic}
 	\vspace{-5mm}
 \end{figure}

 \textbf{Visualization of the Learned Semantics.}
 As shown in Figure~\ref{fig:vis_semantic}, the semantic feature extracted from the input frame is of high semantics and close to human perception.
 The activated regions are concentrated on the human body and the saliency objects, \textit{i.e.}, vehicle wheels and drum.
 We further visualize the bitcost map of the residual part, where the redundant part is removed by subtracting the semantics in the lossy video stream. 
 The residual bitcost map is quite sparse, and further concentrated on the AI task-interested regions.
 We also compare the VVC codec and our method in terms of qualitative results.
 As shown in (e) and (f) of Figure~\ref{fig:vis_semantic}, our method based on VVC (QP=51) demonstrates much clearer object structures and sharper edges than vanilla VVC (QP=47), while consuming few bits (0.08bpp \textit{v.s.} 0.095bpp).

\subsection{\textsf{Analysis on the Improved SMC++ Model}}
In this section, we conduct comprehensive experiments to study the effectiveness of the newly added components in the SMC++ model, specifically: (1) the masked motion learning objective and (2) the Blueprint-guided compression Transformer (Blue-Tr).
Considering that the SMC++ model requires more training iterations than the vanilla SMC model, we have accordingly increased the training iterations for the SMC model as well.
This adjustment gives rise to the SMC$^{adv}$ model, which serves as a more fair baseline.

 \begin{figure}[!t] 
	\centering
	\tabcolsep=-0mm
	\revisefigure{
		\begin{tabular}{ccc}
			\includegraphics[width=0.33 \linewidth]{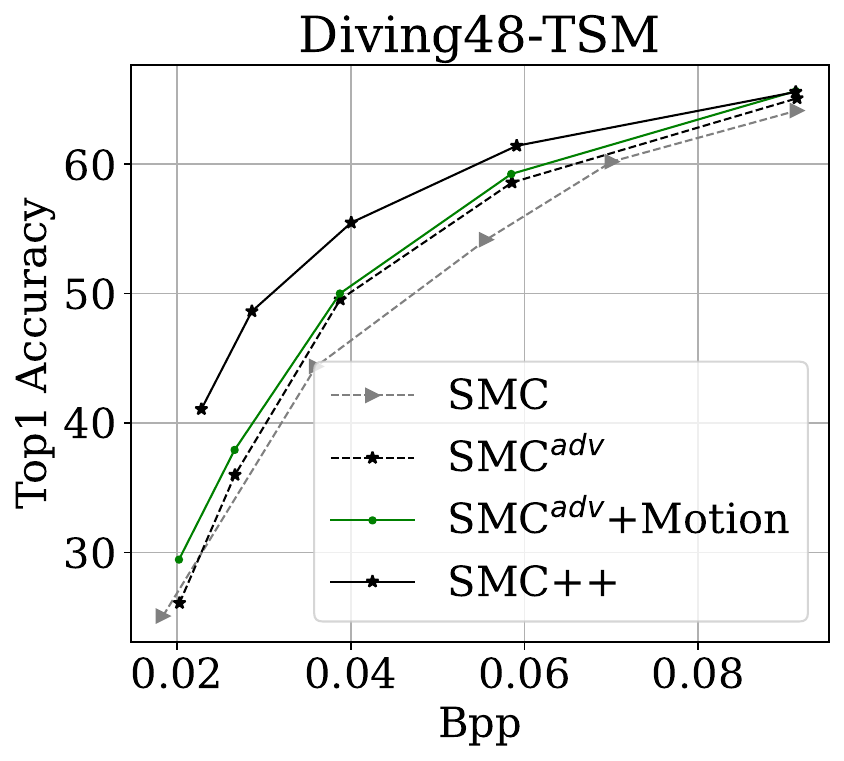}&
			\includegraphics[width=0.33 \linewidth]{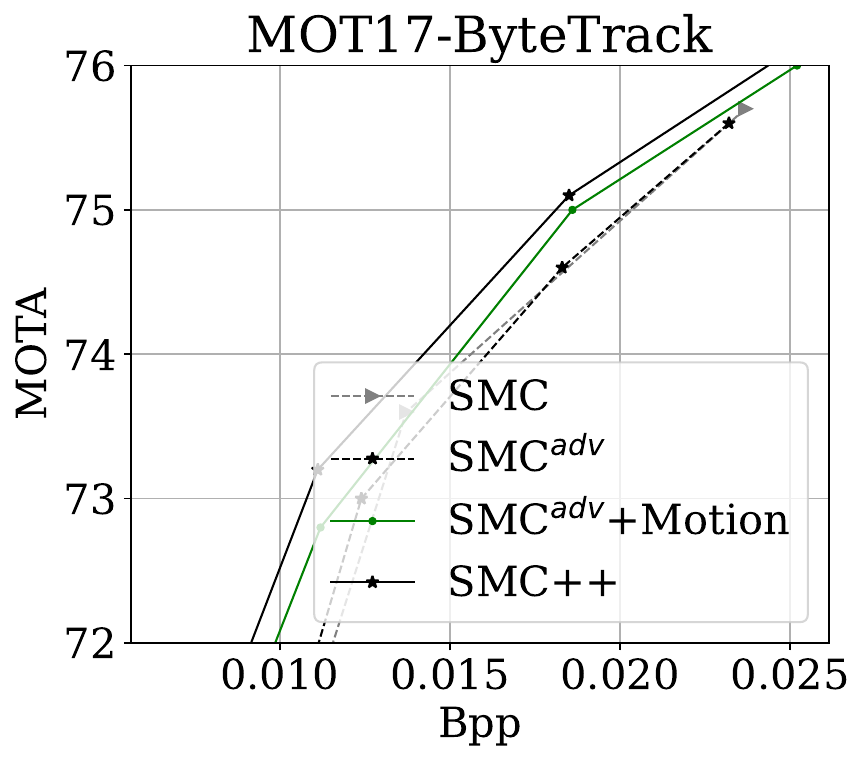}&
				\includegraphics[width=0.33 \linewidth]{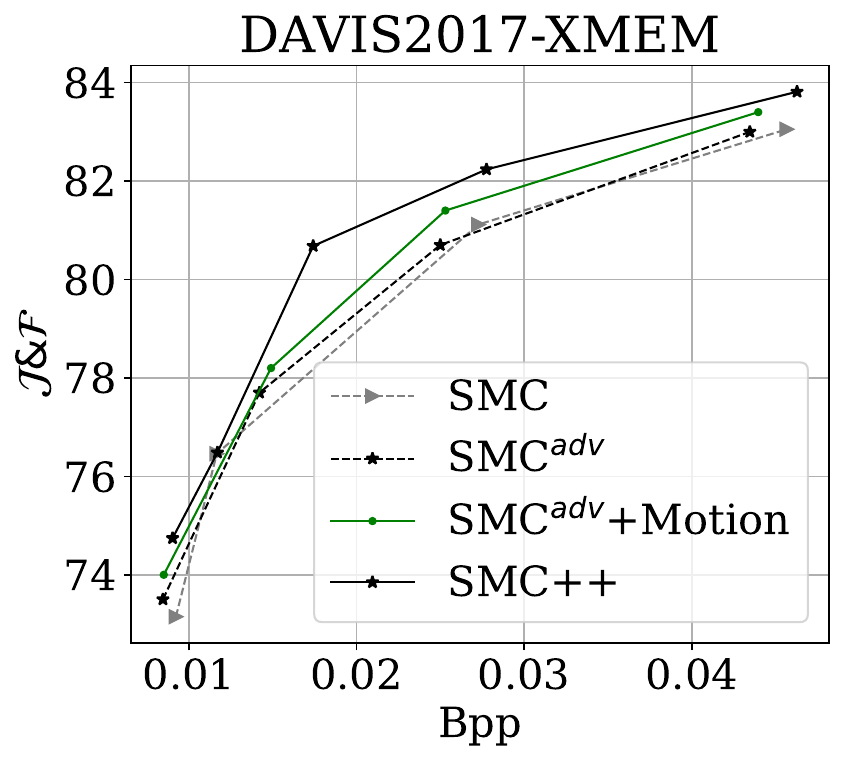}
			 \\
		\end{tabular}
}
\vspace{-2mm}
	\caption{
	\revise{Ablation study of SMC++ framework on action recognition, MOT and VOS tasks.
		`SMC$^{adv}$' shares the same details as the original SMC, but with extended training iterations to align with SMC++ for a fair comparison.
		`SMC$^{adv}$+Motion' model is trained with both masked appearance and motion modeling objectives.
		SMC++ model further improves `SMC$^{adv}$+Motion' model with Blue-Tr.}
	}
\vspace{-5mm}
	\label{fig:smcplus_ablation}
\end{figure}

\textbf{Effectiveness of the New Components.}
As shown in Figure~\ref{fig:smcplus_ablation}, we train different variant models and evaluate them on the Diving48-TSM setting.
First, we observe that by merely increasing the training iterations, neither changing the learning objective nor any network architecture, the resultant SMC$^{adv}$ model has noticeably outperformed the original SMC model in our previous conference paper~\cite{tian2023non}. For instance, we note a substantial improvement of approximately 2\% in performance at bitrate levels of 0.04bpp and 0.06bpp.
Then, after introducing the masked motion learning objective, the resultant SMC$^{adv}$+Motion model achieves a 3.2\% Top1 accuracy advantage over the SMC$^{adv}$ baseline, at the 0.02bpp level.
This is because the additional motion target enhances the temporal modeling capability of our approach, allowing it to better discern video instances with ambiguous actions.
Finally, the incorporation of our new Blue-Tr module leads to a notable enhancement, \textit{i.e.}, the resultant SMC++ model (SMC$^{adv}$+Motion+Blue-Tr) achieves approximately 9\% Top1 accuracy gain at the 0.03bpp bitrate level, when compared with the SMC$^{adv}$+Motion model.
This proves that the specially designed Blue-Tr is powerful at compressing semantic information.

\revise{
Furthermore, we conduct ablation studies on MOT and VOS tasks, to comprehensively validate the effectiveness of the masked motion modeling scheme and the Blue-Tr module.
As shown in Figure~\ref{fig:smcplus_ablation}, both the masked motion objective and the Blue-Tr module contribute to the superior performance on the VOS and MOT tasks. For instance, in the VOS task, adding the masked motion objective results in the SMC$^{adv}$+Motion model achieving 79.76\% in $\mathcal{J\&F}$ at 0.02bpp, which is 0.46\% higher than the 79.30\% attained by SMC$^{adv}$. Furthermore, by integrating the transformer-based Blue-Tr module, our final SMC++ model reaches 81.06\% in $\mathcal{J\&F}$ at 0.02bpp, marking a further improvement of 1.30\% over the SMC$^{adv}$+Motion model.

These results demonstrate that the two new modules introduced in SMC++ are always effective, across tasks of varying granularity, from classification to segmentation.
%
}

\textbf{Different Masked Motion Prediction Objectives.}
Furthermore, we analyze the influence of different masked motion learning targets.
We train three SMC variant models using three common motion representations as prediction targets, \textit{i.e.}, RGB difference map, optical flow, and feature correlation map.
RGB difference map is calculated by pixel-wise subtraction between each two consecutive frames.
The optical flow is obtained with the pre-trained RAFT model~\cite{teed2020raft}.
As for the calculation of the feature correlation map, we first extract the feature map of video frames with the ResNet18 network, then calculate the correlation map~\cite{dosovitskiy2015flownet} between the features of two consecutive frames.
Compared to the optical flow map, the correlation map has a coarser resolution, but depicts more semantics.

 \begin{figure}[!t] 
	\centering
	\newcommand{\widthscalefive}{0.46}
	\tabcolsep=1mm
	\scalebox{1}{
		\begin{tabular}{cc}
			\includegraphics[width=0.21 \textwidth]{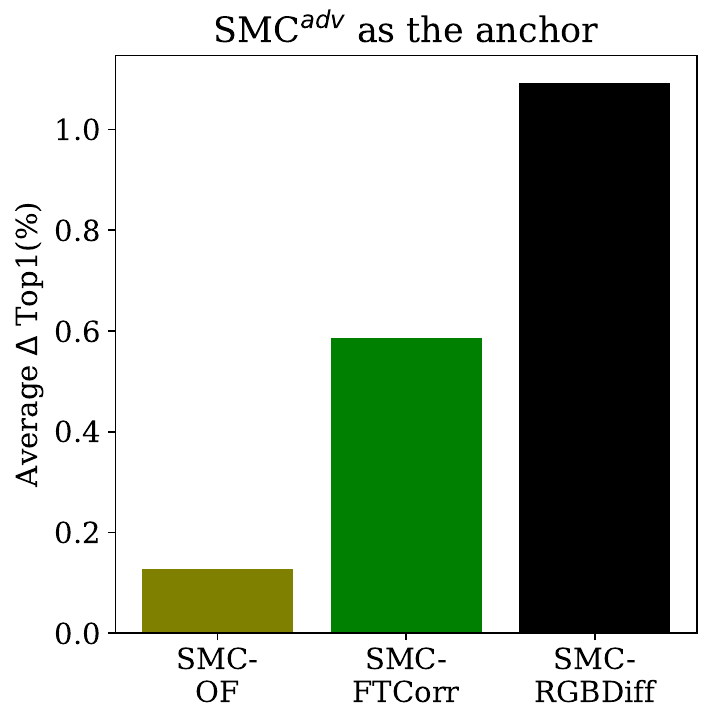}&
			\includegraphics[width=0.24 \textwidth]{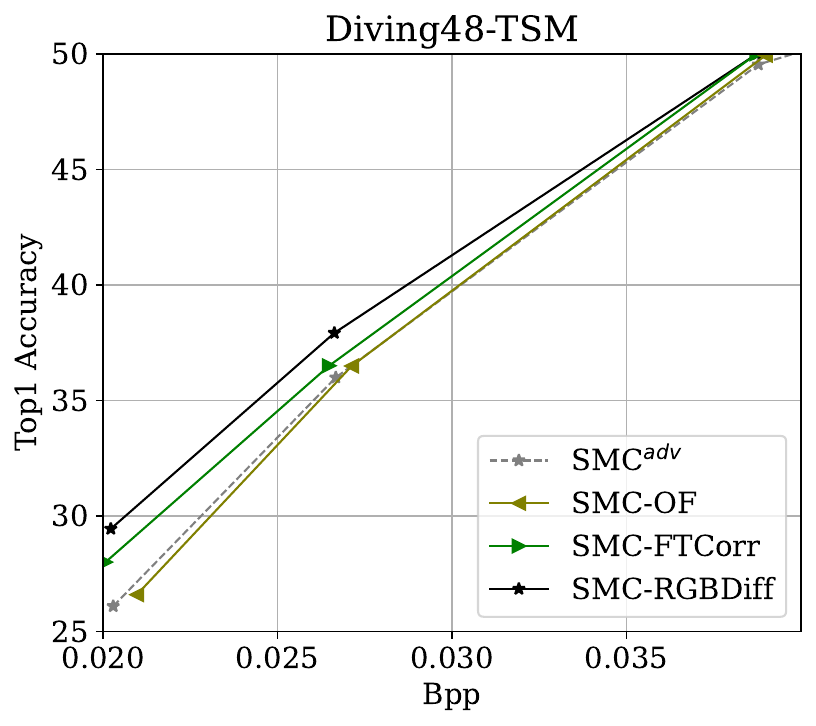} \\
			(a) &(b)
		\end{tabular}
	}
	\vspace{-2mm}
	\caption{
		Comparison of different masked motion targets.
	}
	\vspace{-5mm}
	\label{fig:ablation_smc_plus_motionpred}
\end{figure}

As shown in Figure~\ref{fig:ablation_smc_plus_motionpred} (a), introducing motion prediction branch to the MAE learning objective always leads to the performance improvement.
Compared to the computationally expensive feature-level matching operations in computing optical flow and correlation map, obtaining RGB difference map is more efficient.
However, it demonstrates the strongest result, over 1\% average Top1 accuracy improvement over the baseline SMC$^{adv}$ model, obviously superior to the gains 0.15\% and 0.6\% that are obtained by optical flow and feature correlation.
This is not surprising, as using frame difference as motion representation has effectively enhanced various video analysis tasks~\cite{tian2022ean}.

We further inspect the impact of different motion targets at various bit-rate levels, as shown in Figure~\ref{fig:ablation_smc_plus_motionpred} (b).
At lower bitrate levels such as 0.02bpp, the performance gain of RGB difference is amplified as over 3\% Top1 accuracy.
In contrast, the performance gain of adopting the correlation map is inferior, \textit{i.e.}, about 2\% Top1 accuracy.
When adopting the optical flow map, the performance is even hurt at low bitrates.
We conjecture the reason is that predicting the complex optical flow is non-trivial for the lightweight decoder in the MAE framework,
thus impeding the learning procedure.
At higher bitrate levels (e.g., 0.04bpp), the gains of all motion-enhanced models diminish, as appearance-based semantics are sufficiently discriminative.

\revise{
Finally, we investigate the balancing weight between masked pixel and motion prediction targets, by introducing the tunable hyperparameter $\gamma$ instead of the fixed 0.5, i.e.,
\begin{equation}
	\mathcal{L}_{MAE} = (1-\gamma) \mathcal{L}_{MAE-pixel} + \gamma \mathcal{L}_{MAE-motion}.
\end{equation}

As shown in Table~\ref{tab:ablation_motion_pixel_weight}, setting $\gamma=0.5$ consistently yields optimal or near-optimal performance across various bitrates. For example, at 0.06 bpp, the Top1 accuracy reaches 59.54\%, outperforming both $\gamma=0$ (58.87\%) and $\gamma=1$ (58.91\%). At lower bitrates, such as 0.025 bpp, the benefit of higher $\gamma$ values becomes evident, with $\gamma=0.5$ achieving 35.77\% and $\gamma=1$ achieving 36.13\%. This is likely because motion information is sparser and can be more efficiently compressed at lower bitrates.
}

\begin{table}[!thbp]
	\centering
	\caption{\revise{Comparison of different tunable balancing weights ($\gamma$) between the pixel-based and motion-based MAE learning target in the SMC$^{adv}$+Motion model.}}
	\vspace{-2mm}
	\label{tab:gamma_comparison}
	\tabcolsep=3mm
	\revisefigure{
	~\begin{tabular}{|c|c|c|c|c|}
		\hline
		\multirow{2}{*}{$\gamma$} & \multicolumn{4}{c|}{Diving48-TSM Top1 Accuracy (\%)} \\
		\cline{2-5}
		& @0.025bpp & @0.04bpp & @0.06bpp & @0.08bpp \\
		\hline
		0 & 33.40 & 50.10 & 58.87 & 62.83 \\
		\hline
		0.25 & 34.44 & 50.28 & 58.90 & 62.88 \\
		\hline
		0.5 & 35.77 & \textbf{50.59} & \textbf{59.54} & \textbf{63.45} \\
		\hline
		0.75 & 35.97 & 50.50 & 59.29 & 63.13 \\
		\hline
		1 & \textbf{36.13} & 50.37 & 58.91 & 62.73 \\
		\hline
	\end{tabular}
}
\vspace{-2mm}
	\label{tab:ablation_motion_pixel_weight}
\end{table}

\begin{figure}[!t] 
	\centering
	\tabcolsep=0.2mm
	\scalebox{1}{
	\hspace{-4mm}	\begin{tabular}{cc}
			\includegraphics[width=0.24 \textwidth]{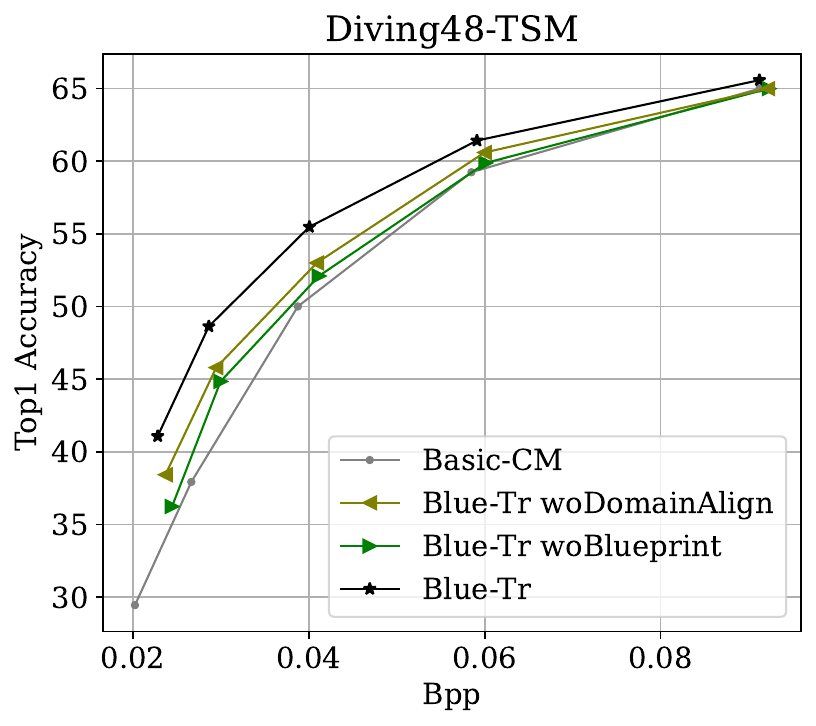}
			&
			\includegraphics[width=0.23 \textwidth]{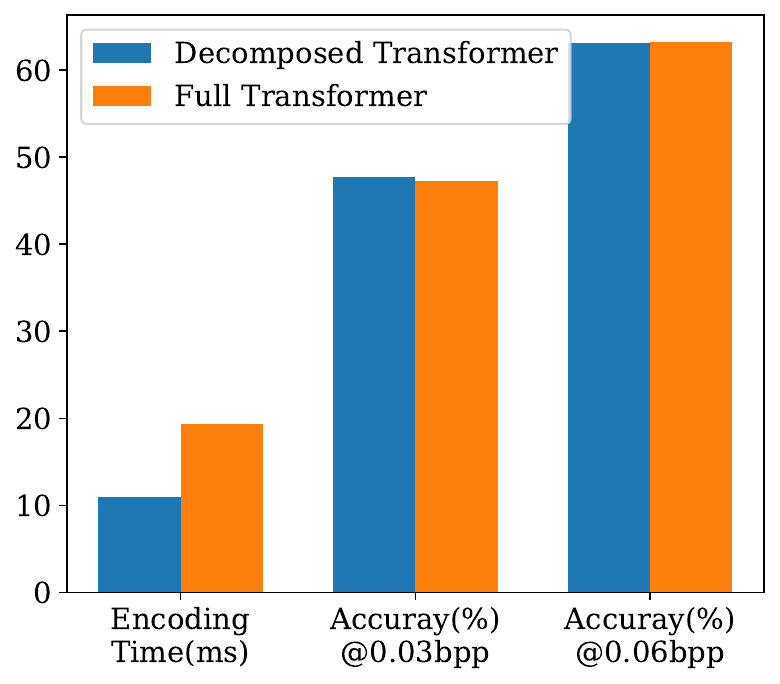}\\
			(a) & (b)
		\end{tabular}
	}
	\vspace{-2mm}
	\caption{
		(a) Ablation study of Blueprint-guide compression Transformer (Blue-Tr).
		(b) Comparison of the adopted decomposed Transformer and the vanilla full Transformer.
	}
	\vspace{-4mm}
	\label{fig:smcplus_ablation_entropy}
\end{figure}

\textbf{Ablation Study on Blue-Tr.} 
As shown in Figure~\ref{fig:smcplus_ablation_entropy} (a), all variant modules of Blue-Tr outperform the CNN-based basic compression module Basic-CM. This proves the advantage of Transformer over CNN in compressing highly abstract semantic information.
Furthermore, we note that removing the blueprint-based alignment, the resultant `Blue-Tr woBlueprint' model depicts a notable decrease in Top1 accuracy, \textit{i.e.}, about 5\% drop at a 0.04bpp bitrate. The reason is that, without explicit feature alignment, the Transformer still struggles to capture redundancies among various contextual features.

Moreover, to assess the impact of domain correction and motion compensation, we removed the Domain-Net from the Feature Aligner. The resulting model `Blue-Tr woDomainAlign' performs obviously worse than the full Blue-Tr, particularly at low bitrates. For instance, at 0.025bpp, there's a performance dip of over 3\%. This decline is attributed to the domain gap between the features of the lossy video (compressed by the base VVC codec) and our semantic features, which is further exacerbated by the severe semantic distortion of VVC in low-bitrate conditions.

Further, we investigate the effectiveness of the decomposed Transformer used in Blue-Tr. Figure~\ref{fig:smcplus_ablation_entropy} (b) demonstrates that our decomposition strategy enhances speed by about 2$\times$ compared to the standard Transformer (10.9ms vs. 20.2ms), while performance remains similar.

\revise{
We explore replacing full attention with Linear attention~\cite{katharopoulos2020transformers} to reduce attention complexity from quadratic to linear.  
As shown in Figure 6 of the supplementary material, the two methods perform similarly at higher bitrates ($>$0.06 bpp). However, at lower bitrates such as 0.03bpp, full attention substantially outperforms Linear attention by about 5\% accuracy, due to the latter's limited capability in global context modeling, which is critical for semantic tasks.  
}

Moreover, we explore how the blueprint semantic feature's dimension impacts system performance. As shown in Table~\ref{tab:blueprint_dim}, smaller dimensions of 32 and 64 yield poorer results due to insufficient information for effective feature alignment. Conversely, larger dimensions like 256 and 512, while not causing drastic performance declines, still lead to a noticeable drop. This is because the bitcost savings from blueprint guidance can't offset the cost of transmitting the blueprint feature itself.

\begin{table}[!t]
	\centering
	\caption{
		Comparison of different Blueprint feature dimensions, on the Diving48-TSM setting at 0.04bpp bitrate level.
	}
	\vspace{-2mm}
	\tabcolsep=3.5mm
	\begin{tabular}{|c|c|c|c|c|c|}
		\hline
		Dimension&  32&64&128&256& 512\\
		\hline
		\makecell[c]{ 
			Top1 (\%)}  &   53.11&54.02&\textbf{54.58}&54.26& 54.13\\
		\hline
	\end{tabular}
	\label{tab:blueprint_dim}
\end{table}

\revise{
Previous works~\cite{lin2020m,sheng2022temporal} have demonstrated that multi-frame reference improves coding performance. To evaluate whether our Blur-Tr module can effectively exploit multi-frame historical information, we incorporate long-term history by aligning multiple previous frames to the current frame using the Feature Aligner in the Blur-Tr module. The aligned features are fused through a residual block, providing a richer coding context compared to using a single previous frame.  
As shown in Table~\ref{tab:ablation_history_number}, incorporating additional previous frames consistently enhances performance, aligning with prior findings~\cite{lin2020m}. Future work will focus on developing more advanced multi-frame fusion strategies tailored to video semantics, further improving video semantic compression.
}

\begin{table}[!t]
	\vspace{-2mm}
	\centering
	\caption{
		\revise{Comparison of multiple previous frames as the coding context for SMC++, on Diving48-TSM at 0.04bpp.}
	}
	\vspace{-2mm}
	\tabcolsep=2.85mm
	\revisefigure{
	\begin{tabular}{|c|c|c|c|c|c|}
		\hline
		Previous Frames&  1&2&3&4& 5\\
		\hline
		\makecell[c]{ 
			Top1 (\%)}  &  54.58 & 54.83 &55.04& 55.07& 55.12\\
		\hline
	\end{tabular}
}
	\label{tab:ablation_history_number}
	\vspace{-3mm}
\end{table}

\revise{
Finally, we introduce SMC++$^{lite}$ to alleviate Blue-Tr's increased parameters and computational costs, by using a bottleneck design in the Blueprint-guided compression Transformer. It applies channel reduction before self-attention and expansion afterward. This design cuts computational complexity while maintaining modeling capability.
Table~\ref{tab_smc_lite} shows SMC++$^{lite}$ significantly outperforms SMC (7.28\% higher accuracy at 0.02bpp), while maintaining parameter efficiency comparable to SMC. Performance nearly matches the full SMC++ model (only 0.4\% lower at 0.02bpp). This confirms our gains come from the novel architecture of Blue-Tr, not just from more parameters.
}

\begin{table}[!thbp]
	\vspace{-2mm}
	\centering
	\caption{\revise{Performance comparison between SMC, SMC++$^{lite}$, and SMC++ models.}}
	\vspace{-2mm}
		\revisefigure{
	\begin{tabular}{|l|c|c|c|c|}
		\hline
		\multirow{2}{*}{Method} & \multirow{2}{*}{Parameters} & \multicolumn{3}{c|}{UCF101-TSM Top1 Accuracy (\%)} \\
		\cline{3-5}
		& & @0.02bpp & @0.04bpp & @0.06bpp \\
		\hline
		SMC & 48.23M & 76.48 & 86.46 & 89.11 \\
		\hline
		SMC++$^{lite}$ & 52.46M & 83.76 & 88.83 & 90.81 \\
		\hline
		SMC++ & 96.18M & 84.16 & 89.17 & 90.83 \\
		\hline
	\end{tabular}
}
	\vspace{-2mm}
	\label{tab_smc_lite}
\end{table}

 \revise{
\textbf{More Advanced Post-Enhancement Strategy.}
Both SMC and SMC++ adopt the GAN loss to ensure the photo-realism of decoded videos. Recently, stable diffusion~\cite{rombach2022high} has demonstrated advanced image generation capabilities. Consequently, we incorporate the GDP model~\cite{fei2023generative}, which leverages the rich generative priors of pre-trained diffusion models~\cite{rombach2022high}, to enhance image quality.  

As shown in Table~\ref{tab_diffusion_enhance}, the GDP-based enhancement substantially reduces FID (e.g., 27.92 at 0.02 bpp, compared to 76.12 without enhancement), indicating notable improvements in image quality. However, the accuracy improvement is marginal (e.g., 84.56\% vs. 84.16\%), suggesting that, for semantic compression task, preserving semantic information is more critical than enhancing visual quality.  
}

\begin{table}[!t]
	\centering
	\caption{\revise{Comparison of different approaches, in terms of Top1 Accuracy/Fr\'{e}chet Inception Distance (FID)~\cite{heusel2017gans}.}}
	\vspace{-2mm}
		\tabcolsep=3.2mm
		\revisefigure{
	\begin{tabular}{|c|c|c|c|}
		\hline
		\multirow{2}{*}{Method} & \multicolumn{3}{c|}{UCF101-TSM Top1 (\%)/FID} \\
		\cline{2-4}
		&@0.02bpp & @0.04bpp & @0.06bpp \\
		\hline
		SMC++  & 84.16/76.12 & 89.17/58.24 & 90.83/36.53 \\
		\hline
		SMC++(GDP) & 84.56/27.92 & 89.32/26.81 & 90.89/25.04 \\
		\hline
	\end{tabular}
}
\vspace{-3mm}
	\label{tab_diffusion_enhance}
\end{table}

\revise{
\textbf{Robustness Evaluation.}
As shown in Figure 7 of the supplementary material, we evaluate SMC++ under varying noise levels (5\%, 10\%, 20\%, and 50\%) on the Diving48-TSM setting.  
As noise intensity increases, accuracy naturally decreases, but the drop remains moderate, e.g., ~2\% with 10\% noise. Notably, SMC++ demonstrates robustness even at 20\% noise levels, likely due to its ability to filter out noise by concentrating on task-relevant features.
}

\subsection{Model Complexity and Running Time}
\label{sec:model_complexity}
As Table~\ref{tab:model_complexity} demonstrates, our basic SMC model includes fewer parameters than the recent learnable codec DCVC-FM, yet more parameters than the previous semantic coding approach VCS. VCS achieves strong feature representation with fewer parameters by employing complex dynamic convolution in its encoder, but this comes at the cost of longer encoding time (1023ms compared to SMC's 952ms).
Furthermore, in the VCS, the semantic extraction and compression modules are tightly decoupled within the encoder network. In contrast, our SMC model separately employs a simple standard Resnet18 for semantic extraction, while isolating the semantic compression process into a distinct module. SMC's simplicity and modular design make it a flexible and easily-extendable baseline.

\revise{
Our SMC++ model further introduces Transformer blocks, increasing the parameter count compared to SMC.
However, the execution times are not obviously affected. This is because the new compression module Blue-Tr in SMC++, operating on 32$\times$ downsampled feature maps, is extremely efficient. 
Despite DCVC-FM having fewer parameters than SMC++ (59.63M vs. 96.18M), its decoding time is longer (654ms vs. 253ms). This difference stems from their distinct designs: DCVC-FM preserves low-level details for superior Rate-PSNR performance, operating primarily in a $4\times$ downscaled feature space, while our method prioritizes semantic compression within a $32\times$ downscaled semantic feature space. This $32\times$ downscaling yields approximately 64 times greater computational efficiency than $4\times$ downscaling, allowing our approach to achieve faster decoding despite having more parameters. The decoding FLOPS comparison (1651G vs. 307G) in Table~\ref{tab:model_complexity} confirms this.

Finally, we provide the complexity and cost of SMC++*, which supports decoding pixel-wise details by appending a lightweight detail decoding network DR-Net to the SMC++ decoder. Compared to SMC++, it only introduces 2.19M parameters while achieving a decoding speed of 296ms, which remains much faster than DCVC-FM (654ms).
}


\begin{table}[!thbp]
	\centering
	\caption{
		\revise{Model complexity and running time of different approaches.
		$^\dag$ denotes the method adopting the VVEnC software as the base coding layer, where the coding time consumed by VVEnC is incorporated.
		SMC++* refers to the SMC++ model augmented with DR-Net and further fine-tuned for the PSNR metric.  
		All evaluations are performed on a machine equipped with a single Nvidia 2080Ti GPU and Intel(R) Xeon(R) Silver 4116 CPU @ 2.10GHz.
		The running time was determined by encoding/decoding 100 frames of 1080p resolution and subsequently calculating the average time required for a single frame.}
	}
	\vspace{-2mm}
	\tabcolsep=2.2mm
	\begin{tabular}{|c|c|c|c|c|}
		\hline
		Method&Parameters & \makecell{Encoding \\ Time} & \makecell{Decoding \\ Time} &\makecell{Decoding \\ FLOPS}\\
		\hline
		VVEnc15.0 & - & 652ms & 142ms &-  \\
		\hline
		DCVC-FM &59.63M & 833ms&654ms & 1651G \\
		\hline
		VCS$^\dag$ &28.60M & 1023ms&198ms & 147G \\
		\hline
		SMC$^\dag$ &48.23M & 952ms&235ms & 199G\\
		\hline
		SMC++$^\dag$ &96.18M & 977ms&253ms & 307G \\
			\hline
		SMC++*$^\dag$ &98.37M & 977ms&296ms& 368G \\
		\hline
	\end{tabular}
	\vspace{-3mm}
	\label{tab:model_complexity}
\end{table}

 \vspace{-4mm}
 \section{Conclusion, Limitation, and Future Work}
 \label{sec:conclusion}
 In this paper, we have focused on a novel unsupervised video semantic compression problem.
 We have proposed a simple baseline model SMC to better cope with this problem, which is equipped with a novel non-semantics suppressed MAE loss.
 We have also proposed an advanced model SMC++, which improves SMC from two aspects, (1) a masked-motion learning objective and (2) a Transformer-based compression module.
 Comprehensive experiments demonstrate our method
 achieves remarkable results on three video analysis tasks with seven datasets.
 
 One limitation is the learned semantics still rely on the training dataset consisting of natural images, which may not perform well in the professional field, such as medical image analysis.
 In the future, we plan to address this by devising the fast domain adaptation strategy.
Moreover, motivated by the powerful semantics contained in large multi-modality models, we will attempt to transfer their rich semantic representation to our video semantic compression problem.

\noindent \textbf{Acknowledgment.} This work was supported by National Natural Science Foundation of China (Grant No.62501337), Shanghai Artificial Intelligence Laboratory, National Natural Science
Foundation of China (Grant No.72293585), National Natural Science Foundation of China (Grant No.72293580), National Natural Science Foundation of China (Grant No.62225112), National Natural Science Foundation of China (Grant No.62571322) and Shanghai Eye
Disease Research Center (2022ZZ01003).

\bibliographystyle{IEEEtran}      
\vspace{-4mm}

{\scriptsize
\bibliography{mybib}

\begin{thebibliography}{100}
\providecommand{\url}[1]{#1}
\csname url@samestyle\endcsname
\providecommand{\newblock}{\relax}
\providecommand{\bibinfo}[2]{#2}
\providecommand{\BIBentrySTDinterwordspacing}{\spaceskip=0pt\relax}
\providecommand{\BIBentryALTinterwordstretchfactor}{4}
\providecommand{\BIBentryALTinterwordspacing}{\spaceskip=\fontdimen2\font plus
\BIBentryALTinterwordstretchfactor\fontdimen3\font minus
  \fontdimen4\font\relax}
\providecommand{\BIBforeignlanguage}[2]{{%
\expandafter\ifx\csname l@#1\endcsname\relax
\typeout{** WARNING: IEEEtran.bst: No hyphenation pattern has been}%
\typeout{** loaded for the language `#1'. Using the pattern for}%
\typeout{** the default language instead.}%
\else
\language=\csname l@#1\endcsname
\fi
#2}}
\providecommand{\BIBdecl}{\relax}
\BIBdecl

\bibitem{sullivan2012overview}
G.~J. Sullivan, J.-R. Ohm, W.-J. Han, and T.~Wiegand, ``Overview of the high
  efficiency video coding (hevc) standard,'' \emph{TCSVT}, 2012.

\bibitem{bross2021overview}
B.~Bross, Y.-K. Wang, Y.~Ye, S.~Liu, J.~Chen, G.~J. Sullivan, and J.-R. Ohm,
  ``Overview of the versatile video coding (vvc) standard and its
  applications,'' \emph{TCSVT}, 2021.

\bibitem{hu2021fvc}
Z.~Hu, G.~Lu, and D.~Xu, ``Fvc: A new framework towards deep video compression
  in feature space,'' in \emph{CVPR}, 2021.

\bibitem{li2021deep}
J.~Li, B.~Li, and Y.~Lu, ``Deep contextual video compression,'' \emph{NeurIPS},
  2021.

\bibitem{yi2021benchmarking}
C.~Yi, S.~Yang, H.~Li, Y.-p. Tan, and A.~Kot, ``Benchmarking the robustness of
  spatial-temporal models against corruptions,'' in \emph{NeurIPS}, 2021.

\bibitem{tanaka2022does}
T.~Tanaka, A.~Harell, and I.~V. Bajić, ``Does video compression impact
  tracking accuracy?'' in \emph{ISCAS}, 2022.

\bibitem{zhang2016joint}
X.~Zhang, S.~Ma, S.~Wang, X.~Zhang, H.~Sun, and W.~Gao, ``A joint compression
  scheme of video feature descriptors and visual content,'' \emph{TIP}, 2016.

\bibitem{chao2015keypoint}
J.~Chao and E.~Steinbach, ``Keypoint encoding for improved feature extraction
  from compressed video at low bitrates,'' \emph{TMM}, 2015.

\bibitem{baroffio2015hybrid}
L.~Baroffio, M.~Cesana, A.~Redondi, M.~Tagliasacchi, and S.~Tubaro, ``Hybrid
  coding of visual content and local image features,'' in \emph{ICIP}, 2015.

\bibitem{duan2018compact}
L.-Y. Duan, Y.~Lou, Y.~Bai, T.~Huang, W.~Gao, V.~Chandrasekhar, J.~Lin,
  S.~Wang, and A.~C. Kot, ``Compact descriptors for video analysis: The
  emerging mpeg standard,'' \emph{IEEE MultiMedia}, 2018.

\bibitem{duan2015overview}
L.-Y. Duan, V.~Chandrasekhar, J.~Chen, J.~Lin, Z.~Wang, T.~Huang, B.~Girod, and
  W.~Gao, ``Overview of the mpeg-cdvs standard,'' \emph{TIP}, 2015.

\bibitem{duan2013compact}
L.-Y. Duan, F.~Gao, J.~Chen, J.~Lin, and T.~Huang, ``Compact descriptors for
  mobile visual search and mpeg cdvs standardization,'' in \emph{ISCAS}, 2013.

\bibitem{galteri2018video}
L.~Galteri, M.~Bertini, L.~Seidenari, and A.~Del~Bimbo, ``Video compression for
  object detection algorithms,'' in \emph{ICPR}, 2018.

\bibitem{choi2018high}
H.~Choi and I.~V. Bajic, ``High efficiency compression for object detection,''
  in \emph{ICASSP}, 2018.

\bibitem{choi2018near}
H.~Choi and I.~V. Bajić, ``Near-lossless deep feature compression for
  collaborative intelligence,'' in \emph{MMSP}, 2018.

\bibitem{chen2019lossy}
Z.~Chen, K.~Fan, S.~Wang, L.-Y. Duan, W.~Lin, and A.~Kot, ``Lossy intermediate
  deep learning feature compression and evaluation,'' in \emph{ACM MM}, 2019.

\bibitem{chen2019toward}
Z.~Chen, K.~Fan, S.~Wang, L.-Y. Duan, W.~Lin, and A.~C. Kot, ``Toward
  intelligent sensing: Intermediate deep feature compression,'' \emph{TIP},
  2019.

\bibitem{singh2020end}
S.~Singh, S.~Abu-El-Haija, N.~Johnston, J.~Ballé, A.~Shrivastava, and
  G.~Toderici, ``End-to-end learning of compressible features,'' in
  \emph{ICIP}, 2020.

\bibitem{feng2022image}
R.~Feng, X.~Jin, Z.~Guo, R.~Feng, Y.~Gao, T.~He, Z.~Zhang, S.~Sun, and Z.~Chen,
  ``Image coding for machines with omnipotent feature learning,'' in
  \emph{ECCV}, 2022.

\bibitem{sheng2024vnvc}
X.~Sheng, L.~Li, D.~Liu, and H.~Li, ``Vnvc: A versatile neural video coding
  framework for efficient human-machine vision,'' \emph{TPAMI}, 2024.

\bibitem{liu2021semantics}
K.~Liu, D.~Liu, L.~Li, N.~Yan, and H.~Li, ``Semantics-to-signal scalable image
  compression with learned revertible representations,'' \emph{IJCV}, 2021.

\bibitem{yang2021towards}
S.~Yang, Y.~Hu, W.~Yang, L.-Y. Duan, and J.~Liu, ``Towards coding for human and
  machine vision: Scalable face image coding,'' \emph{TMM}, 2021.

\bibitem{yan2021sssic}
N.~Yan, C.~Gao, D.~Liu, H.~Li, L.~Li, and F.~Wu, ``Sssic: semantics-to-signal
  scalable image coding with learned structural representations,'' \emph{TIP},
  2021.

\bibitem{choi2021latent}
H.~Choi and I.~V. Bajić, ``Latent-space scalability for multi-task
  collaborative intelligence,'' in \emph{ICIP}, 2021.

\bibitem{choi2022scalable11}
{Choi, Hyomin and Bajić, Ivan V}, ``Scalable image coding for humans and
  machines,'' \emph{TIP}, 2022.

\bibitem{sun2020semantic}
S.~Sun, T.~He, and Z.~Chen, ``Semantic structured image coding framework for
  multiple intelligent applications,'' \emph{TCSVT}, 2020.

\bibitem{lin2023deepsvc}
H.~Lin, B.~Chen, Z.~Zhang, J.~Lin, X.~Wang, and T.~Zhao, ``Deepsvc: Deep
  scalable video coding for both machine and human vision,'' in \emph{ACM MM},
  2023.

\bibitem{bhaskaran1997image}
V.~Bhaskaran and K.~Konstantinides, \emph{Image and video compression
  standards: algorithms and architectures}.\hskip 1em plus 0.5em minus
  0.4em\relax Springer Science \& Business Media, 1997.

\bibitem{wang2004image}
Z.~Wang, A.~C. Bovik, H.~R. Sheikh, and E.~P. Simoncelli, ``Image quality
  assessment: from error visibility to structural similarity,'' \emph{TIP},
  2004.

\bibitem{he2022masked}
K.~He, X.~Chen, S.~Xie, Y.~Li, P.~Dollár, and R.~Girshick, ``Masked
  autoencoders are scalable vision learners,'' in \emph{CVPR}, 2022.

\bibitem{tong2022videomae}
Z.~Tong, Y.~Song, J.~Wang, and L.~Wang, ``Videomae: Masked autoencoders are
  data-efficient learners for self-supervised video pre-training,''
  \emph{NeurIPS}, 2022.

\bibitem{dong2022bootstrapped}
X.~Dong, J.~Bao, T.~Zhang, D.~Chen, W.~Zhang, L.~Yuan, D.~Chen, F.~Wen, and
  N.~Yu, ``Bootstrapped masked autoencoders for vision bert pretraining,'' in
  \emph{ECCV}, 2022.

\bibitem{simonyan2014two}
K.~Simonyan and A.~Zisserman, ``Two-stream convolutional networks for action
  recognition in videos,'' \emph{NeurIPS}, 2014.

\bibitem{vaswani2017attention}
A.~Vaswani, N.~Shazeer, N.~Parmar, J.~Uszkoreit, L.~Jones, A.~N. Gomez,
  {\L}.~Kaiser, and I.~Polosukhin, ``Attention is all you need,''
  \emph{NeurIPS}, 2017.

\bibitem{tian2023non}
Y.~Tian, G.~Lu, G.~Zhai, and Z.~Gao, ``Non-semantics suppressed mask learning
  for unsupervised video semantic compression,'' in \emph{CVPR}, 2023.

\bibitem{wiegand2003overview}
T.~Wiegand, G.~J. Sullivan, G.~Bjontegaard, and A.~Luthra, ``Overview of the
  h.264/avc video coding standard,'' \emph{TCSVT}, 2003.

\bibitem{lu2019dvc}
G.~Lu, W.~Ouyang, D.~Xu, X.~Zhang, C.~Cai, and Z.~Gao, ``Dvc: An end-to-end
  deep video compression framework,'' in \emph{CVPR}, 2019.

\bibitem{li2022hybrid}
J.~Li, B.~Li, and Y.~Lu, ``Hybrid spatial-temporal entropy modelling for neural
  video compression,'' in \emph{ACM MM}, 2022.

\bibitem{li2023neural1}
{Li, Jiahao and Li, Bin and Lu, Yan}, ``Neural video compression with diverse
  contexts,'' in \emph{CVPR}, 2023.

\bibitem{li2024neural}
J.~Li, B.~Li, and Y.~Lu, ``Neural video compression with feature modulation,''
  in \emph{CVPR}, 2024.

\bibitem{hu2025vrvvc}
Q.~Hu, H.~Zhong, Z.~Zheng, X.~Zhang, Z.~Cheng, L.~Song, G.~Zhai, and Y.~Wang,
  ``Vrvvc: Variable-rate nerf-based volumetric video compression,'' in
  \emph{AAAI}, 2025.

\bibitem{hu2025varfvv}
Q.~Hu, Q.~He, H.~Zhong, G.~Lu, X.~Zhang, G.~Zhai, and Y.~Wang, ``Varfvv:
  View-adaptive real-time interactive free-view video streaming with edge
  computing,'' \emph{JSAC}, 2025.

\bibitem{hu20254dgc}
Q.~Hu, Z.~Zheng, H.~Zhong, S.~Fu, L.~Song, X.~Zhang, G.~Zhai, and Y.~Wang,
  ``4dgc: Rate-aware 4d gaussian compression for efficient streamable
  free-viewpoint video,'' in \emph{CVPR}, 2025.

\bibitem{wang2003multiscale}
Z.~Wang, E.~P. Simoncelli, and A.~C. Bovik, ``Multiscale structural similarity
  for image quality assessment,'' in \emph{Asilomar}, 2003.

\bibitem{yang2021perceptual}
R.~Yang, R.~Timofte, and L.~Van~Gool, ``Perceptual learned video compression
  with recurrent conditional gan.'' in \emph{IJCAI}, 2022.

\bibitem{tian2024free}
Y.~Tian, G.~Lu, and G.~Zhai, ``Free-vsc: Free semantics from visual foundation
  models for unsupervised video semantic compression,'' in \emph{ECCV}, 2024.

\bibitem{chen2025just}
Z.~Chen, Y.~Tian, Y.~Sun, W.~Sun, Z.~Zhang, W.~Lin, G.~Zhai, and W.~Zhang,
  ``Just noticeable difference for large multimodal models,'' \emph{arXiv},
  2025.

\bibitem{aibench}
Z.~Zhang \emph{et~al.}, ``Aibench: Towards trustworthy evaluation under the
  45° law,'' \url{https://aiben.ch/}, 2025.

\bibitem{zhang2025lmmsurvey}
------, ``Large multimodal models evaluation: A survey,''
  \url{https://github.com/aiben-ch/LMM-Evaluation-Survey}, 2025, project Page:
  AIBench, available online.

\bibitem{li2025image}
C.~Li, Y.~Tian, X.~Ling, Z.~Zhang, H.~Duan, H.~Wu, Z.~Jia, X.~Liu, X.~Min,
  G.~Lu \emph{et~al.}, ``Image quality assessment: From human to machine
  preference,'' in \emph{CVPR}, 2025.

\bibitem{chen2024gaia}
Z.~Chen, W.~Sun, Y.~Tian, J.~Jia, Z.~Zhang, W.~Jiarui, R.~Huang, X.~Min,
  G.~Zhai, and W.~Zhang, ``Gaia: Rethinking action quality assessment for
  ai-generated videos,'' \emph{NeurIPS}, 2024.

\bibitem{li2025r}
C.~Li, J.~Zhang, Z.~Zhang, H.~Wu, Y.~Tian, W.~Sun, G.~Lu, X.~Min, X.~Liu,
  W.~Lin \emph{et~al.}, ``R-bench: Are your large multimodal model robust to
  real-world corruptions?'' \emph{IEEE JSTSP}, 2025.

\bibitem{huang2021visual}
Z.~Huang, C.~Jia, S.~Wang, and S.~Ma, ``Visual analysis motivated
  rate-distortion model for image coding,'' in \emph{ICME}, 2021.

\bibitem{cai2021novel}
Q.~Cai, Z.~Chen, D.~O. Wu, S.~Liu, and X.~Li, ``A novel video coding strategy
  in hevc for object detection,'' \emph{TCSVT}, 2021.

\bibitem{zhang2021just}
Q.~Zhang, S.~Wang, X.~Zhang, S.~Ma, and W.~Gao, ``Just recognizable distortion
  for machine vision oriented image and video coding,'' \emph{IJCV}, 2021.

\bibitem{veselov2021hybrid}
A.~I. Veselov, H.~Chen, F.~Romano, Z.~Zhijie, and M.~R. Gilmutdinov, ``Hybrid
  video and feature coding and decoding,'' 2021.

\bibitem{duan2020video}
L.~Duan, J.~Liu, W.~Yang, T.~Huang, and W.~Gao, ``Video coding for machines: A
  paradigm of collaborative compression and intelligent analytics,''
  \emph{TIP}, 2020.

\bibitem{hu2020towards}
Y.~Hu, S.~Yang, W.~Yang, L.-Y. Duan, and J.~Liu, ``Towards coding for human and
  machine vision: A scalable image coding approach,'' in \emph{ICME}, 2020.

\bibitem{duan2022jpd}
S.~Duan, H.~Chen, and J.~Gu, ``Jpd-se: High-level semantics for joint
  perception-distortion enhancement in image compression,'' \emph{TIP}, 2022.

\bibitem{dubois2021lossy}
Y.~Dubois, B.~Bloem-Reddy, K.~Ullrich, and C.~J. Maddison, ``Lossy compression
  for lossless prediction,'' \emph{NeurIPS}, 2021.

\bibitem{he2020momentum}
K.~He, H.~Fan, Y.~Wu, S.~Xie, and R.~Girshick, ``Momentum contrast for
  unsupervised visual representation learning,'' in \emph{CVPR}, 2020.

\bibitem{tian2025medical}
Y.~Tian, S.~Wang, and G.~Zhai, ``Medical manifestation-aware
  de-identification,'' in \emph{AAAI}, 2025.

\bibitem{tian2025semantics}
Y.~Tian, S.~Wang, R.~Zhang, Z.~Chen, Y.~Jiang, C.~Li, X.~Zhu, F.~Yan, Q.~Hu,
  X.~Wang \emph{et~al.}, ``Semantics versus identity: A divide-and-conquer
  approach towards adjustable medical image de-identification,'' \emph{ICCV},
  2025.

\bibitem{tian2025towards}
Y.~Tian, K.~Ji, R.~Zhang, Y.~Jiang, C.~Li, X.~Wang, and G.~Zhai, ``Towards
  all-in-one medical image re-identification,'' in \emph{CVPR}, 2025.

\bibitem{yuan2025rofi}
Y.~Tian \emph{et~al.}, ``Rofi: A deep learning-based ophthalmic sign-preserving
  and reversible patient face anonymizer,'' \emph{NPJ Digital Medicine}, 2025.

\bibitem{chen2024cross}
H.~Chen, Z.~Qu, Y.~Tian, N.~Jiang, Y.~Qin, J.~Gao, R.~Zhang, Y.~Ma, Z.~Jin, and
  G.~Zhai, ``A cross-temporal multimodal fusion system based on deep learning
  for orthodontic monitoring,'' \emph{Computers in Biology and Medicine}, 2024.

\bibitem{jiang2025advancing}
Y.~Jiang, P.~Zhang, D.~Yang, Y.~Tian, H.~Lin, and X.~Wang, ``Advancing
  generalizable tumor segmentation with anomaly-aware open-vocabulary attention
  maps and frozen foundation diffusion models,'' in \emph{CVPR}, 2025.

\bibitem{yi2022task}
X.~Yi, H.~Wang, S.~Kwong, and C.-C.~J. Kuo, ``Task-driven video compression for
  humans and machines: Framework design and optimization,'' \emph{TMM}, 2022.

\bibitem{ge2024task}
X.~Ge, J.~Luo, X.~Zhang, T.~Xu, G.~Lu, D.~He, J.~Geng, Y.~Wang, J.~Zhang, and
  H.~Qin, ``Task-aware encoder control for deep video compression,'' in
  \emph{CVPR}, 2024.

\bibitem{zhang2024video}
P.~Zhang, J.~Li, K.~Chen, M.~Wang, L.~Xu, H.~Li, N.~Sebe, S.~Kwong, and
  S.~Wang, ``When video coding meets multimodal large language models: A
  unified paradigm for video coding,'' \emph{arXiv}, 2024.

\bibitem{huang2022hmfvc}
Z.~Huang, C.~Jia, S.~Wang, and S.~Ma, ``Hmfvc: A human-machine friendly video
  compression scheme,'' \emph{TCSVT}, 2022.

\bibitem{tian2022coding}
Y.~Tian, G.~Lu, Y.~Yan, G.~Zhai, L.~Chen, and Z.~Gao, ``A coding framework and
  benchmark towards low-bitrate video understanding,'' \emph{TPAMI}, 2024.

\bibitem{choi2020task}
J.~Choi and B.~Han, ``Task-aware quantization network for jpeg image
  compression,'' in \emph{ECCV}, 2020.

\bibitem{akbari2019dsslic}
M.~Akbari, J.~Liang, and J.~Han, ``Dsslic: Deep semantic segmentation-based
  layered image compression,'' in \emph{ICASSP}, 2019.

\bibitem{huang2021deep}
D.~Huang, X.~Tao, F.~Gao, and J.~Lu, ``Deep learning-based image semantic
  coding for semantic communications,'' in \emph{GLOBECOM}, 2021.

\bibitem{chang2019layered}
J.~Chang, Q.~Mao, Z.~Zhao, S.~Wang, S.~Wang, H.~Zhu, and S.~Ma, ``Layered
  conceptual image compression via deep semantic synthesis,'' in \emph{ICIP},
  2019.

\bibitem{chang2022conceptual}
J.~Chang, Z.~Zhao, C.~Jia, S.~Wang, L.~Yang, Q.~Mao, J.~Zhang, and S.~Ma,
  ``Conceptual compression via deep structure and texture synthesis,''
  \emph{TIP}, 2022.

\bibitem{chang2021thousand}
J.~Chang, Z.~Zhao, L.~Yang, C.~Jia, J.~Zhang, and S.~Ma, ``Thousand to one:
  Semantic prior modeling for conceptual coding,'' in \emph{ICME}, 2021.

\bibitem{chang2022consistency}
J.~Chang, J.~Zhang, Y.~Xu, J.~Li, S.~Ma, and W.~Gao, ``Consistency-contrast
  learning for conceptual coding,'' in \emph{ACM MM}, 2022.

\bibitem{wang2019fast}
S.~Wang, H.~Lu, and Z.~Deng, ``Fast object detection in compressed video,'' in
  \emph{ICCV}, 2019.

\bibitem{li2022end}
C.~Li, X.~Wang, L.~Wen, D.~Hong, T.~Luo, and L.~Zhang, ``End-to-end compressed
  video representation learning for generic event boundary detection,'' in
  \emph{CVPR}, 2022.

\bibitem{wu2018compressed}
C.-Y. Wu, M.~Zaheer, H.~Hu, R.~Manmatha, A.~J. Smola, and P.~Krähbühl,
  ``Compressed video action recognition,'' in \emph{CVPR}, 2018.

\bibitem{shou2019dmc}
Z.~Shou, X.~Lin, Y.~Kalantidis, L.~Sevilla-Lara, M.~Rohrbach, S.-F. Chang, and
  Z.~Yan, ``Dmc-net: Generating discriminative motion cues for fast compressed
  video action recognition,'' in \emph{CVPR}, 2019.

\bibitem{carreira2017quo}
J.~Carreira and A.~Zisserman, ``Quo vadis, action recognition? a new model and
  the kinetics dataset,'' in \emph{CVPR}, 2017.

\bibitem{tian2022ean}
Y.~Tian, Y.~Yan, G.~Zhai, G.~Guo, and Z.~Gao, ``Ean: Event adaptive network for
  enhanced action recognition,'' \emph{IJCV}, 2022.

\bibitem{tian2020self}
Y.~Tian, Z.~Che, W.~Bao, G.~Zhai, and Z.~Gao, ``Self-supervised motion
  representation via scattering local motion cues,'' in \emph{ECCV}, 2020.

\bibitem{khatoonabadi2012video}
S.~H. Khatoonabadi and I.~V. Bajic, ``Video object tracking in the compressed
  domain using spatio-temporal markov random fields,'' \emph{TIP}, 2012.

\bibitem{chen2020improved}
X.~Chen, H.~Fan, R.~Girshick, and K.~He, ``Improved baselines with momentum
  contrastive learning,'' \emph{arXiv}, 2020.

\bibitem{huang2023mgmae}
B.~Huang, Z.~Zhao, G.~Zhang, Y.~Qiao, and L.~Wang, ``Mgmae: Motion guided
  masking for video masked autoencoding,'' in \emph{ICCV}, 2023.

\bibitem{sun2023masked}
X.~Sun, P.~Chen, L.~Chen, C.~Li, T.~H. Li, M.~Tan, and C.~Gan, ``Masked motion
  encoding for self-supervised video representation learning,'' in \emph{CVPR},
  2023.

\bibitem{yang2024motionmae}
H.~Yang, D.~Huang, B.~Wen, J.~Wu, H.~Yao, Y.~Jiang, X.~Zhu, and Z.~Yuan,
  ``Motionmae: Self-supervised video representation learning with motion-aware
  masked auto encoders,'' 2024.

\bibitem{song2022takes}
Y.~Song, M.~Yang, W.~Wu, D.~He, F.~Li, and J.~Wang, ``It takes two: Masked
  appearance-motion modeling for self-supervised video transformer
  pre-training,'' \emph{arXiv}, 2022.

\bibitem{zhao2024masa}
W.~Zhao, H.~Hu, W.~Zhou, Y.~Mao, M.~Wang, and H.~Li, ``Masa: Motion-aware
  masked autoencoder with semantic alignment for sign language recognition,''
  \emph{TCSVT}, 2024.

\bibitem{he2016deep}
K.~He, X.~Zhang, S.~Ren, and J.~Sun, ``Deep residual learning for image
  recognition,'' in \emph{CVPR}, 2016.

\bibitem{reynolds2009gaussian}
D.~A. Reynolds \emph{et~al.}, ``Gaussian mixture models,'' \emph{Encyclopedia
  of Biometrics}, 2009.

\bibitem{papoulis2002probability}
A.~Papoulis and S.~U. Pillai, \emph{Probability, random variables and
  stochastic processes}, 2002.

\bibitem{balle2018variational}
J.~Ball{\'e}, D.~Minnen, S.~Singh, S.~J. Hwang, and N.~Johnston, ``Variational
  image compression with a scale hyperprior,'' 2018.

\bibitem{cheng2020learned}
Z.~Cheng, H.~Sun, M.~Takeuchi, and J.~Katto, ``Learned image compression with
  discretized gaussian mixture likelihoods and attention modules,'' in
  \emph{CVPR}, 2020.

\bibitem{chen2021exploring}
X.~Chen and K.~He, ``Exploring simple siamese representation learning,'' in
  \emph{CVPR}, 2021.

\bibitem{esser2021taming}
P.~Esser, R.~Rombach, and B.~Ommer, ``Taming transformers for high-resolution
  image synthesis,'' in \emph{CVPR}, 2021.

\bibitem{zhang2018unreasonable}
R.~Zhang, P.~Isola, A.~A. Efros, E.~Shechtman, and O.~Wang, ``The unreasonable
  effectiveness of deep features as a perceptual metric,'' in \emph{CVPR},
  2018.

\bibitem{balle2016end}
J.~Balle, V.~Laparra, and E.~P. Simoncelli, ``End-to-end optimized image
  compression,'' in \emph{ICLR}, 2017.

\bibitem{ahmed2020k}
M.~Ahmed, R.~Seraj, and S.~M.~S. Islam, ``The k-means algorithm: A
  comprehensive survey and performance evaluation,'' \emph{Electronics}, 2020.

\bibitem{cover1999elements}
T.~M. Cover, \emph{Elements of information theory}.\hskip 1em plus 0.5em minus
  0.4em\relax John Wiley \& Sons, 1999.

\bibitem{cover1991entropy}
T.~M. Cover, J.~A. Thomas \emph{et~al.}, ``Entropy, relative entropy and mutual
  information,'' \emph{Elements of information theory}, 1991.

\bibitem{poole2019variational}
B.~Poole, S.~Ozair, A.~Van Den~Oord, A.~Alemi, and G.~Tucker, ``On variational
  bounds of mutual information,'' in \emph{ICML}, 2019.

\bibitem{hartigan1979algorithm}
J.~A. Hartigan and M.~A. Wong, ``Algorithm as 136: A k-means clustering
  algorithm,'' \emph{J. R. Stat. Soc. Series C}, 1979.

\bibitem{van2008visualizing}
L.~Van~der Maaten and G.~Hinton, ``Visualizing data using t-sne,'' \emph{JMLR},
  2008.

\bibitem{hu2020coarse}
Y.~Hu, W.~Yang, and J.~Liu, ``Coarse-to-fine hyper-prior modeling for learned
  image compression,'' in \emph{AAAI}, 2020.

\bibitem{hu2022coarse}
Z.~Hu, G.~Lu, J.~Guo, S.~Liu, W.~Jiang, and D.~Xu, ``Coarse-to-fine deep video
  coding with hyperprior-guided mode prediction,'' in \emph{CVPR}, 2022.

\bibitem{chen2017sca}
L.~Chen, H.~Zhang, J.~Xiao, L.~Nie, J.~Shao, W.~Liu, and T.-S. Chua, ``Sca-cnn:
  Spatial and channel-wise attention in convolutional networks for image
  captioning,'' in \emph{CVPR}, 2017.

\bibitem{dai2017deformable}
J.~Dai, H.~Qi, Y.~Xiong, Y.~Li, G.~Zhang, H.~Hu, and Y.~Wei, ``Deformable
  convolutional networks,'' in \emph{ICCV}, 2017.

\bibitem{xu2015empirical}
B.~Xu, N.~Wang, T.~Chen, and M.~Li, ``Empirical evaluation of rectified
  activations in convolutional network,'' \emph{arXiv}, 2015.

\bibitem{mentzer2022vct}
F.~Mentzer, G.~Toderici, D.~Minnen, S.~J. Hwang, S.~Caelles, M.~Lucic, and
  E.~Agustsson, ``Vct: a video compression transformer,'' in \emph{NeurIPS},
  2022.

\bibitem{soomro2012ucf101}
K.~Soomro, A.~R. Zamir, and M.~Shah, ``Ucf101: A dataset of 101 human actions
  classes from videos in the wild,'' \emph{arXiv}, 2012.

\bibitem{kuehne2011hmdb}
H.~Kuehne, H.~Jhuang, E.~Garrote, T.~Poggio, and T.~Serre, ``Hmdb: a large
  video database for human motion recognition,'' in \emph{ICCV}, 2011.

\bibitem{li2018resound}
Y.~Li, Y.~Li, and N.~Vasconcelos, ``Resound: Towards action recognition without
  representation bias,'' in \emph{ECCV}, 2018.

\bibitem{milan2016mot16}
A.~Milan, L.~Leal-Taixé, I.~Reid, S.~Roth, and K.~Schindler, ``Mot16: A
  benchmark for multi-object tracking,'' \emph{arXiv}, 2016.

\bibitem{pont20172017}
J.~Pont-Tuset, F.~Perazzi, S.~Caelles, P.~Arbeláez, A.~Sorkine-Hornung, and
  L.~Van~Gool, ``The 2017 davis challenge on video object segmentation,''
  \emph{arXiv}, 2017.

\bibitem{lin2019tsm}
J.~Lin, C.~Gan, and S.~Han, ``Tsm: Temporal shift module for efficient video
  understanding,'' in \emph{ICCV}, 2019.

\bibitem{feichtenhofer2019slowfast}
C.~Feichtenhofer, H.~Fan, J.~Malik, and K.~He, ``Slowfast networks for video
  recognition,'' in \emph{ICCV}, 2019.

\bibitem{bertasius2021space}
G.~Bertasius, H.~Wang, and L.~Torresani, ``Is space-time attention all you need
  for video understanding?'' in \emph{ICML}, 2021.

\bibitem{zhang2022bytetrack}
Y.~Zhang, P.~Sun, Y.~Jiang, D.~Yu, F.~Weng, Z.~Yuan, P.~Luo, W.~Liu, and
  X.~Wang, ``Bytetrack: Multi-object tracking by associating every detection
  box,'' in \emph{ECCV}, 2022.

\bibitem{cheng2022xmem}
H.~K. Cheng and A.~G. Schwing, ``Xmem: Long-term video object segmentation with
  an atkinson-shiffrin memory model,'' in \emph{ECCV}, 2022.

\bibitem{VVenC}
A.~Wieckowski, J.~Brandenburg, T.~Hinz, C.~Bartnik, V.~George, G.~Hege,
  C.~Helmrich, A.~Henkel, C.~Lehmann, C.~Stoffers, I.~Zupancic, B.~Bross, and
  D.~Marpe, ``Vvenc: An open and optimized vvc encoder implementation,'' in
  \emph{ICMEW}, 2021.

\bibitem{kasturi2008framework}
R.~Kasturi, D.~Goldgof, P.~Soundararajan, V.~Manohar, J.~Garofolo, R.~Bowers,
  M.~Boonstra, V.~Korzhova, and J.~Zhang, ``Framework for performance
  evaluation of face, text, and vehicle detection and tracking in video: Data,
  metrics, and protocol,'' \emph{TPAMI}, 2008.

\bibitem{dosovitskiy2020image}
A.~Dosovitskiy, L.~Beyer, A.~Kolesnikov, D.~Weissenborn, X.~Zhai,
  T.~Unterthiner, M.~Dehghani, M.~Minderer, G.~Heigold, S.~Gelly \emph{et~al.},
  ``An image is worth 16x16 words: Transformers for image recognition at
  scale,'' in \emph{ICLR}, 2021.

\bibitem{kingma2014adam}
D.~P. Kingma and J.~Ba, ``Adam: A method for stochastic optimization,'' 2014.

\bibitem{shao2020finegym}
D.~Shao, Y.~Zhao, B.~Dai, and D.~Lin, ``Finegym: A hierarchical video dataset
  for fine-grained action understanding,'' in \emph{CVPR}, 2020.

\bibitem{li2022nerv}
Z.~Li, M.~Wang, H.~Pi, K.~Xu, J.~Mei, and Y.~Liu, ``E-nerv: Expedite neural
  video representation with disentangled spatial-temporal context,'' in
  \emph{ECCV}, 2022.

\bibitem{chen2021nerv}
H.~Chen, B.~He, H.~Wang, Y.~Ren, S.~N. Lim, and A.~Shrivastava, ``Nerv: Neural
  representations for videos,'' \emph{NeurIPS}, 2021.

\bibitem{li2023high}
M.~Li, Y.~Shi, J.~Wang, and Y.~Huang, ``High visual-fidelity learned video
  compression,'' in \emph{ACM MM}, 2023.

\bibitem{xu2017video}
D.~Xu, Z.~Zhao, J.~Xiao, F.~Wu, H.~Zhang, X.~He, and Y.~Zhuang, ``Video
  question answering via gradually refined attention over appearance and
  motion,'' in \emph{ACM MM}, 2017.

\bibitem{he2024ma}
B.~He, H.~Li, Y.~K. Jang, M.~Jia, X.~Cao, A.~Shah, A.~Shrivastava, and S.-N.
  Lim, ``Ma-lmm: Memory-augmented large multimodal model for long-term video
  understanding,'' in \emph{CVPR}, 2024.

\bibitem{zhang2024psalm}
Z.~Zhang, Y.~Ma, E.~Zhang, and X.~Bai, ``Psalm: Pixelwise segmentation with
  large multi-modal model,'' in \emph{ECCV}, 2024.

\bibitem{chan2022basicvsr++}
K.~C. Chan, S.~Zhou, X.~Xu, and C.~C. Loy, ``Basicvsr++: Improving video
  super-resolution with enhanced propagation and alignment,'' in \emph{CVPR},
  2022.

\bibitem{chan2022generalization}
{Chan, Kelvin CK and Zhou, Shangchen and Xu, Xiangyu and Loy, Chen Change},
  ``On the generalization of basicvsr++ to video deblurring and denoising,''
  \emph{arXiv}, 2022.

\bibitem{chen2017rethinking}
L.-C. Chen, G.~Papandreou, F.~Schroff, and H.~Adam, ``Rethinking atrous
  convolution for semantic image segmentation,'' \emph{arXiv}, 2017.

\bibitem{xie2015holistically}
S.~Xie and Z.~Tu, ``Holistically-nested edge detection,'' in \emph{ICCV}, 2015.

\bibitem{simonyan2014very}
K.~Simonyan and A.~Zisserman, ``Very deep convolutional networks for
  large-scale image recognition,'' \emph{arXiv}, 2014.

\bibitem{oquab2023dinov2}
M.~Oquab, T.~Darcet, T.~Moutakanni, H.~Vo, M.~Szafraniec, V.~Khalidov,
  P.~Fernandez, D.~Haziza, F.~Massa, A.~El-Nouby \emph{et~al.}, ``Dinov2:
  Learning robust visual features without supervision,'' \emph{arXiv}, 2023.

\bibitem{van2017neural}
A.~Van Den~Oord, O.~Vinyals \emph{et~al.}, ``Neural discrete representation
  learning,'' \emph{NeurIPS}, 2017.

\bibitem{lu2020content}
G.~Lu, C.~Cai, X.~Zhang, L.~Chen, W.~Ouyang, D.~Xu, and Z.~Gao, ``Content
  adaptive and error propagation aware deep video compression,'' in
  \emph{ECCV}, 2020.

\bibitem{teed2020raft}
Z.~Teed and J.~Deng, ``Raft: Recurrent all-pairs field transforms for optical
  flow,'' in \emph{ECCV}, 2020.

\bibitem{dosovitskiy2015flownet}
A.~Dosovitskiy, P.~Fischer, E.~Ilg, P.~Hausser, C.~Hazirbas, V.~Golkov, P.~Van
  Der~Smagt, D.~Cremers, and T.~Brox, ``Flownet: Learning optical flow with
  convolutional networks,'' in \emph{ICCV}, 2015.

\bibitem{katharopoulos2020transformers}
A.~Katharopoulos, A.~Vyas, N.~Pappas, and F.~Fleuret, ``Transformers are rnns:
  Fast autoregressive transformers with linear attention,'' in \emph{ICML},
  2020.

\bibitem{lin2020m}
J.~Lin, D.~Liu, H.~Li, and F.~Wu, ``M-lvc: Multiple frames prediction for
  learned video compression,'' in \emph{CVPR}, 2020.

\bibitem{sheng2022temporal}
X.~Sheng, J.~Li, B.~Li, L.~Li, D.~Liu, and Y.~Lu, ``Temporal context mining for
  learned video compression,'' \emph{TMM}, 2022.

\bibitem{rombach2022high}
R.~Rombach, A.~Blattmann, D.~Lorenz, P.~Esser, and B.~Ommer, ``High-resolution
  image synthesis with latent diffusion models,'' in \emph{CVPR}, 2022.

\bibitem{fei2023generative}
B.~Fei, Z.~Lyu, L.~Pan, J.~Zhang, W.~Yang, T.~Luo, B.~Zhang, and B.~Dai,
  ``Generative diffusion prior for unified image restoration and enhancement,''
  in \emph{CVPR}, 2023.

\bibitem{heusel2017gans}
M.~Heusel, H.~Ramsauer, T.~Unterthiner, B.~Nessler, and S.~Hochreiter, ``Gans
  trained by a two time-scale update rule converge to a local nash
  equilibrium,'' \emph{NeurIPS}, 2017.

\end{thebibliography}
}
\vspace{-1cm}
\begin{IEEEbiography}
	[{\includegraphics[width=1in,height=1.25in,clip,keepaspectratio]{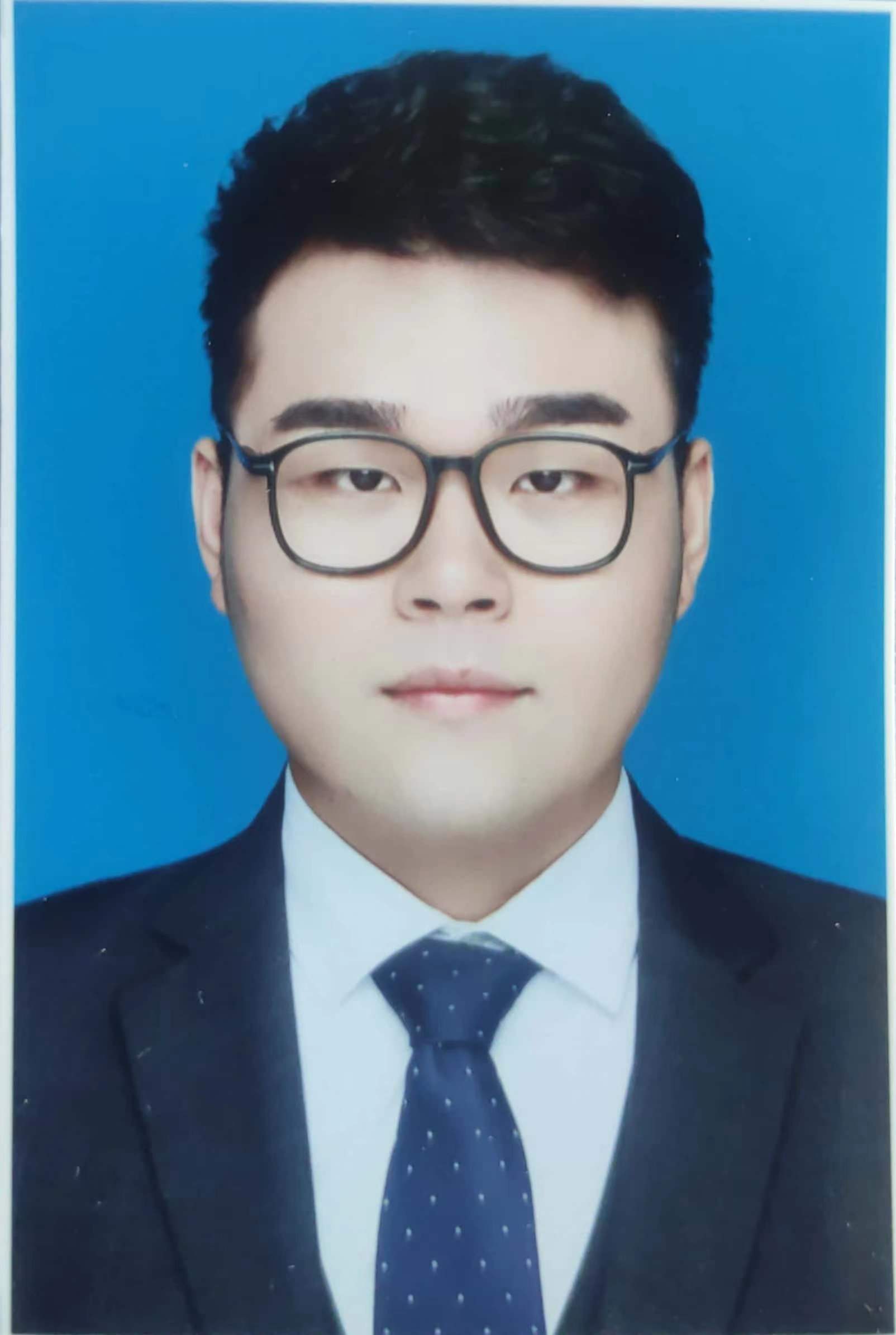}}]
	{Yuan Tian}
	received the B.Sc. degree in electronic engineering from Wuhan University, Wuhan, China, in 2017. He obtained the Ph.D. degree with the Department of Electronic Engineering, Shanghai Jiao Tong University, Shanghai, China, in 2023.
	Currently, he is a researcher at Shanghai Artificial Intelligence Laboratory.
	His works have been published in top-tier journals and conferences (e.g., TPAMI, Nature NPJ Digital Medicine, IJCV, TIP, CVPR, ICCV, ECCV and AAAI).
	His research interests include video processing, generative AI and medical AI.
\end{IEEEbiography}
\vspace{-1cm}
\begin{IEEEbiography}
	[{\includegraphics[width=1in,height=1.25in,clip,keepaspectratio]{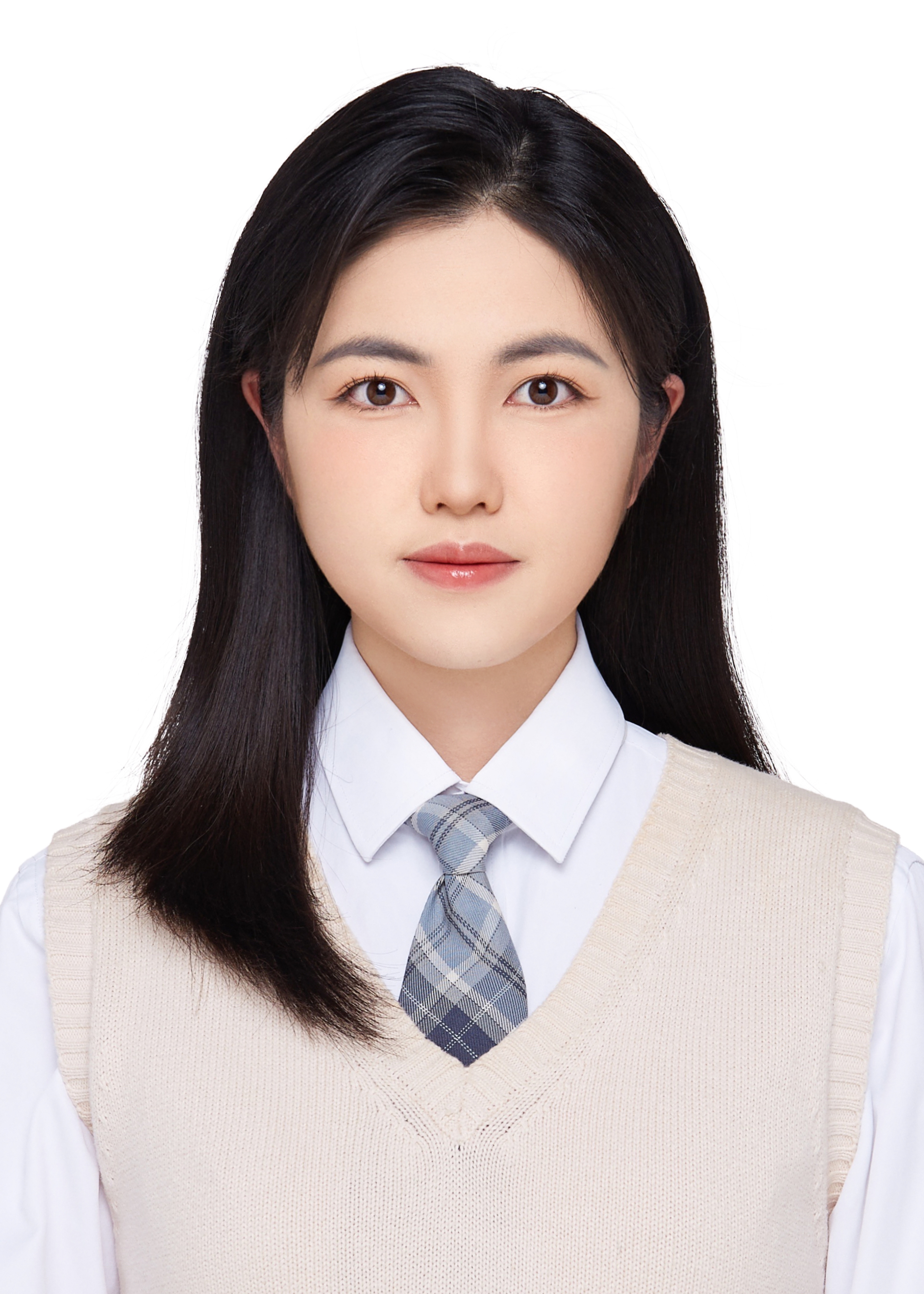}}]
	{Xiaoyue Ling}
	received the B.S. degree in communication engineering from Xidian University, Xi'an, China, in 2024. She is currently pursuing the Ph.D. degree in the Department of Electronic Engineering at Shanghai Jiao Tong University, Shanghai, China. Her research interests include video compression, video coding for machine, and semantic communication.
\end{IEEEbiography}
\vspace{-1cm}
\begin{IEEEbiography}
	[{\includegraphics[width=1in,height=1.25in,clip,keepaspectratio]{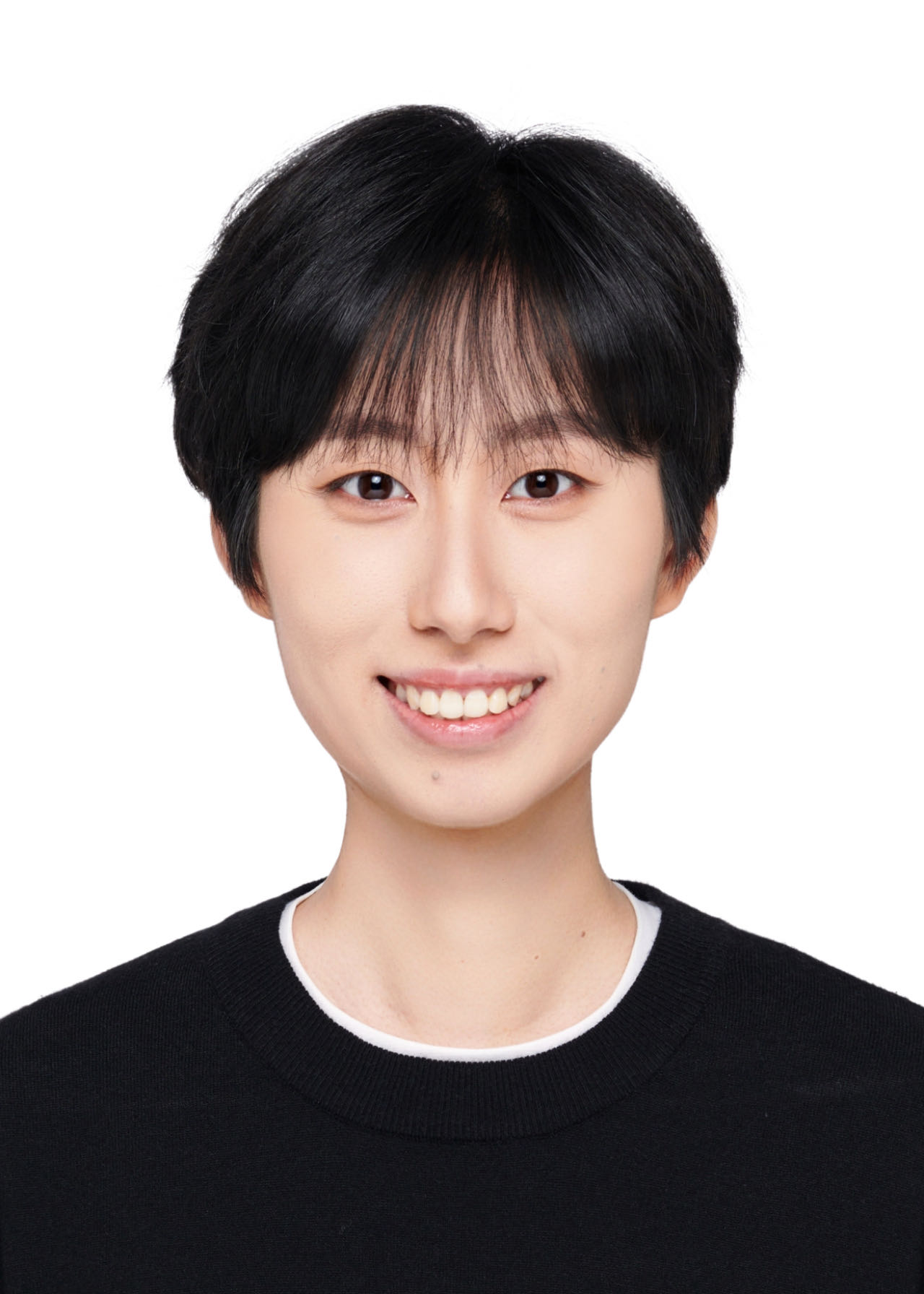}}]
	{Cong Geng}
	 received the B.Sc. degree in electronic engineering from the University of Electronic Science and Technology of China in 2016. She obtained the Ph.D degree with the Department of Electronic Engineering, Shanghai Jiao Tong University in 2023. Now she is an AI algorithm researcher at China Mobile Research Institute. Her research interests focus on generative models and their applications.
\end{IEEEbiography}
\vspace{-1cm}
\begin{IEEEbiography}
	[{\includegraphics[width=1in,height=1.25in,clip,keepaspectratio]{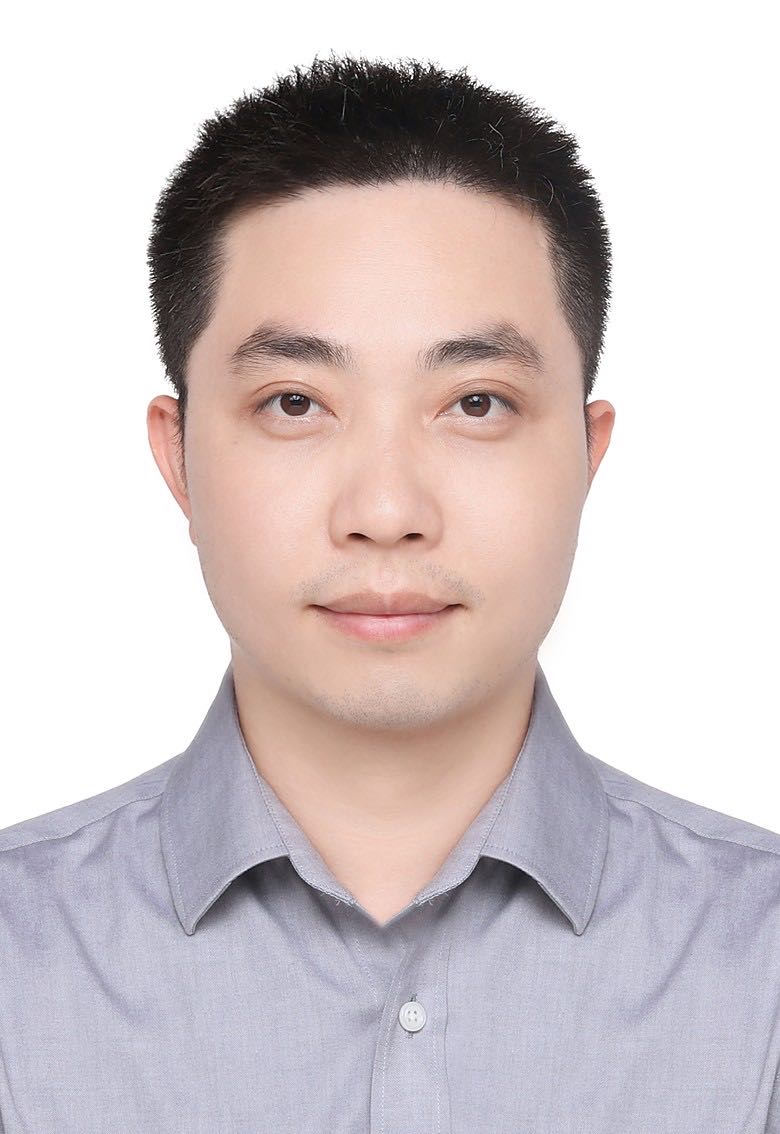}}]
	{Qiang Hu} (Member, IEEE) received the B.S. degree in electronic engineering from University of Electronic Science and Technology of China in 2013, and the Ph.D. degree from Shanghai Jiao Tong University in 2019. He is currently an Assistant Researcher at Cooperative Medianet Innovation Center, Shanghai Jiao Tong University. His research interests focus on 2D/3D video compression. His works have been published in top-tier journals and conferences, such as TIP, JSAC, CVPR.
\end{IEEEbiography}
\vspace{-1cm}
\begin{IEEEbiography}
	[{\includegraphics[width=1in,height=1.25in,clip,keepaspectratio]{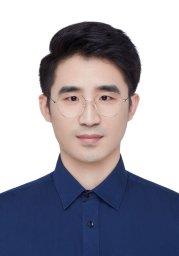}}]
	{Guo Lu}
	received the B.S. degree from the Ocean University of China in 2014 and the Ph.D. degree from Shanghai Jiao Tong University in 2020. Currently, he is an Assistant Professor with the Department of Electronic Engineering, Shanghai Jiao Tong University, Shanghai, China. His works have been published in top-tier journals and conferences (e.g., TPAMI, TIP, CVPR, and ECCV). His research
	interests include image and video processing, video compression, and computer vision.
\end{IEEEbiography}
\vspace{-1cm}
\begin{IEEEbiography}
	[{\includegraphics[width=1in,height=1.25in,clip,keepaspectratio]{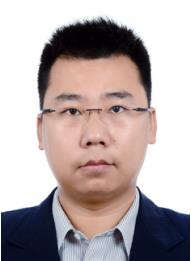}}]
	{Guangtao Zhai} \textit{IEEE Fellow}, received the B.E. and M.E. degrees from Shandong University, Shandong, China, in 2001 and 2004, respectively, and the Ph.D. degree from Shanghai Jiao Tong University, Shanghai, China, in 2009, where he is currently a Professor with the Institute of Image Communication and Information Processing. His research interests include multimedia signal processing and perceptual signal processing.
\end{IEEEbiography}

\end{document}